\documentclass[preprint,12pt,authoryear]{elsarticle}

\usepackage{lineno,hyperref}
\usepackage{subfigure}
\usepackage{booktabs}
\usepackage{multirow}
\usepackage{array}
\usepackage{amsmath}
\usepackage{caption}
\usepackage{enumerate}
\usepackage{rotating}

\journal{Journal of \LaTeX\ Templates}

\begin{document}

% change Figure 1 to Fig.1.
\captionsetup[figure]{
	labelfont={bf},
	name={Fig.},
	labelsep=period
}
\captionsetup[table]{
	labelfont={bf},
	labelsep=newline,
	singlelinecheck=false,
}

\begin{frontmatter}

\title{Reciprocal Translation between SAR and Optical Remote Sensing Images with Cascaded-Residual Adversarial Networks}
% \tnoteref{mytitlenote}

%\author{Shilei Fu\fnref{myfootnote}}
%\author[Feng Xu]{Elsevier Inc}
%\author[Ya-Qiu Jin]{Global Customer Service\corref{mycorrespondingauthor}}

\author{Shilei Fu, Feng Xu, and Ya-Qiu Jin}	
\address{Key Lab for Information Science of Electromagnetic Waves (MoE), Fudan University, Shanghai 200433, China.}


\begin{abstract}
Despite the advantages of all-weather and all-day high-resolution imaging, synthetic aperture radar (SAR) images are much less viewed and used by general people because human vision is not adapted to microwave scattering phenomenon. However, expert interpreters can be trained by comparing side-by-side SAR and optical images to learn the mapping rules from SAR to optical. This paper attempts to develop machine intelligence that are trainable with large-volume co-registered SAR and optical images to translate SAR image to optical version for assisted SAR image interpretation. Reciprocal SAR-Optical image translation is a challenging task because it’s raw data translation between two physically very different sensing modalities. Inspired by recent progresses in image translation studies in computer vision, this paper tackles the problem of SAR-optical reciprocal translation with an adversarial network scheme where cascaded residual connections and hybrid L1-GAN loss are employed. It is trained and tested on both spaceborne GF3 and airborne UAVSAR images. Results are presented for datasets of different resolutions and polarizations and compared with other state-of-the-art methods. The Fréchet inception distance is used to quantitatively evaluate the translation performance. The possibility of unsupervised learning with unpaired SAR and optical images is also explored. Results show that the proposed translation network works well under many scenarios and it could potentially be used for assisted SAR interpretation.
\end{abstract}


\end{frontmatter}

%\linenumbers

\section{Introduction}

Synthetic aperture radar (SAR) is capable of imaging at high resolution in all-day and all-weather conditions. As a cutting-edge technology for space remote sensing, it has found wide applications in earth science, weather change, environmental system monitoring, marine resource utilization, planetary exploration etc.
High resolution and multi-dimension are the two major trends of recent development of spaceborne SAR technology. Imaging resolution has been improved from ten-meter in 1990s (e.g. SIR-C/X-SAR), to meters in 2000s (e.g. Radarsat2), and to sub-meter in 2010s (e.g. TerraSAR-X). Higher resolution greatly enhances our capability to separate individual scatterers of one target, which opens up a new way to deterministically interpret the target image on scatterer-by-scatterer level. Multi-dimension includes multi-polarization, multi-temporal, multi-baselines, multi-frequency, multi-static etc. For example, spaceborne bistatic and multistatic SAR as demonstrated in \cite{zhang2016spaceborne,zhang2016synchronization} is one of the promising multi-dimension SAR technologies in the next decade. New imaging modalities and novel constellation concepts are still under development (\cite{guarnieri2017options, fuster2017interferometric}). Despite the rapid progresses in SAR imaging technologies, the bottleneck challenge remains in the interpretation of SAR imagery and it is becoming more and more urgent as a huge volume of SAR data is being acquired daily by numerous radar satellites in orbits.

Due to its distinct imaging mechanism and the complex electromagnetic (EM) wave scattering process, SAR exhibits very different imaging features from optical images. Some basic differences between SAR images and natural optical images are summarized in \autoref{table1}. Human’s visual system is adapted to the interpretation of optical images. SAR image is difficult to interpret by ordinary people. Although SAR images contain rich information about target and scene, such as geometric structure and material property, they can only be interpreted by well-trained experts. This has now become the major hindrance in utilization of existing SAR archives and further promotion of SAR applications.

\begin{table}
	\scriptsize
	\renewcommand\arraystretch{1.5}
	\setlength{\abovecaptionskip}{0pt}
	\setlength{\belowcaptionskip}{10pt}%设置标题与表格的距离
	\caption{Differences between SAR images and natural optical images.}
	\label{table1} 
	% \resizebox{\textwidth}{!}{
	\begin{tabular}{*{4}{m{1.1in}<{\centering}}}
%	\begin{tabular}{cccc}  
%		\toprule[2pt]
		\hline
		 & Optical images & SAR images & SAR unique phenomena\\
		\hline 
		Wave band & Visible light band & Microwave band & Discontinuity, Scintillation \\
%		\hline
		Focusing mechanism & Real aperture & Coherent synthetic aperture & Speckle noise \\
%		\hline
		Projection scheme & Elevation-Azimuth &	Range-Azimuth &	Layover, Foreshortening, Shadowing \\
%		\hline
		Resolution & Proportional to range & Invariant to range & No perspective distortion \\
%		\hline
		Data format & Color, Intensity & Phase, Amplitude, Polarization & Multi-channel, Complex \\
		\hline
%		\bottomrule[2pt]
	\end{tabular}

\end{table}

\begin{figure}[!htb]
	\scriptsize
	\centering
	\subfigure{
		\begin{minipage}[b]{0.15\linewidth}
			\includegraphics[width=0.8in,height=0.8in]{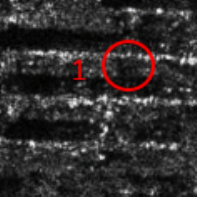}\vspace{4pt}
			\includegraphics[width=0.8in,height=0.8in]{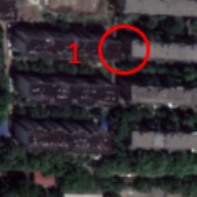}
			\centering{(a)}
		\end{minipage}
	}
	\subfigure{
		\begin{minipage}[b]{0.15\linewidth}
			\includegraphics[width=0.8in,height=0.8in]{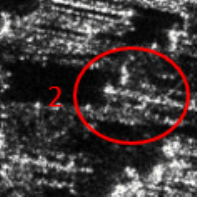}\vspace{4pt}
			\includegraphics[width=0.8in,height=0.8in]{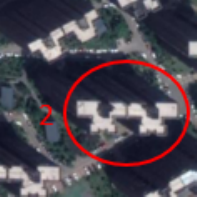}
			\centering{(b)}
		\end{minipage}
	}
	\subfigure{
		\begin{minipage}[b]{0.15\linewidth}
			\includegraphics[width=0.8in,height=0.8in]{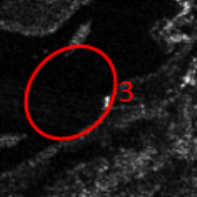}\vspace{4pt}
			\includegraphics[width=0.8in,height=0.8in]{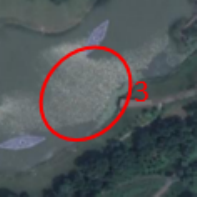}
			\centering{(c)}
		\end{minipage}
	}
	\subfigure{
		\begin{minipage}[b]{0.15\linewidth}
			\includegraphics[width=0.8in,height=0.8in]{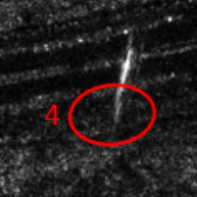}\vspace{4pt}
			\includegraphics[width=0.8in,height=0.8in]{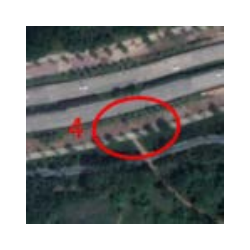}
			\centering{(d)}
		\end{minipage}
	}
	\caption{Distinctions between SAR and optical remote sensing images. The first row lists four example SAR images, and the second row lists the corresponding example optical images. Different qualities of translated optical images are induced by different losses. Each column lists \textbf{(a) low-rise buildings}, \textbf{(b) high-rise buildings}, \textbf{(c) waters} and \textbf{(d) roads}.}
	\label{fig:figure1} 
\end{figure}

Experts of SAR imagery interpretation are often trained by comparing the SAR image side-by-side with the corresponding optical image (e.g. \autoref{fig:figure1}). From such SAR-vs-Optical comparison experiences, experts conclude useful rules which translate between features in SAR and optical remote sensing images. Thereafter, they are able to direct interpret a new image from similar SAR sensors. Ideally, such training could be done in computers with artificial intelligence (AI) by leveraging the recent progresses in AI and deep learning technologies.

The major objective of this work is to develop deep learning application with large amount of co-registered SAR and optical images where SAR image can be translated to optical images and vice versa. The translated optical image can then be used in assisted interpretation of SAR image by ordinary people. Imagine that, with such translation tool, any person without any background knowledge of radar, could be able to understand the primary information contained in SAR image. This could greatly promote the wide application and usage of future and existing archive of SAR remote sensing imagery. Other potential applications include facilitation of data fusion of optical and SAR images, e.g. translating optical image at an earlier date as the reference SAR image for SAR change detection, registering the unpaired SAR and optical images, integrating optical and microwave data into a single image to enhance multi-spectral features, etc.

Deep learning, in particular convolutional neural networks (CNNs), has revolutionized the computer vision regime since the first successful application of CNN in practical image classification task in 2012 \cite{krizhevsky2012imagenet}. It utilized stacked convolution and pooling layers to automatically extract features at different scales via supervised learning. Since 2014, CNN-based approaches have been applied in interpretation of SAR images (\cite{zhu2017deep}), including the typical tasks such as automatic target recognition (ATR) (\cite{song2017zero,chen2016target}), earth surface classification (\cite{zhang2017complex,zhou2016polarimetric}), speckle reduction (\cite{yue2018sar}), change detection (\cite{liu2018deep}), etc.

Generative adversarial networks (GANs) (\cite{goodfellow2014generative}) are a special type of CNNs which are often used to generate synthesized images using actor-critic scheme. It simultaneously trains a generative CNN, the ‘actor’, which tries to generate images as realistic as possible, and a discriminative CNN, the ‘critic’, which tries to identify the synthesized images from the real images. Some relevant works which employed GAN-like approaches to analyze SAR images are reviewed in subsection 2.1.

On the other hand, image translation itself is an interesting problem in computer vision. Some interesting approaches in this regard are reviewed in subsection 2.2. Most of these image translation studies deal with the problem of style transfer where the objective is to change the style of image, e.g. from optical to cartoon picture, or from semantic map to optical picture. 

However, the problem of translation between SAR and optical images is considered slightly more difficult. It involves with two distinct types of raw sensor data. \autoref{table1} lists several major differences of SAR and optical imaging mechanisms and the corresponding distinct phenomena. As a result, the information contents in SAR and optical image are partial overlapped and partial exclusive, which means that only part information is observed by both sensors and each sensor observes other information that is not observable by the other sensor. A successful translation algorithm should be able to: 1) convert the common part of information content from one sensing modality to the other; and ideally, 2) generate the new information content from learned experiences. The former functionality is similar to style-transfer, while the latter is different. We believe that such cross-modality raw data translation requires a specially designed and fine-tuned network scheme and a large volume of co-registered image pairs as training data.

To illustrate the difficulties in translating across the two sensing modalities, \autoref{fig:figure1} gives four example image pairs, i.e. low-rise buildings, high-rise buildings, waters and roads. The differences between SAR and optical images are marked in red circles. For low-rise buildings, the gap between houses cannot be distinguished in SAR images. For high-rise buildings, the projection directions of buildings differ in the two sensing modalities. And buildings in SAR images scatter heavily and have severe layover effect with nearby buildings. For waters, ripples in optical images may not appear in SAR images. For roads, small roads are not nearly noticeable in SAR images due to layovers and shadows.

A SAR-Optical image reciprocal translation GAN architecture is proposed in this paper. It follows the typical image translation GAN architecture (e.g. \cite{isola2017image,zhu2017unpaired,jin2017deep,zhu2017toward}) employing a CNN as the discriminator and a specially-designed network as the generator (translator). The translator uses the multi-scale encoder-decoder CNN as the backbone and incorporates novel multi-scale cascaded-residual connections. To reduce the instability during the training of GAN, a hybrid loss function is used to train the generator which contains two parts: the GAN loss back-propagated from discriminator output, and the L1-distance loss directly applied to the generated sample and true sample.

The proposed method is verified on a large volume SAR-Optical image pairs, i.e. ~10000 samples of 256×256 size patch. The dataset covers different urban/suburban regions and mainly contains earth surfaces such as built-up areas, roads, vegetation, waters and farmlands. Appearances of these terrain objects, e.g. buildings, have great diversity which makes the training and testing more generalizable. The algorithm is tested at different resolutions. The Fréchet inception distance (FID) (\cite{heusel2017gans}) is used as the quantitative measure of the similarity between the reconstructed and the true image. For low-resolution (6m, 10m), the reconstructed result appears to be very similar to ground truths and the FID indicates very high similarity. For high-resolution (0.5m), the reconstructed result appears to be reasonable but not exactly captures the fine geometric features of manmade objects such as buildings and roads. This indicates that the mutual-exclusive part of information content in two sensing modalities becomes significant at higher resolutions. Overall, the translated high resolution optical image can partially serve the purpose of assisted interpretation of SAR images.

The major contributions of this paper are as follows:  
\begin{itemize}
	\item A modified image translation GAN architecture with multiscale cascaded residual connections is proposed for raw image translation between two very different sensing modalities, SAR and optical sensor.
	\item Experiment results on large volume of dataset demonstrate good visual quality and variety that can be achieved by the proposed network. Extensive analyses are conducted with quantitative metrics on space-borne and airborne SAR images. Different factors are analyzed including resolutions, targets, polarization, frequency bands, input image scales etc.
	\item  An extension towards unsupervised learning is tested with the CycleGAN loop (\cite{zhu2017unpaired}). Results demonstrate that using large volume of unpaired SAR and optical images, the performance can be further improved.
\end{itemize}

This paper is organized as follows. Section 2 reviews the relevant recent studies about image translation and SAR generation. Section 3 presents the proposed translation network architecture, loss function and training techniques. Experiments with real SAR images are carried out and results are presented and evaluated in Section 4. Finally in Section 5, conclusions are drawn and the future perspectives of application are discussed.

\section{Related works}
\subsection{GANs in SAR image analysis}
GANs or generative neural networks have been applied in various tasks in SAR image analysis. For instance, \cite{song2017zero} proposed zero-shot learning scheme for SAR ATR which is able to extract target orientation and orientation-invariant intrinsic features via training SAR generative networks. \cite{Ma2018Super} proposed a novel super-resolution method namely dense residual GAN and utilized the memory mechanism to extract hierarchical features for better reconstruction of remote sensing images. \cite{guo2017synthetic} employed GAN to synthesize the desired SAR images according to angle information and its feasibility was validated by comparison with real images and ray-tracing results. \cite{gao2018semi} introduced the multi-classifier to the discriminator so that the images’ labels of the Moving and Stationary Target Acquisition and Recognition (MSTAR) were made the most of to synthesize realistic SAR targets. \cite{hughes2018mining} produced hard-negative samples using GAN and validated its applicability to improve training in data sparse applications such as SAR-optical image matching. \cite{ao2018dialectical} proposed a “Dialectical GAN” based on Spatial Gram matrices and a WGAN-GP framework to transfer low-resolution Sentinel-1 to high-resolution TerraSAR-X images.

\subsection{Image translation}
Image translation is conversion between two types of images. It constructs an intermediate latent space, which maps the two image fields ${{X}_{1}}$, ${{X}_{2}}$ to the latent space Z through two encoders and then reconstructs ${{X}_{2}}$, ${{X}_{1}}$ respectively through decoders (\cite{liu2017unsupervised}). Currently, the mainstream image translation methods are based on GANs. Compared with traditional loss functions such as L2 and L1 norms, GANs can generate sharper and more realistic images. Slightly different from conventional GANs, image translation networks are conditional GANs (\cite{mirza2014conditional}) and require the original image as input. Existing studies of image translation can be mainly divided into two categories. 

The first category employs end-to-end image mapping strategy which directly converts the original image to the translated domain. \cite{taigman2016unsupervised} translated optical images to cartoon images using the domain transfer network. \cite{isola2017image} proposed the Pix2Pix network to translate optical aerial image to maps. It is based on U-Net architecture (\cite{ronneberger2015u}). \cite{zhu2017unpaired} proposed CycleGAN for unpaired image-to-image translation and changed the properties of targets in the images, such as the animal categories and the seasons of the scenes. \cite{liu2017unsupervised} made a shared-latent space assumption based on CycleGAN. \cite{chen2017photographic} synthesized photographic urban scenes with semantic segmentation maps using a cascaded refinement structure. 

The second category employs the style transfer strategy. It assumes that the images in two different domains share the same large-scale features, but have distinct small-scale features. Thus, it first separates the style from the contents of images and then replaces the style of the original image with the translated style. It assumes that the contents of an image are conveyed in the low-frequency large-scale edges while the styles are represented by the high-frequency small-scale textures. \cite{gatys2016image} proposed the Gram Matrix, the correlation coefficient matrix between each feature maps as extracted by CNNs, to represent the style of images. \cite{johnson2016perceptual} trained a feed-forward network to solve the optimization problem proposed by Gatys in real-time rather than adjusting the input according to the target. FaderNets, proposed by \cite{lample2017fader}, trained the discriminator to disentangle some specific style attributes and invariant intrinsic features from the encoded representation.

\subsection{Fusion of SAR and optical images}
Many works have been carried out in the multi-sensor image fusion regime (\cite{byun2013area}). The main objective of data fusion is to integrate complementary information from multi-sensor images of the same region into an enhanced image which appears better than any of the original ones. Hybrid pansharpening method, the weighted combination method (WC method), the integration method (MR method) based on the magnitude ratio of the two images are often employed (\cite{byun2013area}). The fused images are well-defined and diversely textured. \cite{garzelli2002wavelet} leveraged co-registered SAR image to improve the quality of optical images, which extracts specific information from SAR image and complements with optical image so that the targets could appear clearer.

Another type of fusion work is SAR and optical image registration. The focus is to explore the consistent features between the two sensing modalities. \cite{fan2018sar} designed a uniform nonlinear diffusion-based Harris feature extraction method to explore many more well-distributed feature points with potential of being correctly matched. \cite{liu2018deep} transformed the two types of images into a feature space where their feature representations became more consistent using a deep convolutional coupling network. \cite{merkle2018exploring} synthesized artificial SAR-like patches from optical images and matched them with the true SAR patches utilizing NCC, SIFT or BRISK. Especially, Liu’s method is very instructive for the content consistence of the feature space in the SAR and optical image translation.

A third type of fusion is to combine multi-temporal SAR and optical image to generate images of different observation time. \cite{he2018multi} used a conventional conditional GAN to generate the optical images from optical images at a different date with the aid of a SAR image acquired at the both dates. It also tried to directly generate optical image from a single SAR image but found that the existing GANs failed to do so. \cite{schmitt2018sen1} trained the network Pix2Pix on a large number of SEN1-2 patch pairs and got good predicted optical image patches. Some earlier attempts (\cite{wang2018generating,enomoto2018image}) try to converted coarse-resolution SAR or simulated SAR image to visible images using conditional GANs but results show that terrain objects such as buildings cannot be translated.

\section{SAR-Optical Reciprocal Translation Network}
\subsection{Translation framework}
\begin{figure}
	\centering 
	\includegraphics[width=4.5in]{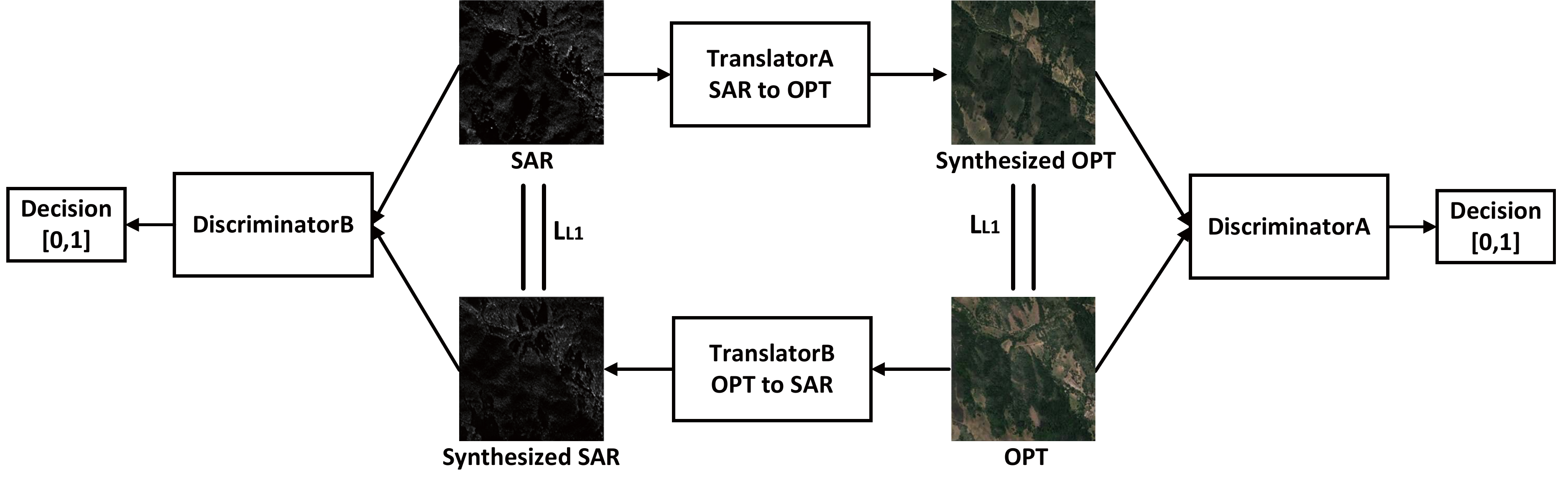} 
	\caption{Schematic diagram of the translation network (\cite{isola2017image}) during training. A pair of translators are trained together. Each translator consists of an encoder and a decoder. The two discriminators are trained separately. ‘SAR’, ‘OPT’, ‘Synthesized OPT’ and ‘Synthesized SAR’ respectively represent the true SAR image, the true optical image, the fake optical image and the fake SAR image. The two vertical lines connecting ‘SAR’ and ‘Synthesized SAR’ mean that the network should make them equal.}
	\label{fig:figure2} 
\end{figure}

The proposed framework is shown in \autoref{fig:figure2}. It has two reciprocal directions of translation, i.e. SAR to Optical and Optical to SAR. Each direction consists of two adversarial deep networks, i.e. a multi-scale convolutional encoder-and-decoder network as the translator (generator) vs. a convolutional network as the discriminator. The translator takes in SAR image, maps it to the latent space via the encoder, and then remaps it to a translated optical image. The discriminator takes in both the translated optical image and the true optical image which is co-registered with the original SAR image, and outputs the classification results. The discriminator learns to identify the translated optical images from the true optical images, while the translator network learns to convert the SAR image to an optical image as realistic as possible to fool the discriminator. On the other direction, the network is constructed exactly in the same manner with the only difference being optical as input and SAR as translated image.

The discriminator is a conventional CNN for binary classification task. The translator has multi-scale convolutional layers for encoder and decoder where direct paths are connected from the encoder to the decoder at different scales. Besides the direct paths in the latent space, residual connections in the input image space are further incorporated at each scales. A conventional binary classification loss is employed to train the discriminator, while its opposite loss, together with a L1 norm loss, is used to train the translator. These are explained in detail in the following subsections.

\subsection{Network architecture}
\begin{figure}
	\centering 
	\includegraphics[width=2.5in, angle=90]{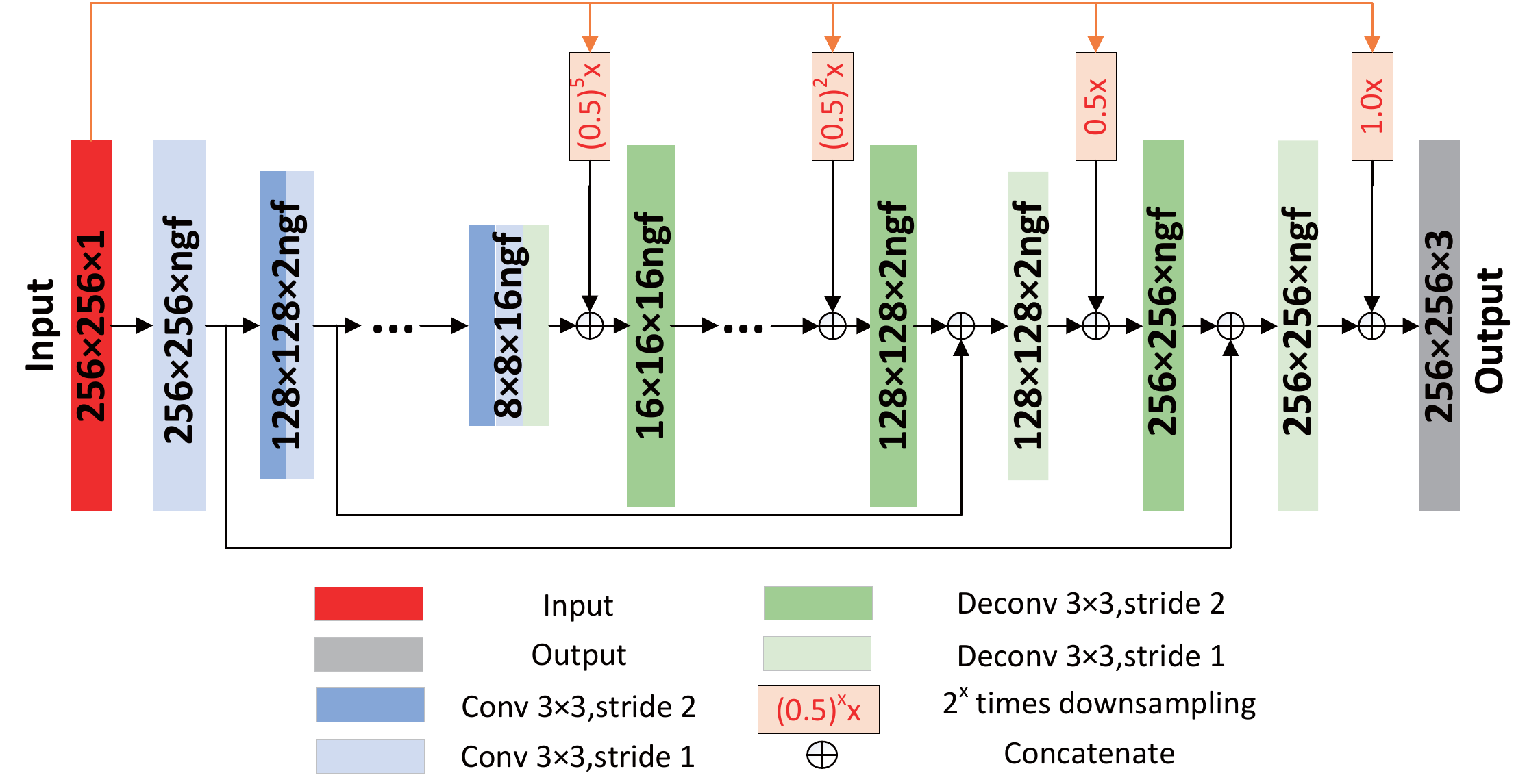} 
	\caption{Translator network architecture with cascaded-residual connections. The input data size is 256×256×1 and the output data size is 256×256×3. The first two numbers represent the size of the feature maps and the third number represents the channel of the feature map. The concatenation from the encoder and the input to the decoder is signified by lines with arrows.}
	\label{fig:figure3} 
\end{figure}

\autoref{fig:figure3} shows the architecture and parameters of the translator network, which is named as CRAN. It follows the main structure of the U-net (\cite{ronneberger2015u}) and the Pix2Pix (\cite{isola2017image}) with certain modifications. On the encoder side, the input image is convolved at one scale and downsampled to the next scale repeatedly for 6 times. On the decoder side, the latent feature map is deconvolved and upsampled back to the original scales. Notably, we include direct links from encoder to decoder. In addition, the network structure of CRAN contains multiscale cascaded residual connections from input to the multiple decoder stages. This is different from conventional ResNet connections such as the one employed in the encoder part in \cite{zhu2017toward}. On the other hand, the network employed in \cite{jin2017deep} contains a single skip connection from input to the last stage of U-net output which, according to our experiences, has low capability in generating image details than the cascaded pattern as used in this paper. We believe that such multi-scale cascaded residual connections are effective in generating vivid high resolution images. In order to increase the depth of the network, at each time upsampling the feature maps, it first concatenates the encoder’s feature maps to the current ones and deconvolves. Then concatenate the residual block to the former output feature maps and deconvolve again. This results in the increase of the decoder’s receptive field. Thus the receptive field of the encoder and that of the decoder will be asymmetrical, which may degrade performance. The solution is to convolve feature maps of each scale twice in the encoder.

\begin{figure}
	\centering 
	\includegraphics[width=1.5in, angle=90]{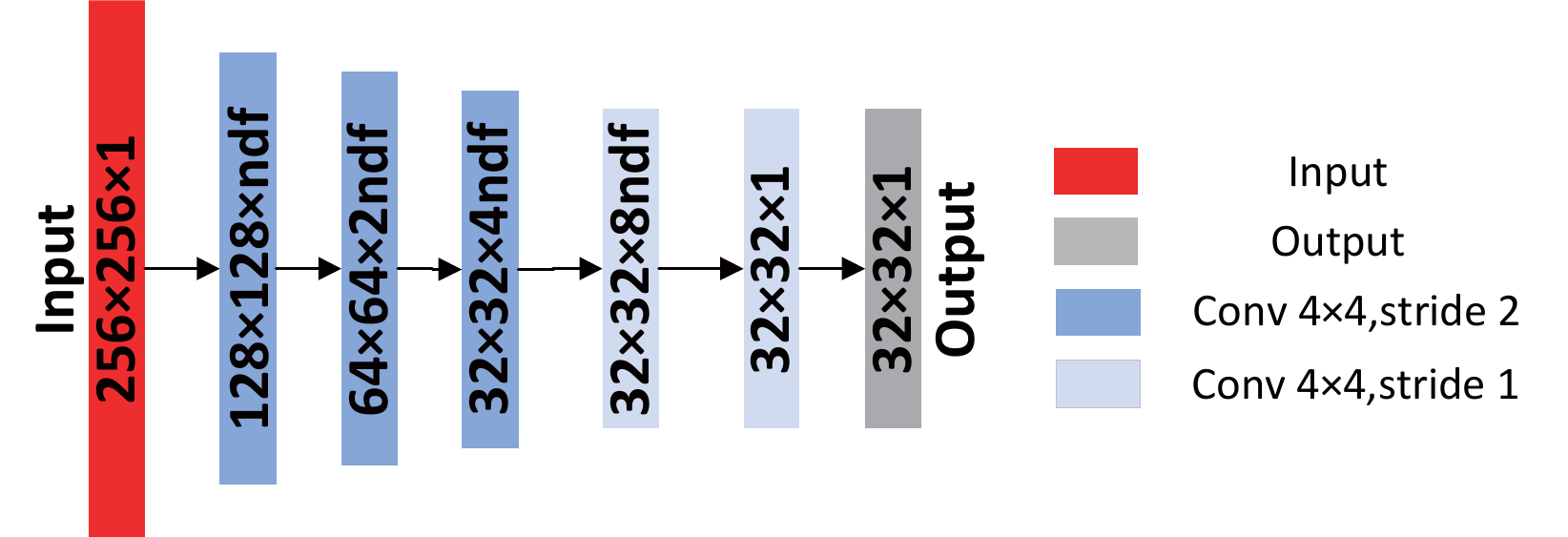} 
	\caption{Discriminator network architecture. The input data size is 256×256×1 and the output probability map size is 32×32×1. The first two numbers represent the size of the feature maps and the third number represents the channel of the feature map.}
	\label{fig:figure4} 
\end{figure}

Regarding the hyperparameters, in the translator, the convolutional kernel is 3×3, the encoder and the decoder each have 12 layers, and the receptive field per pixel of the input image is 191×191. In the discriminator, the kernel is 4×4, and it has 5 layers and the receptive field is 70×70. The benchmark number of feature maps in the generator is set to 50. The number of feature maps doubles at each downsampling and diminishes double after each upsampling. Those in the discriminator are respectively 64, 128, 256, 512 and 1. That means in the discriminator, the feature maps extracted from the input image are finally mapped into a 32×32 matrix and every value corresponds to a 70×70 patch of the input. By contrast, the difference between the discriminative matrix of the true image and that of the reconstructed image could determine how similar the spatial structures of the two images are. The total number of the generator’s weights is approximately 53.75 million and that of the discriminator is 2.76 million. The network architectures are depicted in detail in \autoref{fig:figure3} and \autoref{fig:figure4}.

\subsection{Loss functions}
Loss functions are critical for training of the networks. The discriminator is trained with a binary classification log-loss (\cite{goodfellow2014generative}), i.e.
\begin{flalign}
\label{equ:equation1}
& L(D)=-{{E}_{x\sim{{p}_{data}}(i)}}[logD(x)]-{{E}_{z\sim{{p}_{data}}(j)}}[log(1-D(T(z)))] &
\end{flalign}
where $i=0,1$ in ${{p}_{data}}(i)$ demonstrate the distributions of the true optical and SAR images respectively. ${{E}_{x\sim{{p}_{data}}(i)}}$ denotes that $x$ obeys the distribution ${{p}_{data}}(i)$, and ${{E}_{z\sim{{p}_{data}}(j)}}$ denotes that $z$ obeys the distribution ${{p}_{data}}(j)$. When $z$ denotes the original input SAR (or optical) image, $T(z)$ denotes the translated optical (or SAR) image and $x$ denotes the corresponding true optical (or SAR) image. $D(\cdot)$ denotes the output probability map of the discriminator. For the discriminator, minimizing $L(D)$ is equivalent to classifying $x$ as 1 and $T(z)$ as 0.

Following the adversary scheme (\cite{goodfellow2014generative}), the loss function of the translator is
\begin{flalign}
\label{equ:equation2}
& {{L}_{GAN}}(T)=-\sum\limits_{i}{{{E}_{z\sim{{p}_{data}}(i)}}[log(D(T(z)))]} &
\end{flalign}
where ${{L}_{GAN}}(T)$ is the sum loss of the two translated networks. Opposite to the goal of the discriminator, the translator is aimed at synthesizing ‘realistic’ images to fool the discriminator to classify them as 1. 

\cite{isola2017image} found that the adversary loss function is better to be hybrid with traditional loss, such as L1 or L2 loss. Therefore, L1 norm loss is used to hybrid with the GAN loss, i.e. L1 distance between the translated image $T(z)$ and the true image $x$:
\begin{flalign}
\label{equ:equation3}
& {{L}_{L1}}(T)=\sum\limits_{i,j}{{{E}_{x\sim{{p}_{data}}(i),z\sim{{p}_{data}}(j)}}[||x-T(z)||_{1}^{1}]} &
\end{flalign}

Combine the above two equations together with appropriate weights and derive the final loss function $L(T)$ of the translators.
\begin{flalign}
\label{equ:equation4}
& L(T)={{L}_{GAN}}(T)+\beta {{L}_{L1}}(T) &
\end{flalign}

$L(T)$ is the objective function for two translators, whose parameters are simultaneously updated. The two discriminators are allocated with the independent loss function $L(D)$ and trained separately.

\begin{figure}[!htb]
	\scriptsize
	\centering
	\subfigure{
		\begin{minipage}[b]{0.12\linewidth}
			\includegraphics[width=0.7in,height=0.7in]{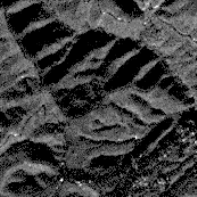}\vspace{4pt}
			\includegraphics[width=0.7in,height=0.7in]{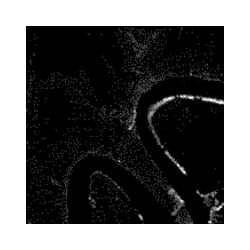}\vspace{4pt}
			\includegraphics[width=0.7in,height=0.7in]{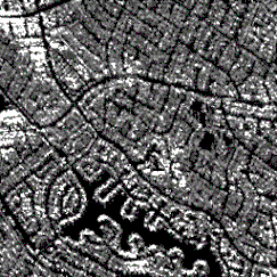}
			\centering{(a)}
		\end{minipage}
	}
	\subfigure{
		\begin{minipage}[b]{0.12\linewidth}
			\includegraphics[width=0.7in,height=0.7in]{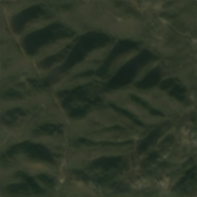}\vspace{4pt}
			\includegraphics[width=0.7in,height=0.7in]{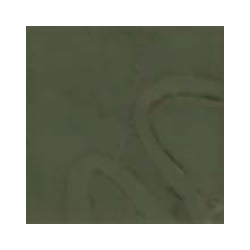}\vspace{4pt}
			\includegraphics[width=0.7in,height=0.7in]{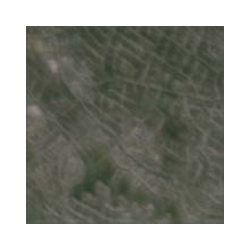}
			\centering{(b)}
		\end{minipage}
	}
	\subfigure{
		\begin{minipage}[b]{0.12\linewidth}
			\includegraphics[width=0.7in,height=0.7in]{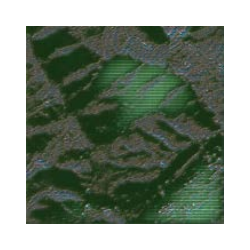}\vspace{4pt}
			\includegraphics[width=0.7in,height=0.7in]{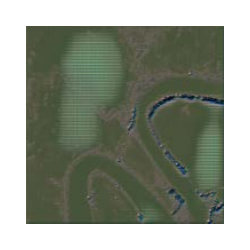}\vspace{4pt}
			\includegraphics[width=0.7in,height=0.7in]{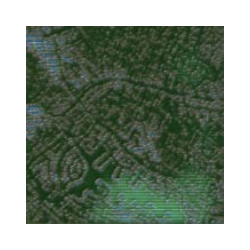}
			\centering{(c)}
		\end{minipage}
	}
	\subfigure{
		\begin{minipage}[b]{0.12\linewidth}
			\includegraphics[width=0.7in,height=0.7in]{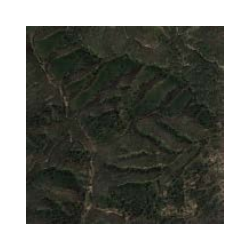}\vspace{4pt}
			\includegraphics[width=0.7in,height=0.7in]{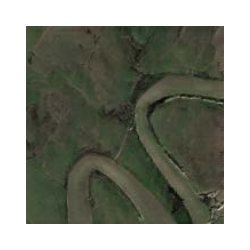}\vspace{4pt}
			\includegraphics[width=0.7in,height=0.7in]{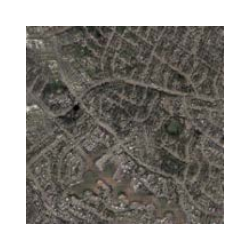}
			\centering{(d)}
		\end{minipage}
	}
	\subfigure{
		\begin{minipage}[b]{0.12\linewidth}
			\includegraphics[width=0.7in,height=0.7in]{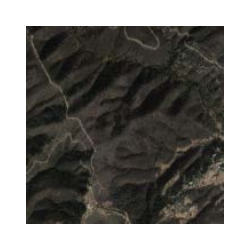}\vspace{4pt}
			\includegraphics[width=0.7in,height=0.7in]{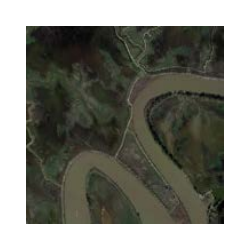}\vspace{4pt}
			\includegraphics[width=0.7in,height=0.7in]{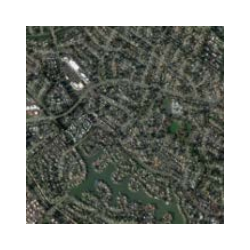}
			\centering{(e)}
		\end{minipage}
	}
	\caption{Different qualities of translated optical images are induced by different losses. The first column lists the \textbf{(a) input SAR images}; the intermediate three columns are respectively \textbf{(b) translated optical images with L1}, \textbf{(c) translated optical images with GAN} and \textbf{(d) translated optical images with L1+GAN}; the last column are the corresponding \textbf{(e) optical ground truths}.}
	\label{fig:figure5} 
\end{figure}

A quick experiment is conducted to show the efficacy of proposed loss function. Different losses contribute differently to the qualities of reconstructed results (\cite{isola2017image}). In \autoref{fig:figure5}, it is found that reconstructed optical images trained under L1-only loss are blurred and low-frequency features such as contours can be learned while high-frequency fine textures are missing. The model trained under GAN-only loss can learn the details and the targets are more prominent, but large-scale smooth features and their spatial distributions are not well reconstructed. We also notice that some artifacts appear in homogenous areas such as water. Besides, training with GAN-only loss often encounters the well-known ‘mode collapse’ problem (\cite{arjovsky2017wasserstein}). Mode collapse is a fatal training problem of GAN where $T(z)$ is collapsed to a fixed sample to maintain low loss but sacrificing the diversity. These issues can be alleviated by using the hybrid L1 and GAN loss, in which case, the synthesized images can have both low-frequency and high-frequency characteristics. Moreover, the network can also be trained more stably with the hybrid loss.

\subsection{Training Strategy}
\begin{figure}
	\centering 
	\includegraphics[width=4.5in]{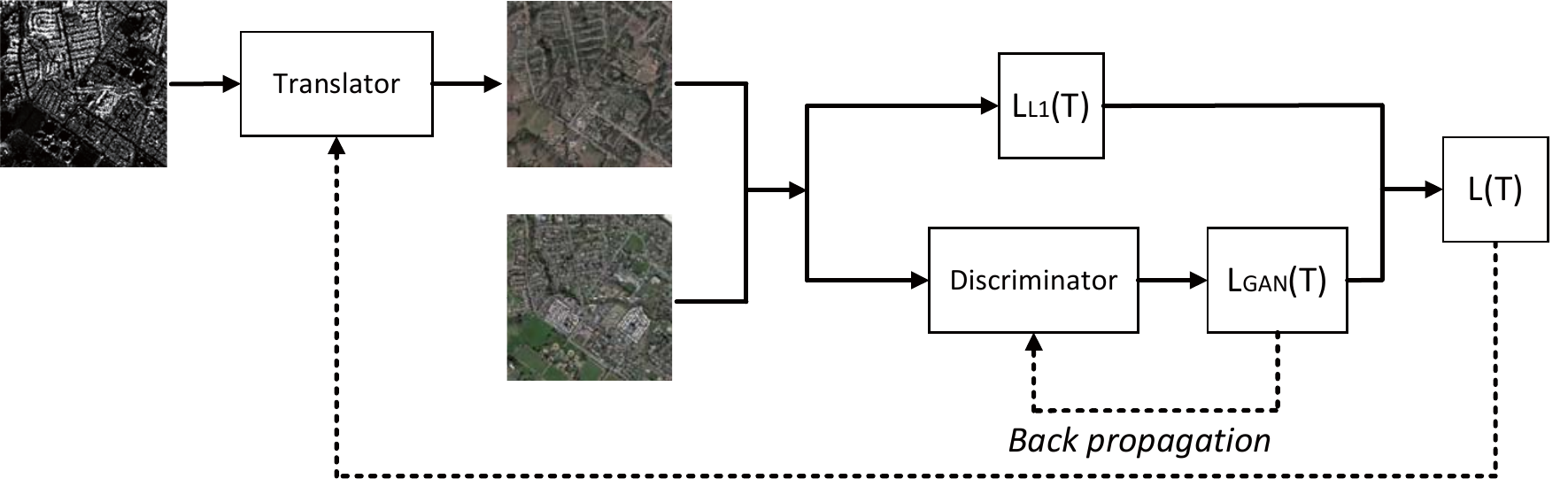} 
	\vspace{-0.1in}
	\caption{The conceptual process of training the adversarial networks. The left image is the real SAR image, the upper right is the synthesized optical image and the lower right is the real optical image.}
	\label{fig:figure6} 
\end{figure}

Stochastic gradient descent algorithm with adaptive moment estimation (Adam) can be used to train the two translators/discriminators simultaneously. Following GAN training strategy, one iteration consists of the following step (see \autoref{fig:figure6}):

a) Forward Pass – Weights of translators and discriminators are randomly initialized. A mini-batch of SAR images are then sent to the translator A to synthesize fake optical images, while a mini-batch of optical images are sent to the translator B to synthesize fake SAR images. Next, the fake and real optical images are sequentially sent to the same discriminator A, which generates two probability maps respectively. The fake and real SAR images are sent to the discriminator B and the discriminator B also generates corresponding probability maps.

b) Backward Pass – The two probability maps of optical images are compared in the loss to optimize the discriminator A, while those of SAR images are compared to optimize the discriminator B. The sigmoid function is selected as the activation function for the discriminator, which functions as a binary classifier. The discriminator is trained to distinguish the fake as 0 and the real as 1. The discriminator classifies the image patches separately. This not only limits the receptive field, but also provides more samples for the training (\cite{shrivastava2017learning}). Both of the two losses are also added as the GAN loss for the translators, which have to maximize them. That means the aim of both the translators is to generate sufficiently-realistic images to fool the discriminators. The real and fake ones are also compared directly to ensure the positional mappings of targets are correct. Thus, the joint losses are applied as the final loss function of the two translators. Then the backpropagation is applied to adjust the trainable parameters in the two translators simultaneously.

The forward process alternates with the backward process. The batch size is set as 1. The technique of GPU parallel acceleration with 4 NVIDIA Titan X is employed, which means four pairs of SAR and optical images are used to train the network each time simultaneously. After the gradients of the four threads are all calculated, the mean gradients are used to update the optimizers. The backward pass is a single thread. After finishing the back propagation, another four pairs of images are sent in. Traversing all the images is considered as an epoch. Then reshuffle the images and traverse next epoch.

We made several comparison experiments and set the parameter $\beta=20$ in Eq. \ref{equ:equation4}, with which the initial values of ${{L}_{GAN}}(T)$ and ${{L}_{L1}(T)}$ are approximately equal and the model is more stable and generates better results. We set the learning rate to 0.0002. Adam optimizer with $\beta1=0.5$ is used. The input images are linearly mapped to the interval $[-1,1]$. Leaky ReLU is selected as the activation function. Batch normalization is used before the activation function except the first or last layer. All the trainable parameters are initialized as the truncated normal distribution with mean 0 and standard deviation 0.02. These hyperparameters are mainly selected based on the implementation of Pix2Pix (\cite{isola2017image}). When the batch size is set too small, due to the difference between training samples, a slight oscillation occurs in the gradient decent and the curve of the loss convergence appears steady declining with oscillations. Early stop is adopted during training. When the training loss does not decrease for four epochs in a row, the training is forced to stop.

In terms of training time, it takes only about 0.4 second per batch and 18 minutes to run through an epoch (10284 pairs of samples). Another 3 minutes are needed to test the test samples and save the test results and the checkpoint of the model. So it costs around 21 minutes to finish training and testing for an epoch.

\subsection{Towards unsupervised learning}
\begin{figure}
	\centering 
	\includegraphics[width=4.5in]{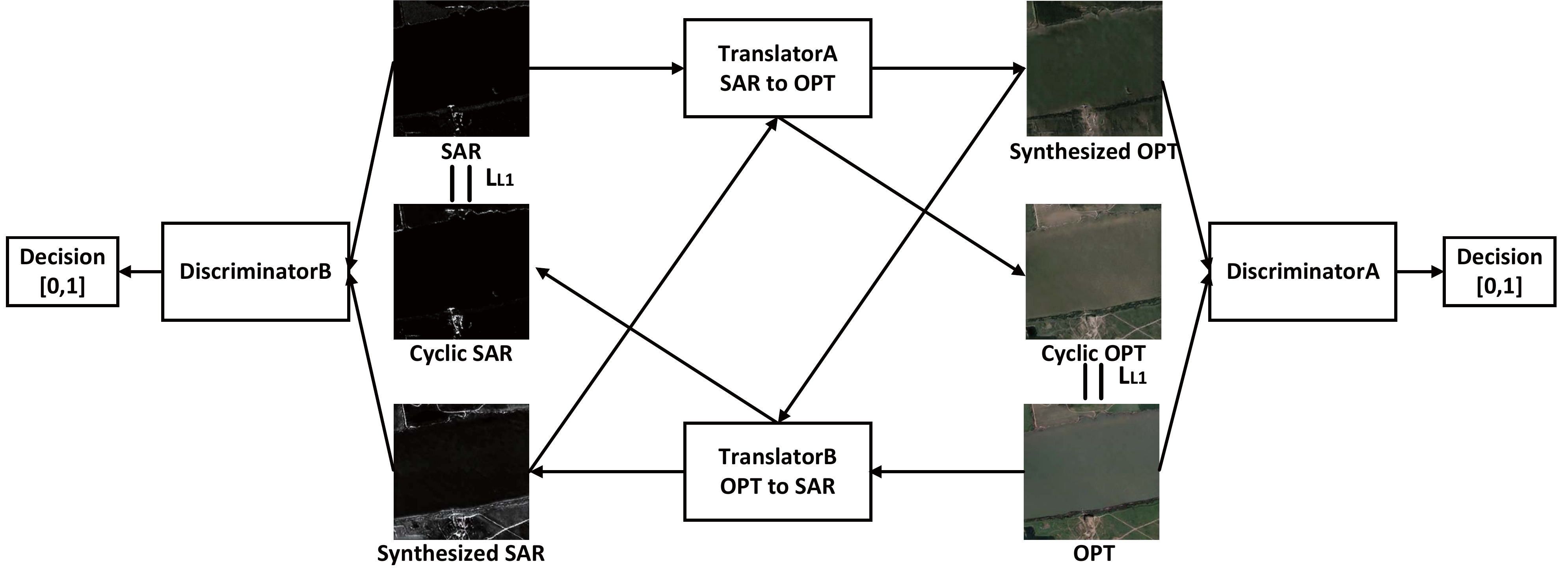} 
	\vspace{-0.1in}
	\caption{Modified network scheme for unsupervised learning with CycleGAN loops (\cite{zhu2017unpaired}).}
	\label{fig:figure7} 
\end{figure}

Supervised learning with well co-registered optical and SAR image pairs produces good results. However, such dataset is not always available and even available, would require a significant amount of effort for image registration. Thus, this paper also explores the possibility of unsupervised learning with unpaired SAR and optical images. CycleGAN (\cite{zhu2017unpaired}) proposes a cyclic loop which could be leveraged for this purpose. As shown in \autoref{fig:figure7}, the SAR image is first fed to the translator A and synthesizes fake optical image. Then the fake optical image is used to synthesize the cyclic fake SAR images by the translator B. On the other hand, the optical image is used to synthesize fake SAR image which is then further used to synthesize the cyclic fake optical images. The cyclic images are compared with the corresponding true images in a pixel-by-pixel fashion, while the synthesized fake images are fed into the ‘critic’ discriminator networks. The translator A and translator B networks are trained alternatively during these two loops together with the discriminator networks. Later in subsection 4.5, we demonstrate how such unsupervised learning could further improve the performance of a translator initially trained with a small number of co-registered image pairs.

\section{Experiments and Analyses}
\subsection{Datasets}
SAR data used in this study mainly comes from the spaceborne GF3 SAR from China (\cite{Gaofen3data}) and the airborne UAVSAR system from NASA (\cite{UAVSARdata}). The information of those data used in our experiments is listed in the following \autoref{table2}. Optical data used is downloaded from Google Map around November, 2018, with pixel resolution 0.51m for GF3 SAR data and 1.02m for UAVSAR data respectively.

%\begin{table}
%	\scriptsize
%	\caption{Information about the two datasets employed for experiments.}
%	\label{table2} \centering %
%	\begin{tabular}{*{4}{m{1.0in}<{\centering}}}
%		\toprule[2pt]
%		   &  & UAVSAR & GF3 \\
%		\hline
%		\multirow{7}*{SAR} & Resolution & 6m & 0.51m \\
%		\cline{2-4}
%		    & Polarization & Quad-Pol & HH or VV \\
%		 \cline{2-4}
%		    & Angle of Incidence & $90^{\circ}$ & $40.6642^{\circ}$ and $36.0820^{\circ}$ \\
%		 \cline{2-4}
%		    & Acquisition Mode & PolSAR & SL \\
%		 \cline{2-4}
%		    & Frequency Band & 80MHz & 240MHz \\
%		 \cline{2-4}
%		    & Day of Acquisition & 2010-04-09, 2013-05-13 and etc. & 2017-01-02 and 2016-08-15 \\ 
%		 \cline{2-4}
%		    & Location & California, US & Wuhan and Hefei, China \\
%		\hline
%		\specialrule{0em}{3pt}{3pt}
%		\multirow{2}*{Optical} & Resolution & 1.02m & 0.51m \\
%		\cline{2-4}
%		& Day of Acquisition &  & \\
%		\bottomrule[2pt]
%	\end{tabular}
%\end{table}
\begin{table}[!htbp]
	\scriptsize
	\renewcommand\arraystretch{1.5}
	\setlength{\abovecaptionskip}{0pt}
	\setlength{\belowcaptionskip}{10pt}%设置标题与表格的距离
	\caption{Information about the two datasets employed for experiments.}
	\label{table2} 
\begin{tabular}{cccc}
%	\multicolumn{4}{l}{\scriptsize{\textbf{Table 2}}}\\
%	\multicolumn{4}{l}{\scriptsize{Information about the two datasets employed for experiments.}}\\\specialrule{0.05em}{3pt}{3pt}
	\hline	
	&  & UAVSAR & GF3 \\
	\hline
	\multirow{7}*{SAR} & Resolution & 6m & 0.51m \\
%	\cline{2-4}
	& Polarization & Quad-Pol & HH or VV \\
%	\cline{2-4}
	& Angle of Incidence & $90^{\circ}$ & $40.6642^{\circ}$ and $36.0820^{\circ}$ \\
%	\cline{2-4}
	& Acquisition Mode & PolSAR & SL \\
%	\cline{2-4}
	& Frequency Band & 80MHz & 240MHz \\
%	\cline{2-4}
	& Day of Acquisition & 2010-04-09, 2013-05-13, etc. & 2017-01-02, 2016-08-15 \\ 
%	\cline{2-4}
	& Location & California, US & Wuhan and Hefei, China \\
	\hline
	\specialrule{0em}{3pt}{3pt}
	\multirow{3}*{Optical} & Resolution & 1.02m & 0.51m \\
%	\cline{2-4}
	& Day of Acquisition & 2018-11-25 & 2018-06-05, 2018-05-28\\
%	\cline{2-4}
	& Geographic Coordinate System & WGS 84 & WGS 84 \\
	\hline
\end{tabular}
\end{table}

\begin{figure}
	\centering 
	\includegraphics[width=4.5in]{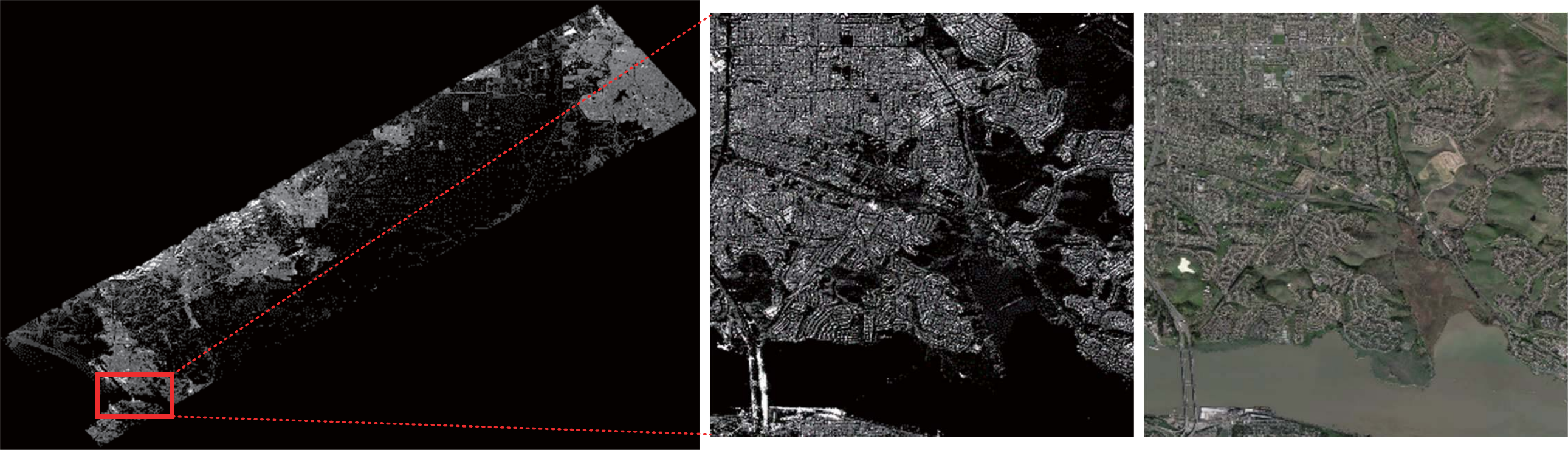} 
	\vspace{-0.1in}
	\caption{(left) UAVSAR image acquired in California, US. (center) Zoomed region. (right) Corresponding optical image of the zoomed region.}
	\label{fig:figure8} 
\end{figure}

UAVSAR (Uninhabited Airborne Vehicle Synthetic Aperture Radar) radar system is an L-band polarimetric instrument developed by NASA. As shown in \autoref{fig:figure8}, UAVSAR data used here mainly consists of five types of earth surfaces, buildings, vegetation (mountains are usually covered with trees and classified as vegetation here), farmlands, waters and deserts. It has pixel resolution of about 6.2×4.9m. The samples are 256×256 patches cropped from the original large SAR and optical images without any overlapping, which avoids the direct correlation between the training and test samples. Then we acquire a total of 12394 pairs of co-registered samples.

\begin{figure}
	\centering 
	\includegraphics[width=2.0in, angle=-90]{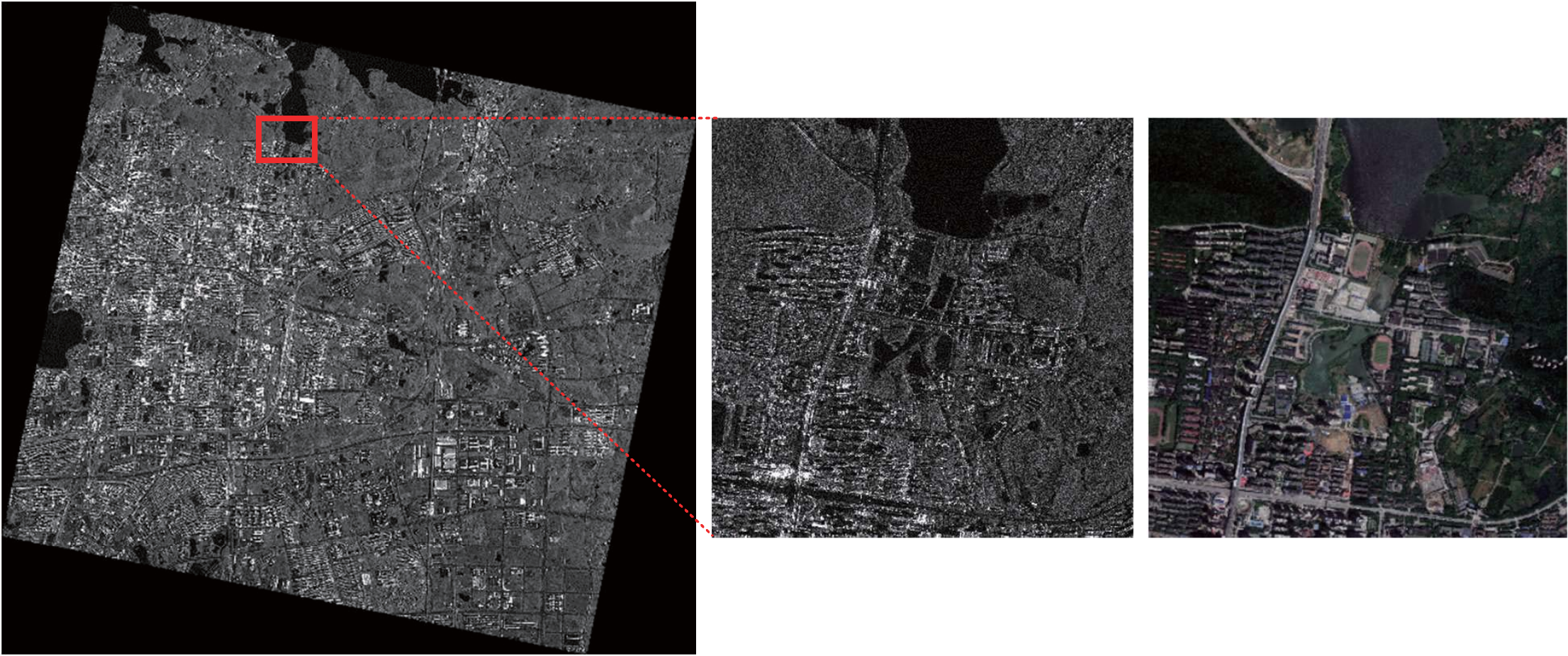} 
	\vspace{-0.1in}
	\caption{(left) A large SAR image with a 0.51m resolution in Hongshan District, Wuhan City, Hubei Province, China. (center) Zoomed region. (right) Corresponding optical image of the zoomed region.}
	\label{fig:figure9} 
\end{figure}

GF3 satellite is China’s first C-band multi-polarization SAR satellite. Two large scenes of GF3 images are used in the study with a resolution of 0.51m. The dataset contains different urban/suburban regions. In \autoref{fig:figure9}, the GF3 SAR image after geocoding is shown on the left. It mainly contains five terrain surfaces, i.e. buildings, roads (highways or overpass), vegetation, waters (lakes, rivers or seas) and farmlands. Buildings can be further divided into low-rise and high-rise buildings. 

A simple preprocessing step is to normalize the pixel value of the SAR images to $[-1,1]$. Due to the considerably large range of pixel values, we have to determine a suitable threshold value to normalize the SAR image without changing the contrast. The normalized pixel value of the SAR image is defined as the following Eq. \ref{equ:equation5}.
\begin{flalign}
\label{equ:equation5}
& \hat{x}=\left\{
\begin{matrix}
-1, & \text{if }x\leq 0;  \hfill \\
1, & \text{if }x\geq \bar{x};  \hfill \\
2x/\bar{x}-1, & \text{  }otherwise.  \hfill \\
\end{matrix}
\right. &
\end{flalign}
where $x$ and $\hat{x}$ represent the pixel values of SAR images before and after normalization. $\bar{x}$ is $\lambda $ times the mean value of the image $x$, defined as
\begin{flalign}
\label{equ:equation6}
& \bar{x}=\lambda (\sum\nolimits_{i=1}^{N}{{{x}_{i}}})/(N-n) &
\end{flalign}
where ${{x}_{i}}$ is the i-th pixel of the image $x$, $N$ is the total number of pixels and $n$ is the total number of pixels in the element 0. Here set $\lambda=2000$.

Another preprocessing step is to perform speckle filtering on the GF3 SAR images using a fast nonlocal despeckling filter (\cite{cozzolino2014fast}). We found that speckle filtering can improve the quality of the final synthesize images. We have a total of 12854 pairs of co-registered samples, $20\%$ of which are randomly selected as test samples while the rest as training samples. During the preparation of the dataset, it is found that, due to the difference of acquisition time of SAR and optical images, some new buildings shown in the recent optical images were not captured in the SAR image. This may adversely affect the final results.

\subsection{Quantitative evaluation}
Here we design experiments to test the performance of our model for SAR images of different resolutions and different polarization modes. The experiment for different resolutions adopts medium resolution UAVSAR and high resolution GF3 datasets respectively; the experiment for different polarization modes uses single-polarized and full-polarized UAVSAR data.
\subsubsection{Resolution}
\begin{figure}[!htb]
	\scriptsize
	\centering
	\subfigure{
		\begin{minipage}[b]{0.12\linewidth}
			\includegraphics[width=0.7in,height=0.7in]{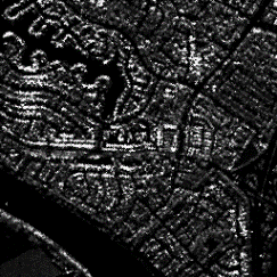}\vspace{4pt}
			\includegraphics[width=0.7in,height=0.7in]{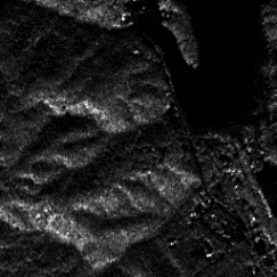}\vspace{6pt}
			\includegraphics[width=0.7in,height=0.7in]{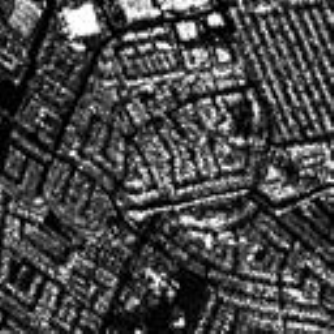}\vspace{4pt}
			\includegraphics[width=0.7in,height=0.7in]{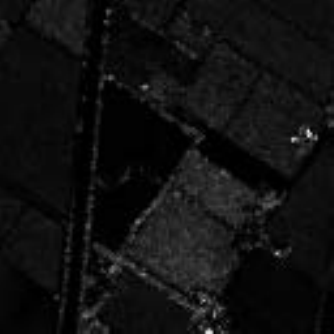}
			\centering{(a)}
		\end{minipage}
	}
	\subfigure{
		\begin{minipage}[b]{0.12\linewidth}
			\includegraphics[width=0.7in,height=0.7in]{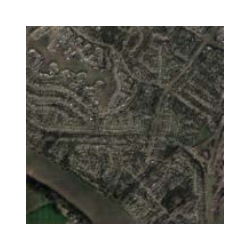}\vspace{4pt}
			\includegraphics[width=0.7in,height=0.7in]{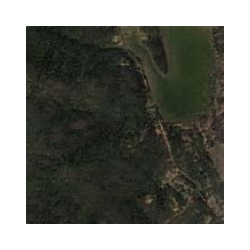}\vspace{6pt}
			\includegraphics[width=0.7in,height=0.7in]{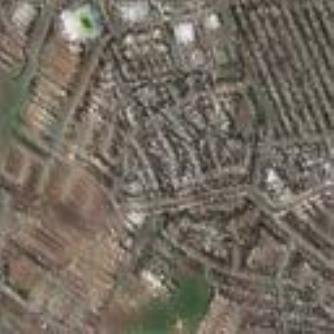}\vspace{4pt}
			\includegraphics[width=0.7in,height=0.7in]{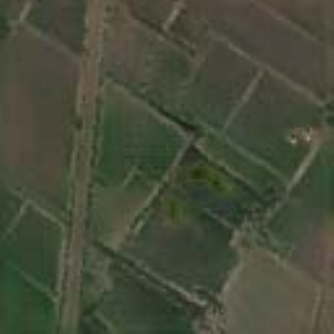}
			\centering{(b)}
		\end{minipage}
	}
	\subfigure{
		\begin{minipage}[b]{0.12\linewidth}
			\includegraphics[width=0.7in,height=0.7in]{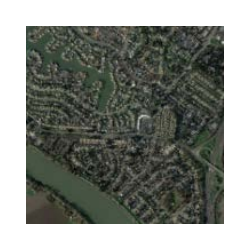}\vspace{4pt}
			\includegraphics[width=0.7in,height=0.7in]{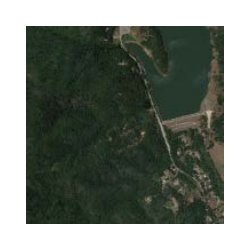}\vspace{6pt}
			\includegraphics[width=0.7in,height=0.7in]{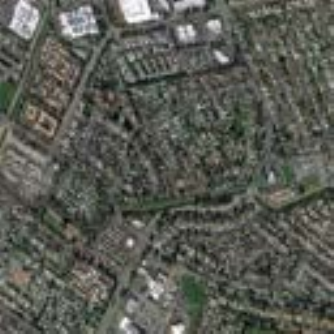}\vspace{4pt}
			\includegraphics[width=0.7in,height=0.7in]{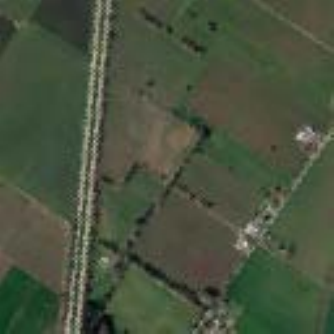}
			\centering{(c)}
		\end{minipage}
	}
	\subfigure{
		\begin{minipage}[b]{0.12\linewidth}
			\includegraphics[width=0.7in,height=0.7in]{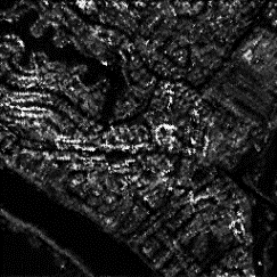}\vspace{4pt}
			\includegraphics[width=0.7in,height=0.7in]{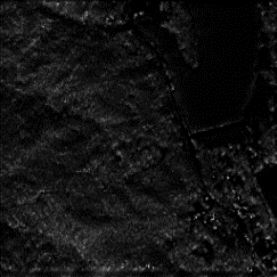}\vspace{6pt}
			\includegraphics[width=0.7in,height=0.7in]{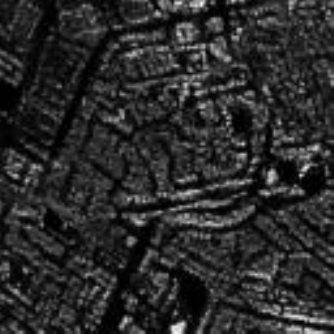}\vspace{4pt}
			\includegraphics[width=0.7in,height=0.7in]{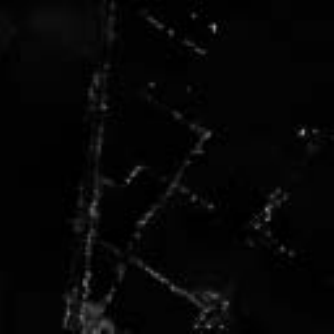}
			\centering{(d)}
		\end{minipage}
	}
	\caption{Example translation images with UAVSAR (test samples). Images in each row from left to right are the \textbf{(a) real SAR image} and its \textbf{(b) translated optical image}, the \textbf{(c) real optical image} and its \textbf{(d) translated SAR image}. The first two rows are chosen from \textbf{6m UAVSAR}, and the last two are from \textbf{10m UAVSAR}.}
	\label{fig:figure10} 
\end{figure}

The UAVSAR images are resampled to resolution of 6m and 10m and then used to train the proposed network. An example of translated SAR and optical images are show in \autoref{fig:figure10} where a good visual quality is achieved.

\begin{figure}[!htb]
	\scriptsize
	\centering
	\subfigure{
		\begin{minipage}[b]{0.12\linewidth}
			\includegraphics[width=0.7in,height=0.7in]{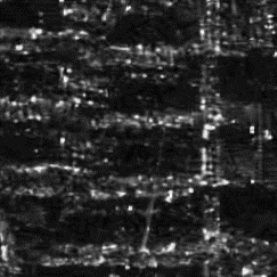}\vspace{4pt}
			\includegraphics[width=0.7in,height=0.7in]{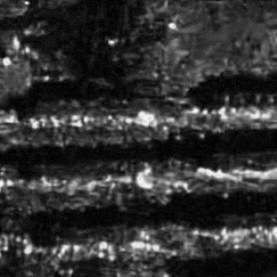}\vspace{4pt}
			\includegraphics[width=0.7in,height=0.7in]{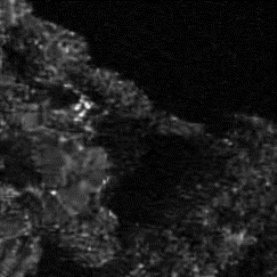}\vspace{4pt}
			\includegraphics[width=0.7in,height=0.7in]{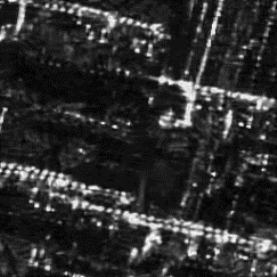}
			\centering{(a)}
		\end{minipage}
	}
	\subfigure{
		\begin{minipage}[b]{0.12\linewidth}
			\includegraphics[width=0.7in,height=0.7in]{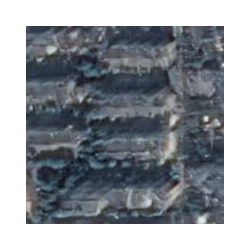}\vspace{4pt}
			\includegraphics[width=0.7in,height=0.7in]{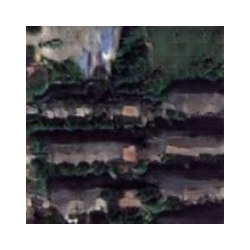}\vspace{4pt}
			\includegraphics[width=0.7in,height=0.7in]{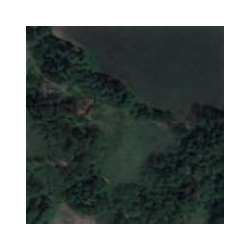}\vspace{4pt}
			\includegraphics[width=0.7in,height=0.7in]{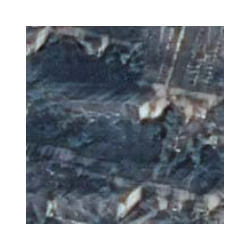}
			\centering{(b)}
		\end{minipage}
	}
	\subfigure{
		\begin{minipage}[b]{0.12\linewidth}
			\includegraphics[width=0.7in,height=0.7in]{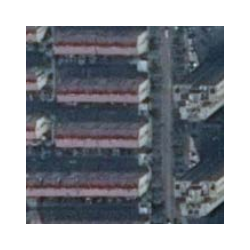}\vspace{4pt}
			\includegraphics[width=0.7in,height=0.7in]{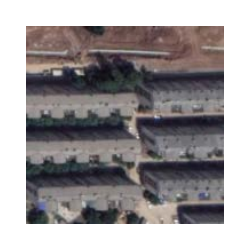}\vspace{4pt}
			\includegraphics[width=0.7in,height=0.7in]{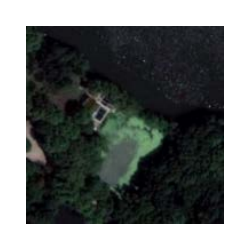}\vspace{4pt}
			\includegraphics[width=0.7in,height=0.7in]{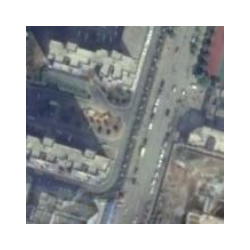}
			\centering{(c)}
		\end{minipage}
	}
	\subfigure{
		\begin{minipage}[b]{0.12\linewidth}
			\includegraphics[width=0.7in,height=0.7in]{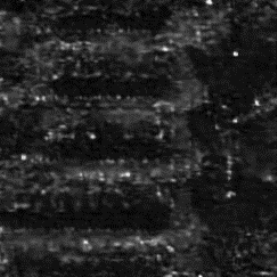}\vspace{4pt}
			\includegraphics[width=0.7in,height=0.7in]{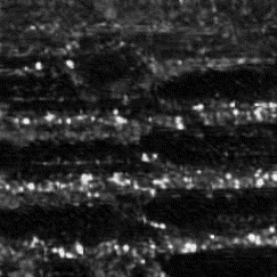}\vspace{4pt}
			\includegraphics[width=0.7in,height=0.7in]{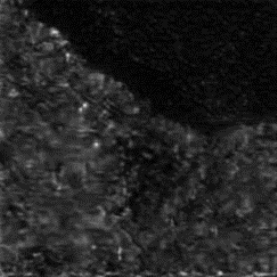}\vspace{4pt}
			\includegraphics[width=0.7in,height=0.7in]{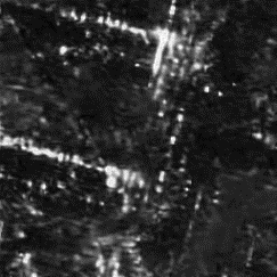}
			\centering{(d)}
		\end{minipage}
	}
	\caption{Example translation images with GF3 data. Images in each row from left to right are the \textbf{(a) real SAR image} and its \textbf{(b) translated optical image}, the \textbf{(c) real optical image} and its \textbf{(d) translated SAR image}.}
	\label{fig:figure11} 
\end{figure}

\autoref{fig:figure11} shows examples of high-resolution GF3 images translated by the proposed networks. The first row is a training sample and the rest three rows are test samples. It is found that earth surfaces like waters and vegetation can be easily reconstructed in both training and test cases. Low-rise buildings can be rebuilt into cubes, but their edges are not well-aligned. If the buildings are too close, the open space between them is difficult to distinguish. For high-rise buildings, the ones shown in the training samples appear to be reasonably realistic. However, the ones in the test sample in the bottom row appear to be smeared. It seems like that the network got confused by the viewing angles. Apparently, for tall 3D terrain objects, both SAR and optical images are very sensitive to the view angles. Without incorporation of the projection mechanism, the proposed network is not able to generalize in this dimension. 

\begin{figure}
	\centering 
	\includegraphics[width=3.5in]{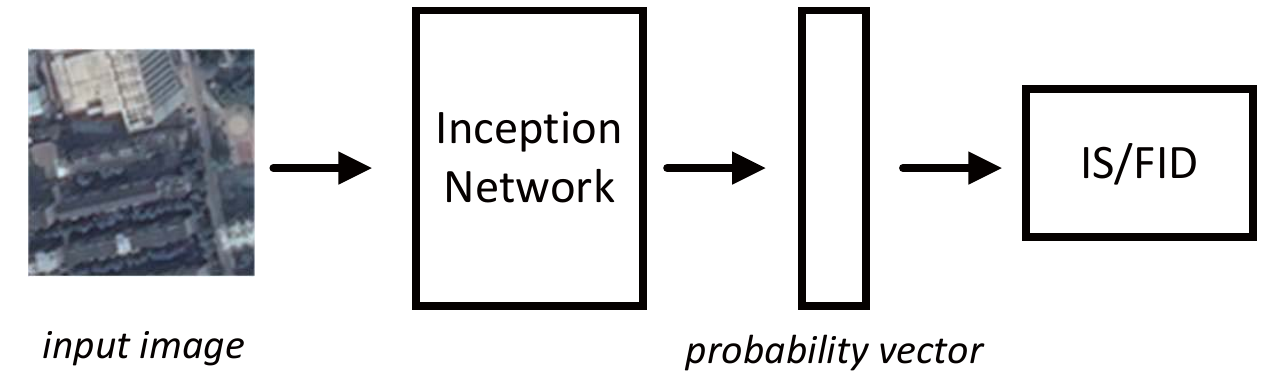} 
	\caption{The conceptual process of calculating IS/FID.}
	\label{fig:figure12} 
\end{figure}

It is necessary to quantitatively measure the difference between the translated images and true ones. Traditional methods, such as L1, Peak Signal-to-Noise Ratio (PSNR) and Structural Similarity (SSIM) (\cite{wang2004image}), could be used to measure the similarity between two images. However, these methods still compare the similarity in terms of pixel values but rather than in the sense of perceptual similarity. Inception score (IS) (\cite{salimans2016improved}) and Fréchet inception distance (FID) (\cite{heusel2017gans}) are usually used to quantitatively evaluate the quality and variety of images generated by GANs. Both of them encode the input image to a feature vector by using the inception network, shown in \autoref{fig:figure12}, which functions as the human visual perception. If the two images are identical, their encoded feature vectors should be the same.

Different from IS, FID uses the statistics of real world samples and compares them to the statistics of synthetic samples. FID between the Gaussian distribution with mean and covariance $({{m}_{1}},{{C}_{1}})$ and the Gaussian distribution with $({{m}_{2}},{{C}_{2}})$ is defined as $||{{m}_{1}}-{{m}_{2}}||_{2}^{2}+Tr({{C}_{1}}+{{C}_{2}}-2{{({{C}_{1}}{{C}_{2}})}^{1/2}})$. Lower FID is better, corresponding to more similar real and generated samples. It is found that the FIDs of the generated 0.51m optical and SAR images are respectively 154.7532 and 53.0067. The values are quite large, which indicates that the model performs badly. However, the reconstructed images are very good from the perspective of human eye. The buildings, farmlands, green areas, etc. in each image are generally well classified. The textures are also allocated. Nevertheless, an exception exists. The textures of buildings vary widely and are hard to match one-to-one with ground truths. For large-scale urban scenes, high-frequency parts such as noise and details in ground truths are difficult to learn because those in each sample differ greatly. Our main purpose is to reconstruct their main contours.

The number of samples to calculate the Gaussian statistics (mean and covariance) should be greater than the dimension of the last coding layer, here 2048 for the inception pool 3 layer (\cite{heusel2017gans}). Otherwise the covariance is not full rank, which will result in complex numbers and nans by calculating the square root. Here, we use 2048 pairs of test samples to calculate the FID to estimate the capability of the generators. 

\autoref{table3} lists the tested FID values for the different resolution datasets mentioned above. Randomly select 2048 pairs of samples from each dataset and calculate the corresponding FID value. Repeat three times and use the mean as the ability of our model to train this kind of dataset. It indicates that 6m data performs better than 0.51m and 10m data. The 10m results are not ideal due to their small features hard to extract.
\begin{table}
	\scriptsize
	\renewcommand\arraystretch{1.5}
	\setlength{\abovecaptionskip}{0pt}
	\setlength{\belowcaptionskip}{10pt}%设置标题与表格的距离
	\caption{FIDs of different datasets.}
	\label{table3}
%	\centering
	\begin{tabular}{*{5}{m{0.17\textwidth}<{\centering}}}
%	\begin{tabular}{ccccc}
		\hline
			& 0.51m single-pol GF3 & 6m single-pol UAVSAR & 10m single-pol UAVSAR & 6m full-pol UAVSAR \\
		\hline
		Optical & 154.8 & 106.4 & 138.4 & 85.6 \\
		\hline
		SAR & 53.0 & 56.0 & 64.7 & 52.8 \\
		\hline
	\end{tabular}
\end{table}

\begin{table}
	\scriptsize
	\renewcommand\arraystretch{1.5}
	\setlength{\abovecaptionskip}{0pt}
	\setlength{\belowcaptionskip}{10pt}%设置标题与表格的距离
	\caption{Decreasing FID with increasing the number of samples (for the case of 6m full-pol UAVSAR in \autoref{table3}).}
	\label{table4} 
%	\centering %
	\begin{tabular}{*{12}{m{0.27in}<{\centering}}}
		\hline
		Num & 500 & 1000 & 2048 & 3000 & 4000 & 5000 & 6000 & 7000 & 8000 & 9000 & 10000 \\
		\hline
		Opt. & 125.0 & 102.9 & 85.6 & 81.2 & 77.9 & 75.9 & 74.8 & 74.1 & 73.4 & 72.7 & 72.1 \\
		\hline
		SAR & 86.9 & 68.8 & 52.8 & 49.4 & 46.8 & 45.9 & 44.5 & 43.2 & 42.5 & 42.0 & 41.9 \\
		\hline
	\end{tabular}
\end{table}

Note that FID could be further reduced by increasing the number of samples. As shown in \autoref{table4} for the case of 6m full-pol UAVSAR, its FID could be reduced to 72 for optical and 42 for SAR if given 10000 samples.

\subsubsection{Polarization}
The SAR data used in this study so far is all single-pol, i.e. HH or VV single-pol. Full-pol data contains rich polarimetric information. It is worth to investigate how the performance might improve if full-pol SAR is used. For simplicity, the pauli color-coded image is used as an proxy of full-pol SAR data.

The basic form of polarimetric SAR data is the Sinclair scattering matrix (\cite{lee2009polarimetric}) with horizontal and vertical polarisations. It can be expressed as a $2\times2$ matrix containing four components $S_{HH}$, $S_{HV}$, $S_{VH}$ and $S_{VV}$, where $H$, $V$ respectively denotes the horizontal and vertical polarisations. ${S_{HH}}$ and ${S_{VV}}$ are co-polarized components; ${S_{HV}}$ and ${S_{VH}}$ are cross-polarized components. Different polarimetric channels contain partial electromagnetic information. Some targets may be imaged more clearly in the cross-polarized channels than that in the co-polarized channels, and vice versa (\cite{jin2013polarimetric}).

Then we convert the full polarimetric information into pseudo-color coded images via Pauli decomposition. Pauli decomposition is to decompose the scattering matrix $S$ into different scattering components, i.e. $a$ is the single-bounce surface scattering intensity; $b$ is the dihedral scattering intensity with incidence angle $0{}^\circ$; $c$ is the volumetric scattering.
\begin{flalign}
\label{equ:equation7}
& a=\frac{S_{HH}+S_{VV}}{\sqrt{2}}, b=\frac{S_{HH}-S_{VV}}{\sqrt{2}}, c=\frac{S_{HV}+S_{VH}}{\sqrt{2}}, d=j \frac{S_{HV}-S_{VH}}{\sqrt{2}} &
\end{flalign}

The Pauli image is a pseudo-color image coded using the intensities of these three components, i.e.
\begin{flalign}
\label{equ:equation8}
& \mathrm{I}=\left[\left|S_{HH}-S_{VV}\right|^{2}, 4\left|S_{H V}\right|^{2},\left|S_{HH}+S_{VV}\right|^{2}\right]^{T} / 2 &
\end{flalign}

\begin{figure}[!htb]
	\scriptsize
	\centering
	\subfigure{
		\begin{minipage}[b]{0.12\linewidth}
			\includegraphics[width=0.7in,height=0.7in]{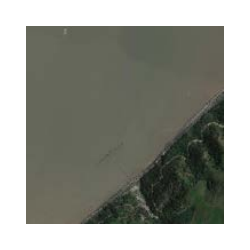}\vspace{4pt}
			\includegraphics[width=0.7in,height=0.7in]{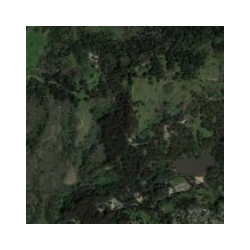}\vspace{4pt}
			\includegraphics[width=0.7in,height=0.7in]{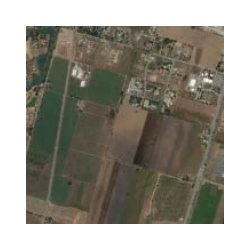}\vspace{4pt}
			\includegraphics[width=0.7in,height=0.7in]{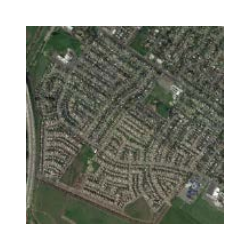}
			\centering{(a)}
		\end{minipage}
	}
	\subfigure{
		\begin{minipage}[b]{0.12\linewidth}
			\includegraphics[width=0.7in,height=0.7in]{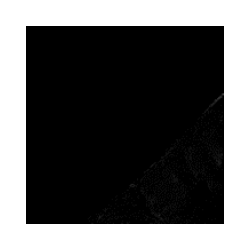}\vspace{4pt}
			\includegraphics[width=0.7in,height=0.7in]{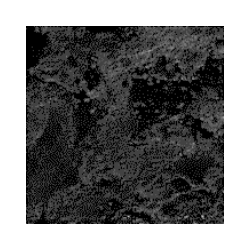}\vspace{4pt}
			\includegraphics[width=0.7in,height=0.7in]{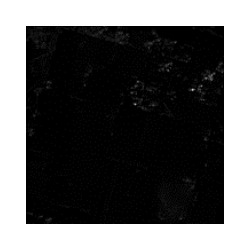}\vspace{4pt}
			\includegraphics[width=0.7in,height=0.7in]{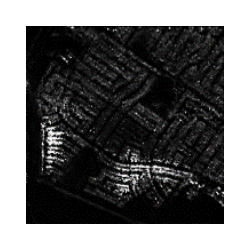}
			\centering{(b)}
		\end{minipage}
	}
	\subfigure{
		\begin{minipage}[b]{0.12\linewidth}
			\includegraphics[width=0.7in,height=0.7in]{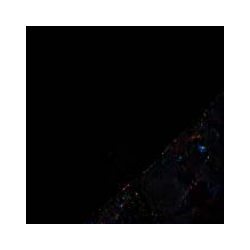}\vspace{4pt}
			\includegraphics[width=0.7in,height=0.7in]{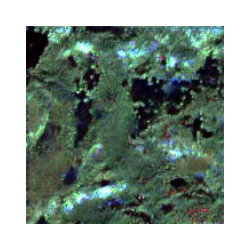}\vspace{4pt}
			\includegraphics[width=0.7in,height=0.7in]{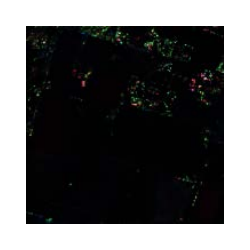}\vspace{4pt}
			\includegraphics[width=0.7in,height=0.7in]{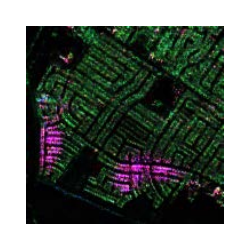}
			\centering{(c)}
		\end{minipage}
	}
	\subfigure{
		\begin{minipage}[b]{0.12\linewidth}
			\includegraphics[width=0.7in,height=0.7in]{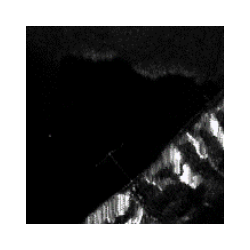}\vspace{4pt}
			\includegraphics[width=0.7in,height=0.7in]{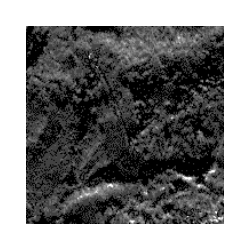}\vspace{4pt}
			\includegraphics[width=0.7in,height=0.7in]{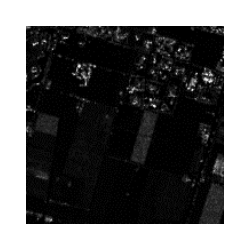}\vspace{4pt}
			\includegraphics[width=0.7in,height=0.7in]{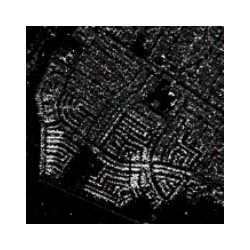}
			\centering{(d)}
		\end{minipage}
	}
	\subfigure{
		\begin{minipage}[b]{0.12\linewidth}
			\includegraphics[width=0.7in,height=0.7in]{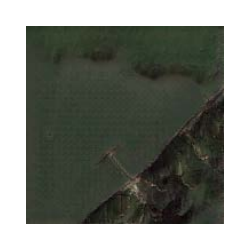}\vspace{4pt}
			\includegraphics[width=0.7in,height=0.7in]{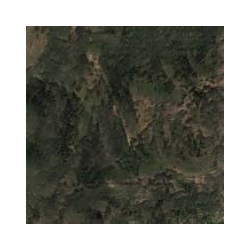}\vspace{4pt}
			\includegraphics[width=0.7in,height=0.7in]{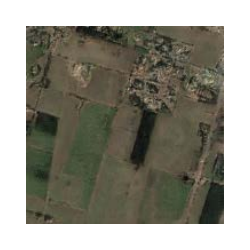}\vspace{4pt}
			\includegraphics[width=0.7in,height=0.7in]{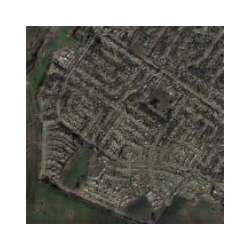}
			\centering{(e)}
		\end{minipage}
	}
	\subfigure{
		\begin{minipage}[b]{0.12\linewidth}
			\includegraphics[width=0.7in,height=0.7in]{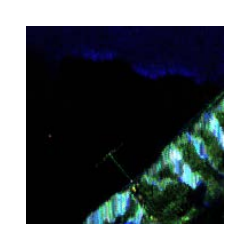}\vspace{4pt}
			\includegraphics[width=0.7in,height=0.7in]{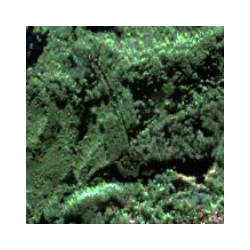}\vspace{4pt}
			\includegraphics[width=0.7in,height=0.7in]{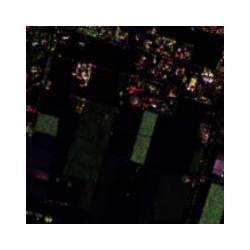}\vspace{4pt}
			\includegraphics[width=0.7in,height=0.7in]{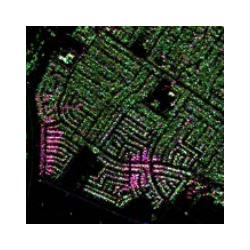}
			\centering{(f)}
		\end{minipage}
	}
	\subfigure{
		\begin{minipage}[b]{0.12\linewidth}
			\includegraphics[width=0.7in,height=0.7in]{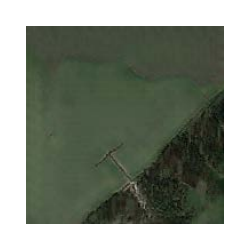}\vspace{4pt}
			\includegraphics[width=0.7in,height=0.7in]{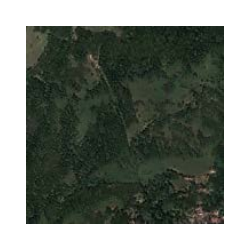}\vspace{4pt}
			\includegraphics[width=0.7in,height=0.7in]{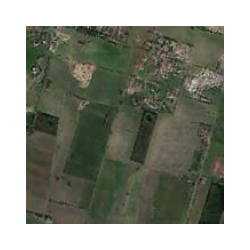}\vspace{4pt}
			\includegraphics[width=0.7in,height=0.7in]{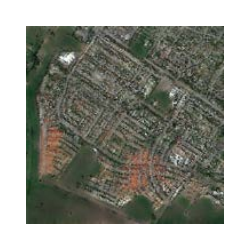}
			\centering{(g)}
		\end{minipage}
	}
	\caption{Images listed above in each row are \textbf{(a) the optical ground truth} and its \textbf{(b) translated single-polarized SAR image} and \textbf{(c) translated full-polarized SAR image}, the \textbf{(d) single-polarized SAR ground truth} and its \textbf{(e) translated optical image}, the \textbf{(f) full-polarized SAR ground truth} and its \textbf{(g) translated optical image} in order. Each row lists a kind of earth surfaces: waters, vegetation, farmlands and buildings.}
	\label{fig:figure13} 
\end{figure}

Finally, we carry out an experiment to train our model with full-pol images and single-pol images in the same region respectively, and compare the translation performance. As shown in \autoref{table3}, it indicates that 6m full-polarized data has the best performance, especially the reconstructed optical images are much better than those from 6m single-pol data. Four examples of different kinds of earth surfaces, waters, vegetation, farmlands and buildings are shown \autoref{fig:figure13}, respectively. Apparently, the optical images translated from full-pol SAR images are more vivid and realistic than those from single-pol images.

\begin{figure}[!htb]
	\scriptsize
	\centering
	\subfigure{
		\begin{minipage}[b]{0.12\linewidth}
			\includegraphics[width=0.7in,height=0.7in]{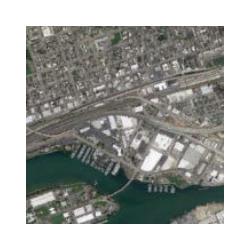}\vspace{4pt}
			\includegraphics[width=0.7in,height=0.7in]{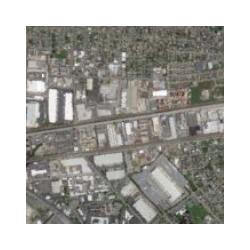}\vspace{4pt}
			\includegraphics[width=0.7in,height=0.7in]{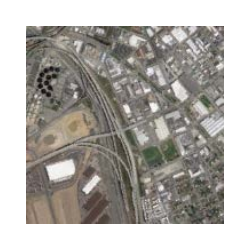}
			\centering{(a)}
		\end{minipage}
	}
	\subfigure{
		\begin{minipage}[b]{0.12\linewidth}
			\includegraphics[width=0.7in,height=0.7in]{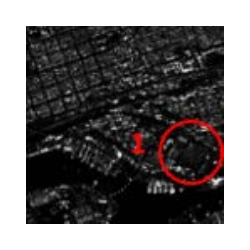}\vspace{4pt}
			\includegraphics[width=0.7in,height=0.7in]{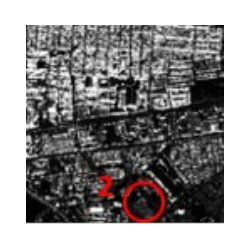}\vspace{4pt}
			\includegraphics[width=0.7in,height=0.7in]{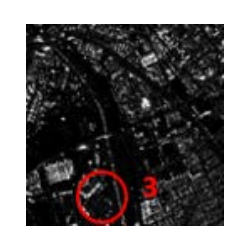}
			\centering{(b)}
		\end{minipage}
	}
	\subfigure{
		\begin{minipage}[b]{0.12\linewidth}
			\includegraphics[width=0.7in,height=0.7in]{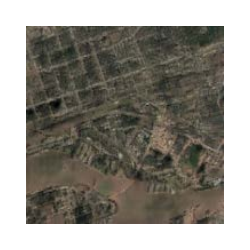}\vspace{4pt}
			\includegraphics[width=0.7in,height=0.7in]{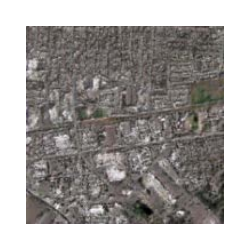}\vspace{4pt}
			\includegraphics[width=0.7in,height=0.7in]{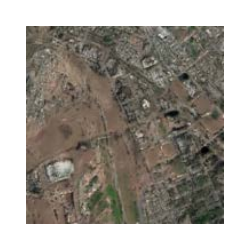}
			\centering{(c)}
		\end{minipage}
	}
	\subfigure{
		\begin{minipage}[b]{0.12\linewidth}
			\includegraphics[width=0.7in,height=0.7in]{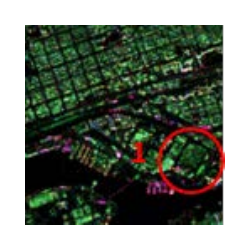}\vspace{4pt}
			\includegraphics[width=0.7in,height=0.7in]{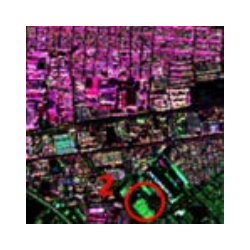}\vspace{4pt}
			\includegraphics[width=0.7in,height=0.7in]{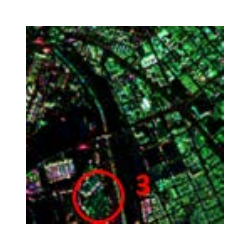}
			\centering{(d)}
		\end{minipage}
	}
	\subfigure{
		\begin{minipage}[b]{0.12\linewidth}
			\includegraphics[width=0.7in,height=0.7in]{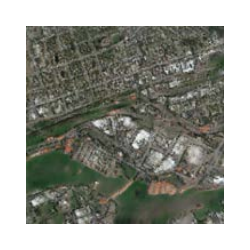}\vspace{4pt}
			\includegraphics[width=0.7in,height=0.7in]{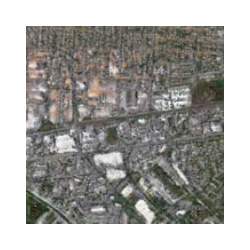}\vspace{4pt}
			\includegraphics[width=0.7in,height=0.7in]{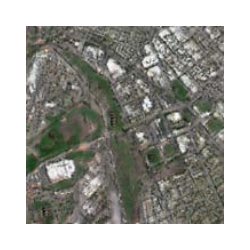}
			\centering{(e)}
		\end{minipage}
	}
	\caption{Images in each row from left to right are the \textbf{(a) real optical image}, the real \textbf{(b)single-pol SAR image} and its \textbf{(c) translated optical image}, the real \textbf{(d) full-pol SAR image} and its \textbf{(e) translated optical image}.}
	\label{fig:figure14} 
\end{figure}

\autoref{fig:figure14} further investigates into few interesting cases. In each case, one building in single-pol image and the corresponding full-pol image is marked correspondingly in each row. Note that these buildings are all easily observable in the full-pol image but not in the single-pol image. This is mainly because of the imbalance of scattering power distribution over different polarization channels. The optical image translated by the full-pol SAR image appear to be much more realistic and closer to true image. Apparently, it is benefited from the additional rich information conveyed in the full-pol SAR image.

\subsection{Comparison with existing translation networks}
To evaluate the performance of the proposed method in the context of existing image translation approaches, here, we compare it with CycleGAN (\cite{zhu2017unpaired}) and Pix2Pix (\cite{isola2017image}) using the 0.51m GF3 images. Note that the CycleGAN implemented here shares the same network structure with Pix2Pix, but is trained with the cyclic loop strategy. In order to ensure the fairness of comparison, the discriminators and the receptive fields of the generators are the same. The number of the generators’ layers and that of the total trainable parameters are the same. The parameters of the network are randomly initialized. To remove the slight dependence of training result on the initialization, each network is repeatedly trained for 3 times with the same data and then the best result is chosen.

\begin{figure}[!htb]
	\scriptsize
	\centering
	\subfigure{
		\begin{minipage}[b]{0.12\linewidth}
			\includegraphics[width=0.7in,height=0.7in]{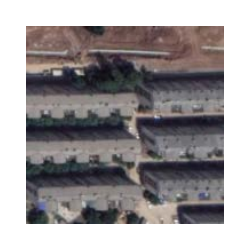}\vspace{4pt}
			\includegraphics[width=0.7in,height=0.7in]{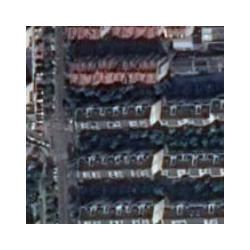}\vspace{4pt}
			\includegraphics[width=0.7in,height=0.7in]{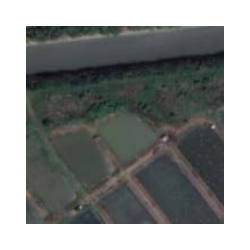}\vspace{4pt}
			\includegraphics[width=0.7in,height=0.7in]{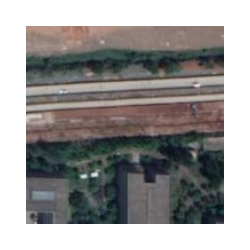}
			\centering{(a)}
		\end{minipage}
	}
	\subfigure{
		\begin{minipage}[b]{0.12\linewidth}
			\includegraphics[width=0.7in,height=0.7in]{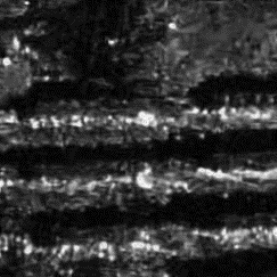}\vspace{4pt}
			\includegraphics[width=0.7in,height=0.7in]{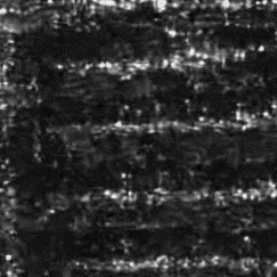}\vspace{4pt}
			\includegraphics[width=0.7in,height=0.7in]{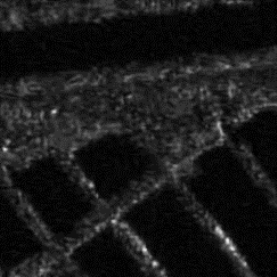}\vspace{4pt}
			\includegraphics[width=0.7in,height=0.7in]{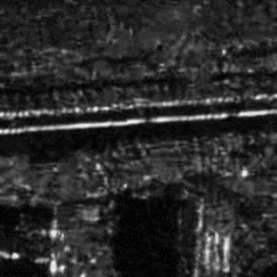}
			\centering{(b)}
		\end{minipage}
	}
	\subfigure{
		\begin{minipage}[b]{0.12\linewidth}
			\includegraphics[width=0.7in,height=0.7in]{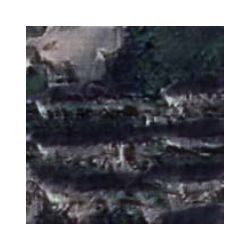}\vspace{4pt}
			\includegraphics[width=0.7in,height=0.7in]{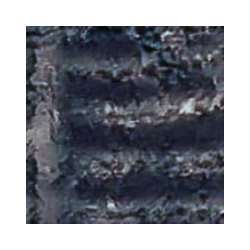}\vspace{4pt}
			\includegraphics[width=0.7in,height=0.7in]{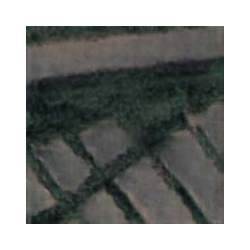}\vspace{4pt}
			\includegraphics[width=0.7in,height=0.7in]{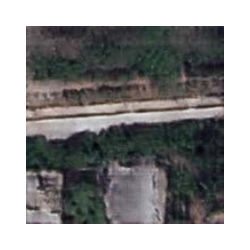}
			\centering{(c)}
		\end{minipage}
	}
	\subfigure{
		\begin{minipage}[b]{0.12\linewidth}
			\includegraphics[width=0.7in,height=0.7in]{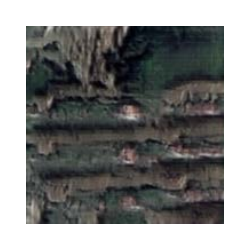}\vspace{4pt}
			\includegraphics[width=0.7in,height=0.7in]{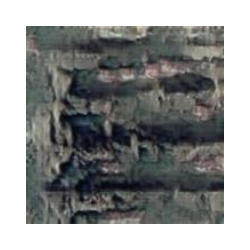}\vspace{4pt}
			\includegraphics[width=0.7in,height=0.7in]{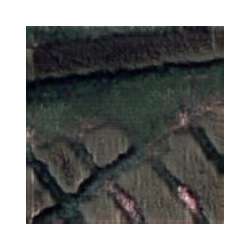}\vspace{4pt}
			\includegraphics[width=0.7in,height=0.7in]{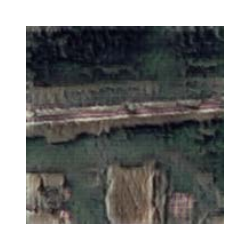}
			\centering{(d)}
		\end{minipage}
	}
	\subfigure{
		\begin{minipage}[b]{0.12\linewidth}
			\includegraphics[width=0.7in,height=0.7in]{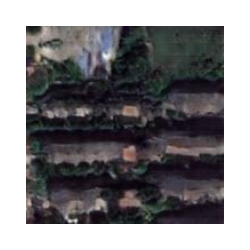}\vspace{4pt}
			\includegraphics[width=0.7in,height=0.7in]{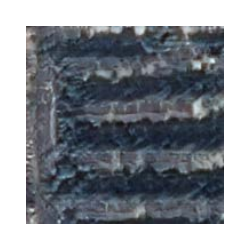}\vspace{4pt}
			\includegraphics[width=0.7in,height=0.7in]{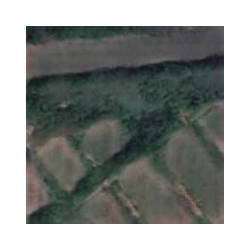}\vspace{4pt}
			\includegraphics[width=0.7in,height=0.7in]{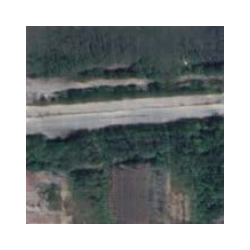}
			\centering{(e)}
		\end{minipage}
	}
	\caption{Comparison of SAR-Optical translation by different methods. Images in each row from left to right are the \textbf{(a) real optical image}, the \textbf{(b)input SAR image}, its \textbf{(c) translated optical image by CycleGAN}, the \textbf{(d) translated optical image by Pix2Pix} and the \textbf{(e) translated optical image by CRAN}. Each row lists a kind of earth surfaces: buildings, buildings, farmlands and roads.}
	\label{fig:figure15} 
\end{figure}

In \autoref{fig:figure15}, four representative pairs of different earth surfaces are selected. It can be found that in the first two rows, the buildings reconstructed by our method are more natural. In the third and fourth rows, waters, roads and vegetation reconstructed by CycleGAN are similar to the proposed method.

In \autoref{table5}, the three metrics PSNR, SSIM and FID are all employed. PSNR is the inverse of the sum of pixel difference between the reference image and the measured image and SSIM is a metric to evaluate image similarity from brightness, contrast and structure aspects. The larger the value of PSNR or SSIM, the more similar the two images are. FID is used to measure the distance from generated samples to real world samples, and the smaller FID, the more analogous the two datasets are. On the 0.51m  and 6m single-pol datasets, our proposed method outperforms CycleGAN and Pix2Pix in the other three indicators, especially the FID score improved greatly. Note that for the cases of 10m UAVSAR and 6m full-pol UAVSAR datasets, it is generally better than the other two methods.

\begin{table}
%	\small
	\scriptsize
	\renewcommand\arraystretch{1.5}
	\setlength{\abovecaptionskip}{0pt}
	\setlength{\belowcaptionskip}{10pt}
	\caption{Result comparisons of different methods with different datasets using different evaluation methods.}
	\label{table5} 
%	\centering %
	\begin{tabular}{cccccccc}
		\hline
		Dataset & Method & \multicolumn{2}{c}{SSIM} & \multicolumn{2}{c}{PSNR} & \multicolumn{2}{c}{FID} \\
		\hline
		
		\multirow{3}*{0.51m single-pol GF3} & CycleGAN & 0.2535 & 0.2656 & 15.7171 & 14.9675 & 62.1420 & 185.3181 \\
		
		& Pix2Pix & 0.2194 & 0.2317 & 15.4978 & 14.4686 & 77.6901 & 212.5304 \\
		
		& CRAN & \textbf{0.2595} & \textbf{0.2799} & \textbf{15.9172} & \textbf{15.5820} & \textbf{53.0067} & \textbf{154.7532} \\
		
		\hline
		
		\multirow{3}*{6m single-pol UAVSAR} & CycleGAN & 0.3585 & 0.3005 & 19.5424 & 16.1030 & 50.5496 & 132.1710 \\
		
		& Pix2Pix & 0.3407 & 0.3081 & 19.6044 & 15.7463 & \textbf{48.5541} & \textbf{99.7782} \\
		
		& CRAN & \textbf{0.3640} & \textbf{0.3092} & \textbf{20.2907} & \textbf{16.1323} & 56.0201 & 106.3988 \\
		
		\hline
		
		\multirow{3}*{10m single-pol UAVSAR} & CycleGAN & 0.2879 & 0.2973 & \textbf{18.5911} & 16.2957 & \textbf{53.2890} & \textbf{113.288} \\
		
		& Pix2Pix & \textbf{0.2917} & 0.3072 & 18.3707 & 16.0357 & 63.5519 & 146.7449 \\
		
		& CRAN & 0.2819 & \textbf{0.3346} & 18.3092 & \textbf{16.4238} & 64.7359 & 138.3651 \\
		
		\hline
		
		\multirow{3}*{6m full-pol UAVSAR} & CycleGAN & 0.3418 & 0.3254 & 18.3431 & 16.0414 & \textbf{46.0073} & 95.69 \\
		
		& Pix2Pix & 0.3716 & \textbf{0.3308} & \textbf{19.5295} & 16.0421 & 65.1980 & 94.9724 \\
		
		& CRAN	& \textbf{0.3768} & 0.3109 & 19.2188 & \textbf{16.1489} & 52.7645 & \textbf{85.5704} \\
		
		\hline
	\end{tabular}
\end{table}

Note that the selected quantitative metrics can only be used as a general reference of image generation performance. In some cases, it may not faithfully and precisely reflect the actual visual appearance of the generated image. Two cases are given in \autoref{fig:figure16} below. The metrics of these cases are also provided above the corresponding images. As we can see from the result that some cases appears better visually but measured with slightly lower metrics. For example, for the images $a$, $b$ and $c$ selected from 6m full-pol UAVSAR dataset, the optical image $b$ translated by CRAN appears to be better than the image $c$ generated by CycleGAN, but the values of SSIM and PSNR are actually smaller than the latter; for the images $d$, $e$, $f$ taken from 10m single-pol UAVSAR data, similarly, $e$ generated by CRAN appears visually better than $f$, but the latter has a larger PSNR value.
\begin{figure}[!htb]
	\scriptsize
	\centering
	\vspace{-0.05in} 
	\begin{tabular}{*{3}{m{0.8in}<{\centering}}}
		& \textbf{{\scriptsize SSIM:0.188}} & \textbf{{\scriptsize SSIM:0.190}} \\
		& \textbf{{\scriptsize PSNR:14.948}} & \textbf{{\scriptsize PSNR:15.599}} \\
	\end{tabular}
	\vspace{-0.05in}
	
	\subfigure{\includegraphics[width=0.8in,height=0.8in]{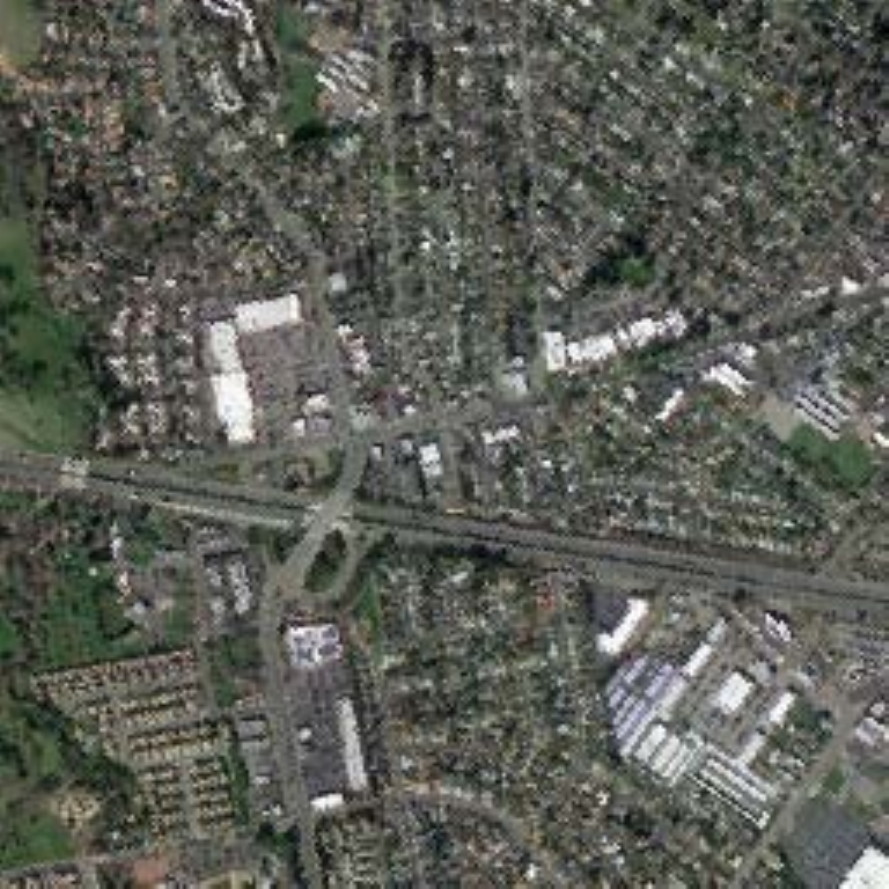}}
	\hspace{0.1in}
	\subfigure{\includegraphics[width=0.8in,height=0.8in]{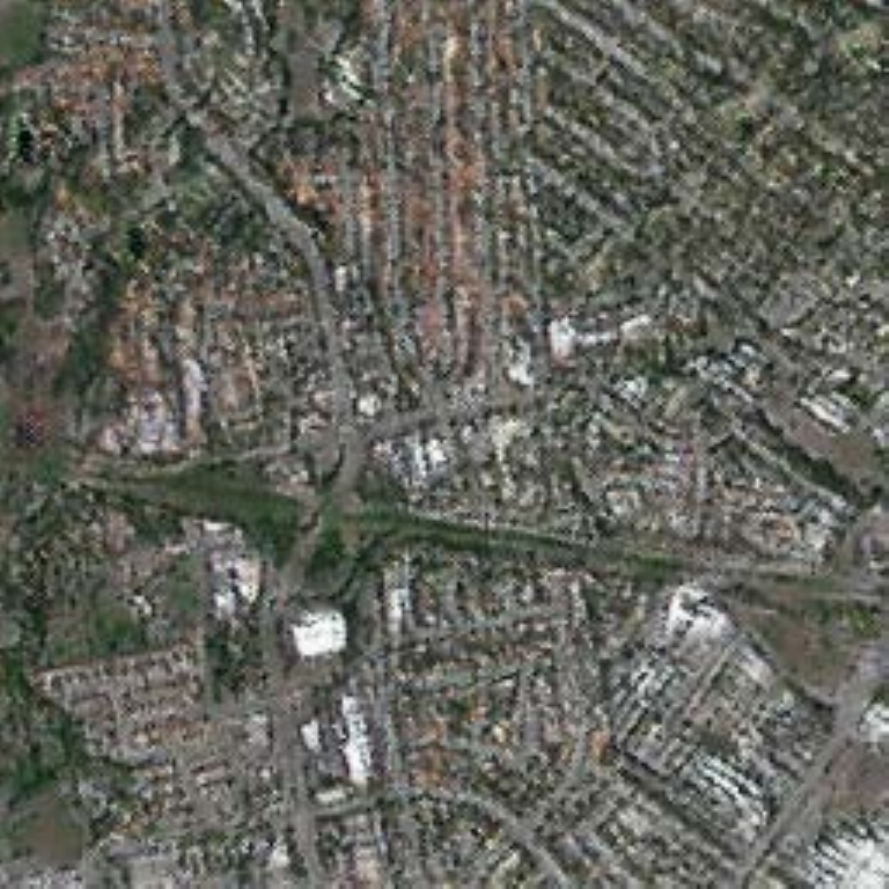}}
	\hspace{0.1in}
	\subfigure{\includegraphics[width=0.8in,height=0.8in]{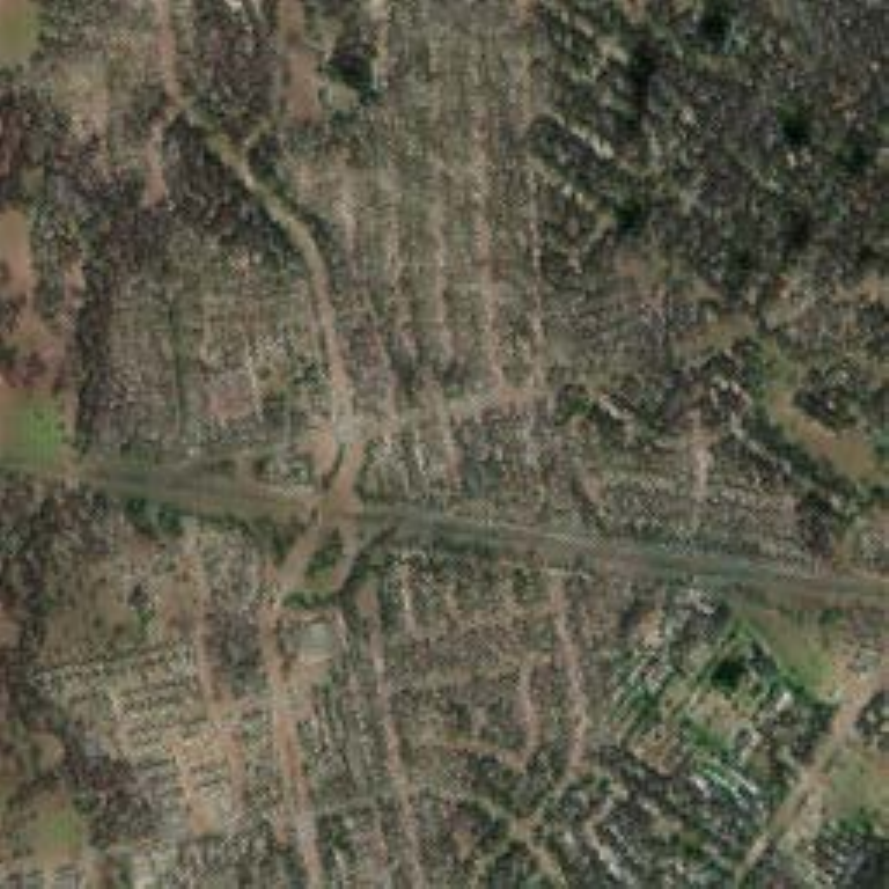}} \\
	\vspace{-0.05in}
	\begin{tabular}{*{3}{m{0.8in}<{\centering}}}
		(a) & (b) & (c) \\
	\end{tabular}
	\vspace{0.01in}
	
	\begin{tabular}{*{3}{m{0.8in}<{\centering}}}
		 & \textbf{{\scriptsize SSIM:0.308}} & \textbf{{\scriptsize SSIM:0.296}} \\
		 & \textbf{{\scriptsize PSNR:17.884}} & \textbf{{\scriptsize PSNR:18.577}} \\
	\end{tabular}
	\vspace{-0.05in}
	
	\subfigure{\includegraphics[width=0.8in,height=0.8in]{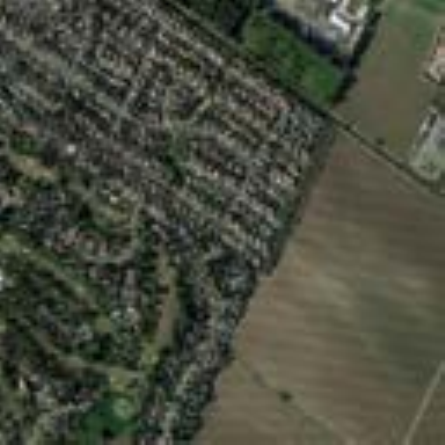}}
	\hspace{0.1in}
	\subfigure{\includegraphics[width=0.8in,height=0.8in]{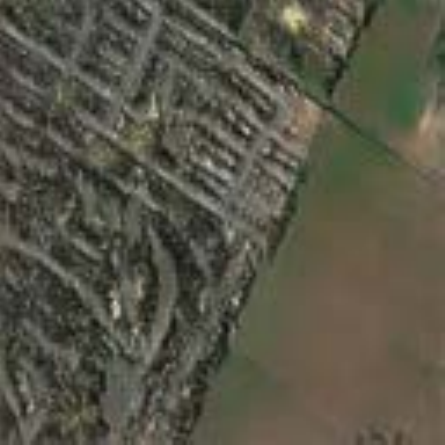}}
	\hspace{0.1in}
	\subfigure{\includegraphics[width=0.8in,height=0.8in]{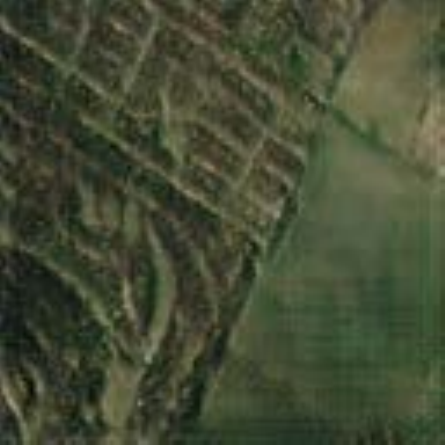}} \\
	\vspace{-0.05in}
	\begin{tabular}{*{3}{m{0.8in}<{\centering}}}
		(d) & (e) & (f) \\
	\end{tabular}
	\vspace{-0.05in}
	\caption{Comparison of SAR-Optical translation by different methods to verify the not exactly correctness of evaluation metrics.The first row is chosen from 6m full-pol UAVSAR dataset: \textbf{(a)Ground Truth}, \textbf{(b)Trans. Opt. by CRAN}, \textbf{(c)Trans. Opt. by CycleGAN}; the second row is chosen from 10m single-pol UAVSAR dataset: \textbf{(d)Ground Truth}, \textbf{(e)Trans. Opt. by CRAN}, \textbf{(f)Trans. Opt. by Pix2Pix}.}
	\label{fig:figure16} 
\end{figure}

\subsection{Generalization to different SAR platforms}
Generalization capability is critical to make the proposed method applicable in practical scenarios. One key aspect is generalization to different geographic scenes. From the cases presented in previous subsections, the test samples are acquired from different regions than the training samples, where the low FID has demonstrated that the proposed method can be generalized to different scenes. Another critical test is generalization to different SAR platforms, e.g. a model is trained with data from one SAR platform but used to translate SAR image from another platform.

\begin{figure}[!htb]
	\scriptsize
	\centering
	\subfigure{
		\begin{minipage}[b]{0.22\linewidth}
			\includegraphics[width=1.2in,height=1.2in]{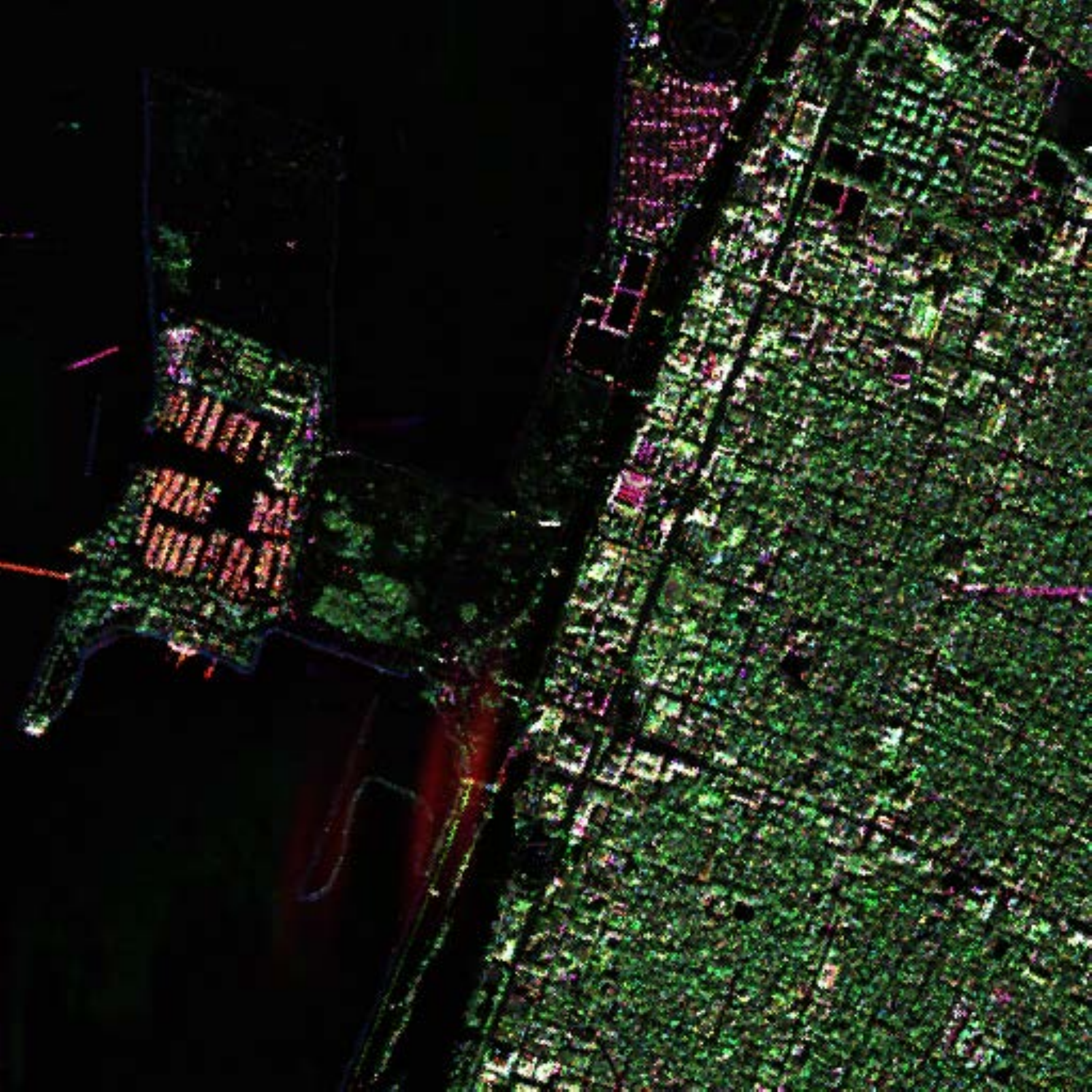}\vspace{4pt}
			\includegraphics[width=1.2in,height=1.2in]{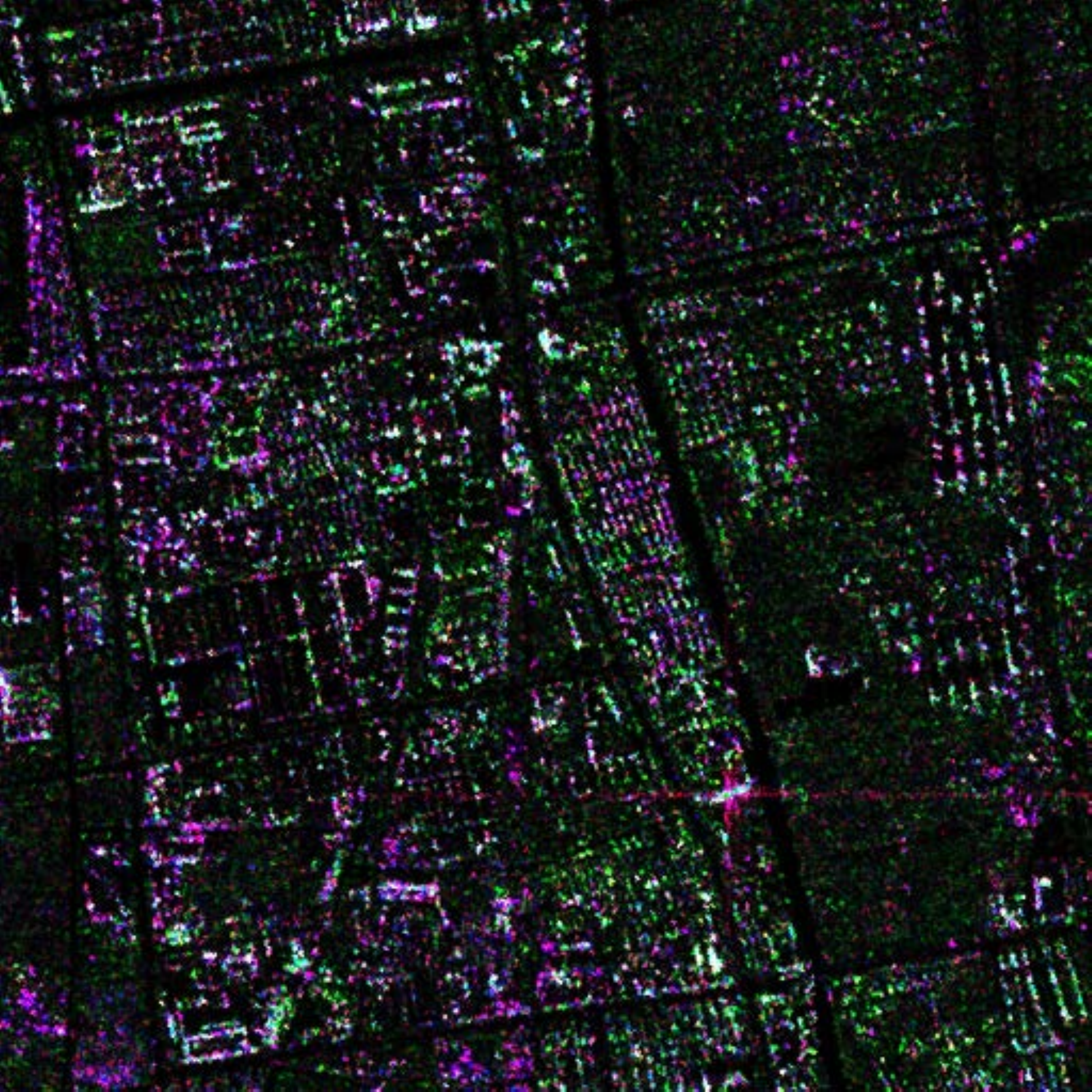}\vspace{4pt}
			\includegraphics[width=1.2in,height=1.2in]{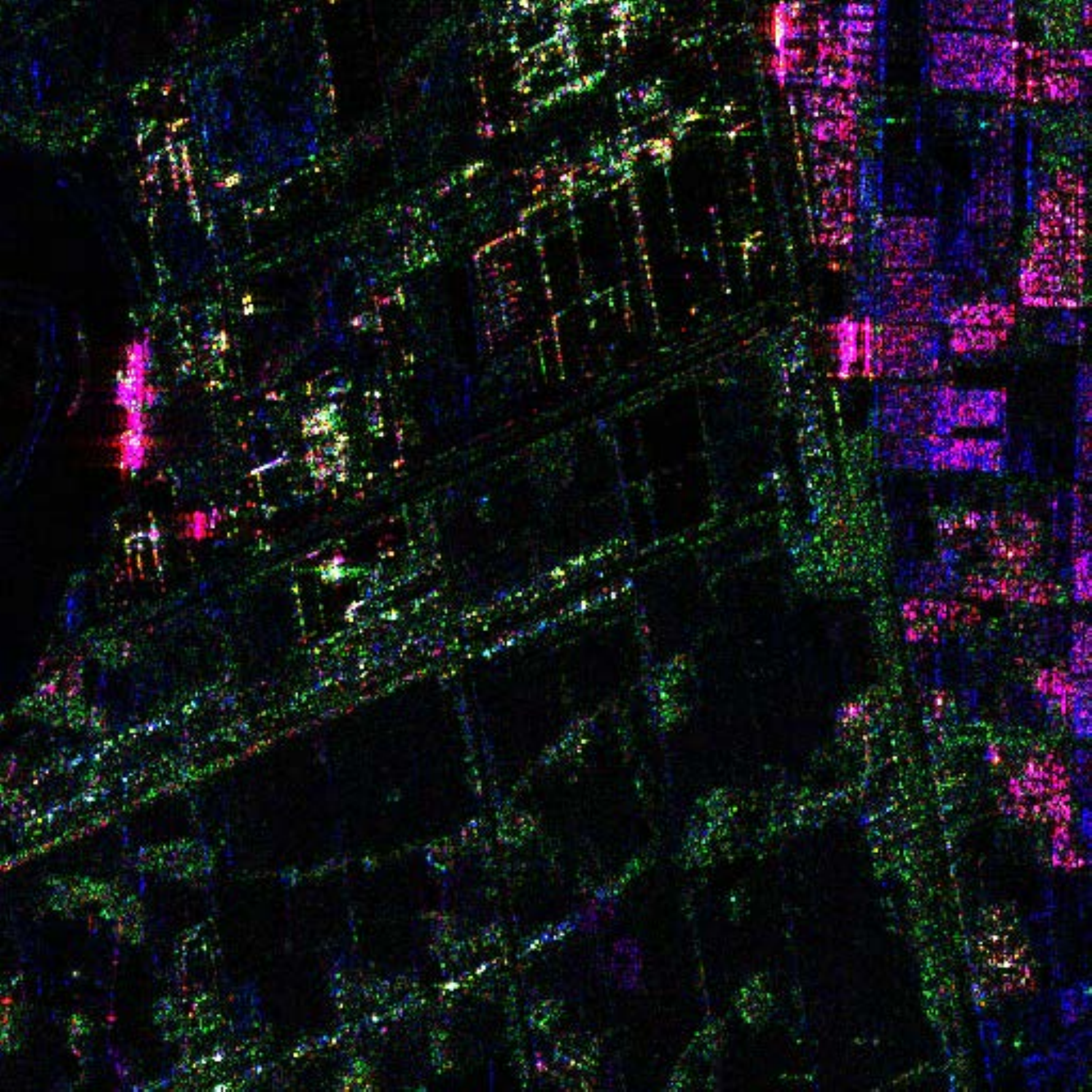}
			\centering{(a)}
		\end{minipage}
	}
	\subfigure{
		\begin{minipage}[b]{0.22\linewidth}
			\includegraphics[width=1.2in,height=1.2in]{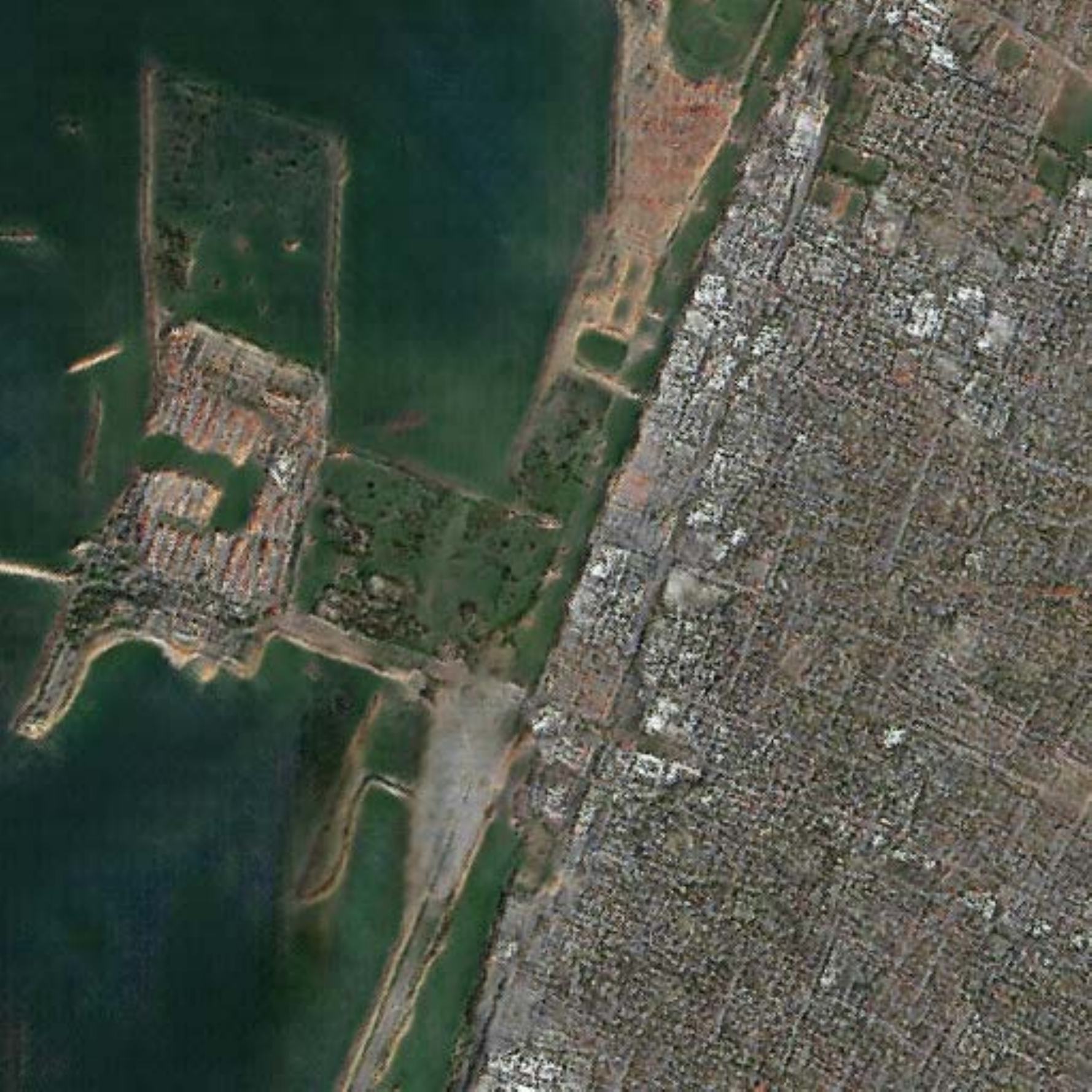}\vspace{4pt}
			\includegraphics[width=1.2in,height=1.2in]{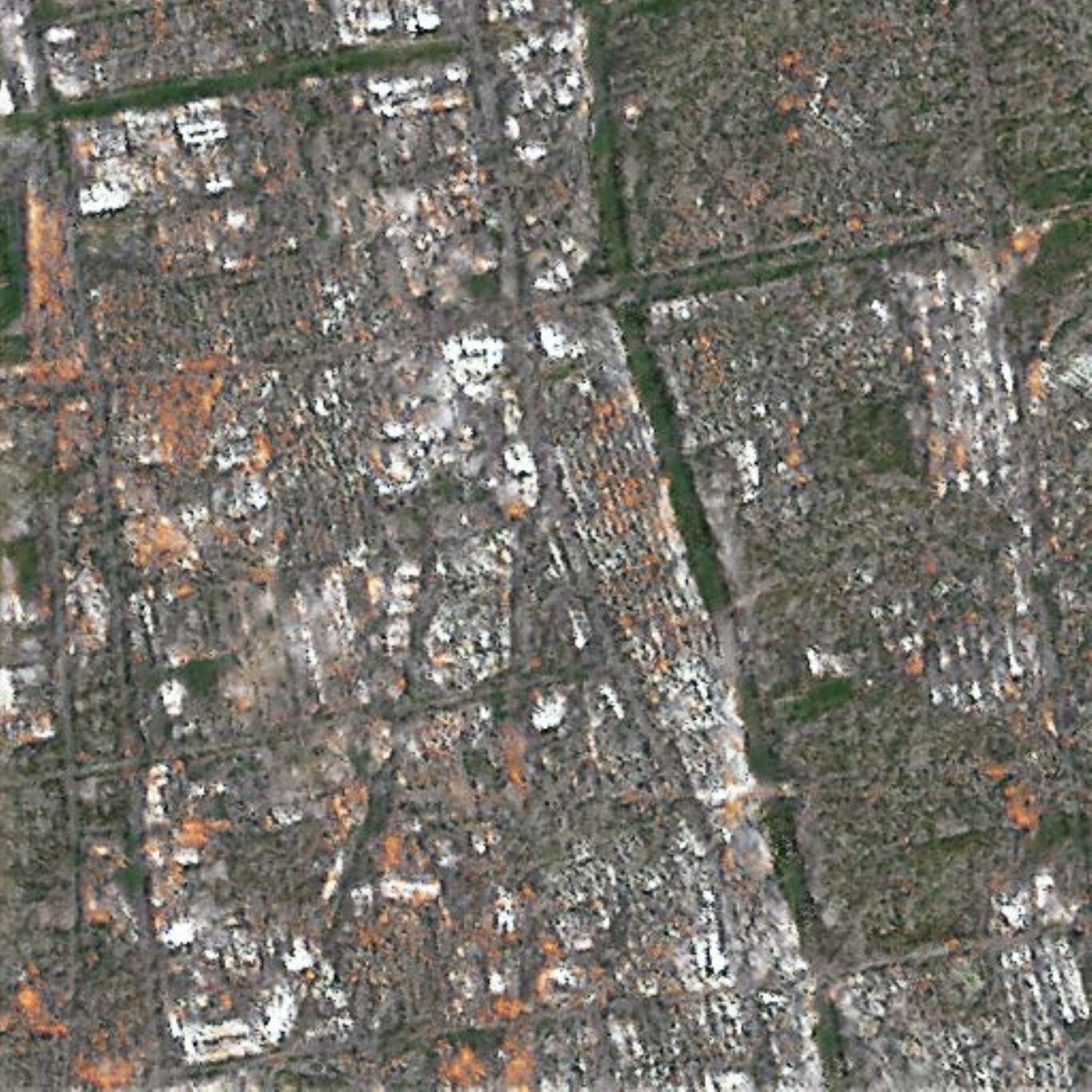}\vspace{4pt}
			\includegraphics[width=1.2in,height=1.2in]{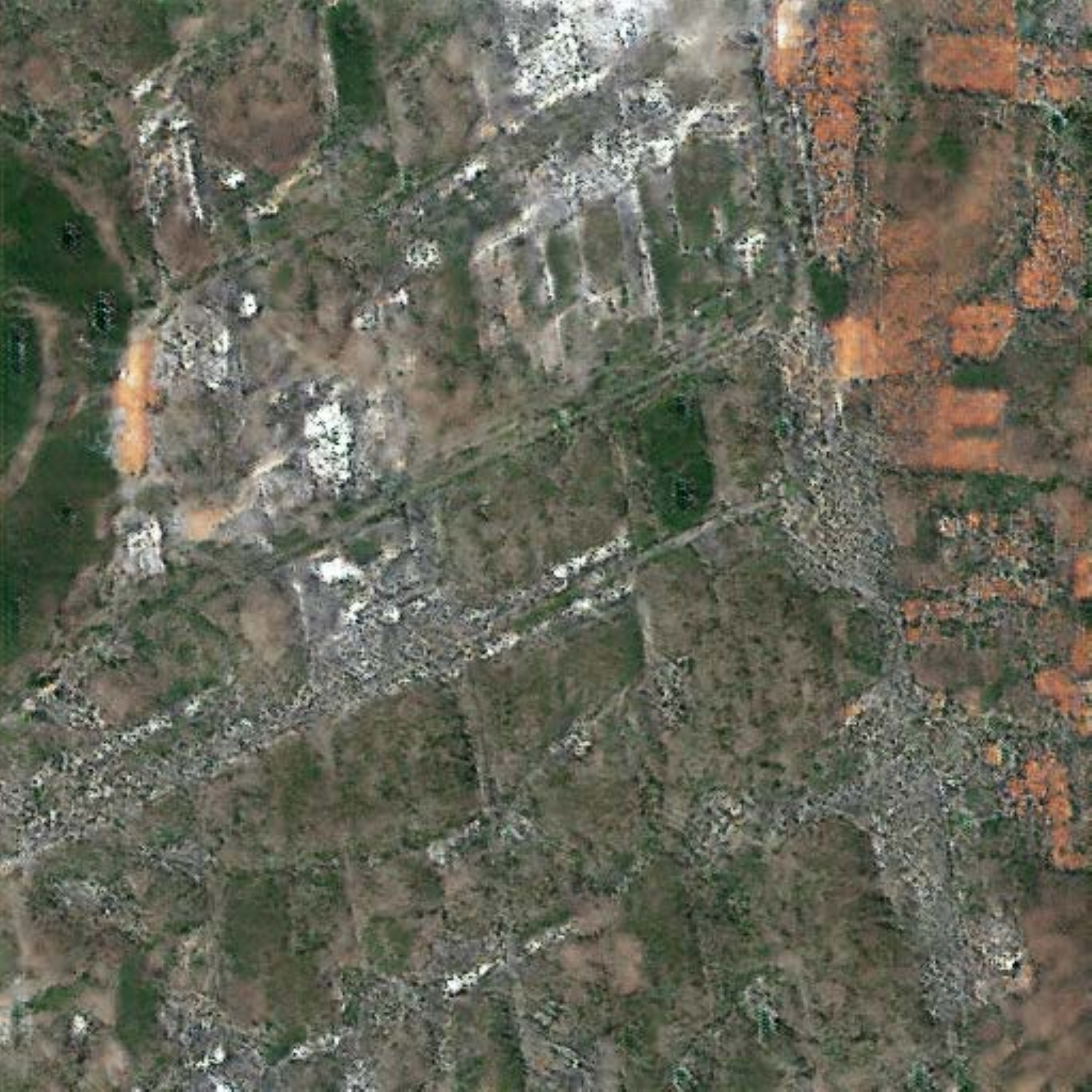}
			\centering{(b)}
		\end{minipage}
	}
	\subfigure{
		\begin{minipage}[b]{0.22\linewidth}
			\includegraphics[width=1.2in,height=1.2in]{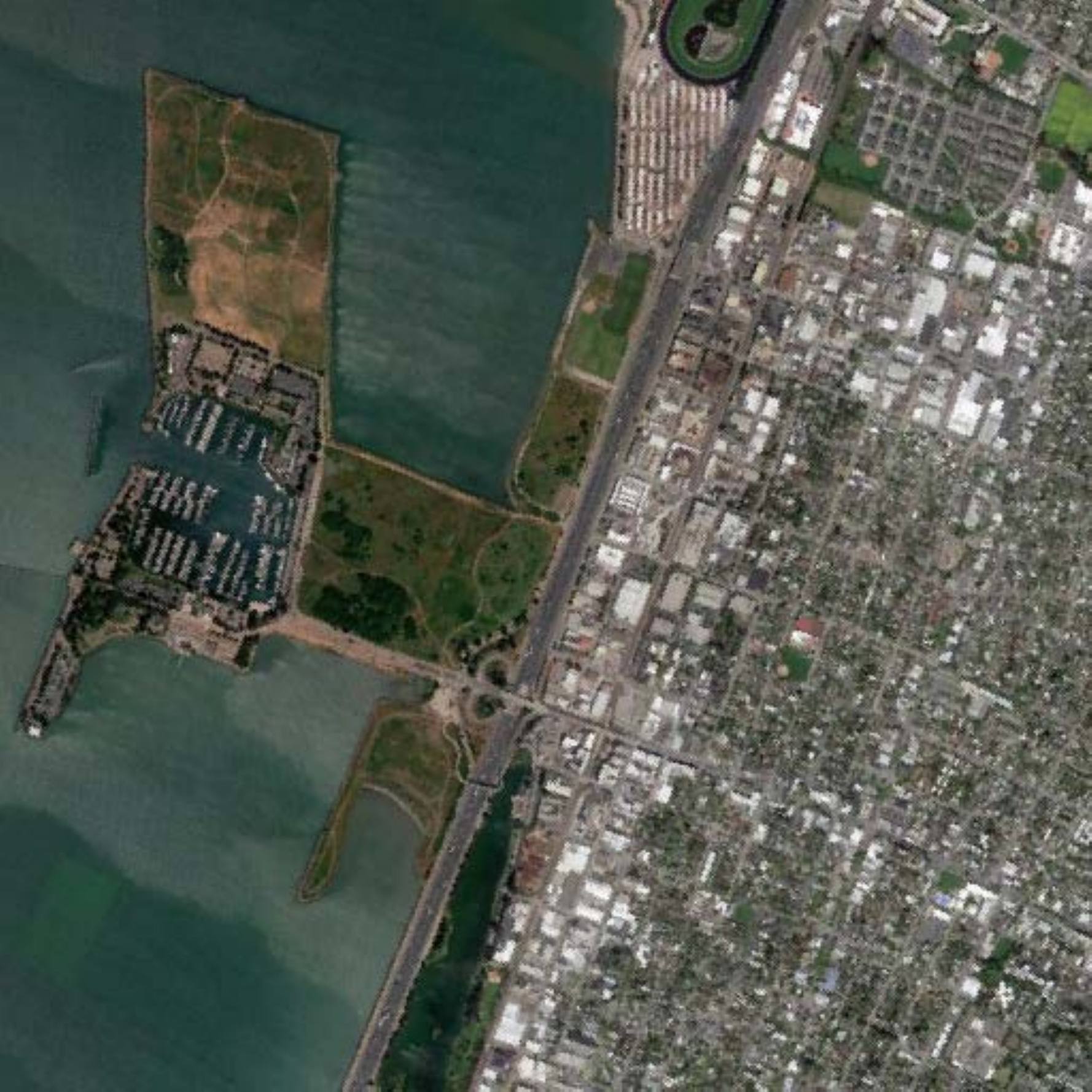}\vspace{4pt}
			\includegraphics[width=1.2in,height=1.2in]{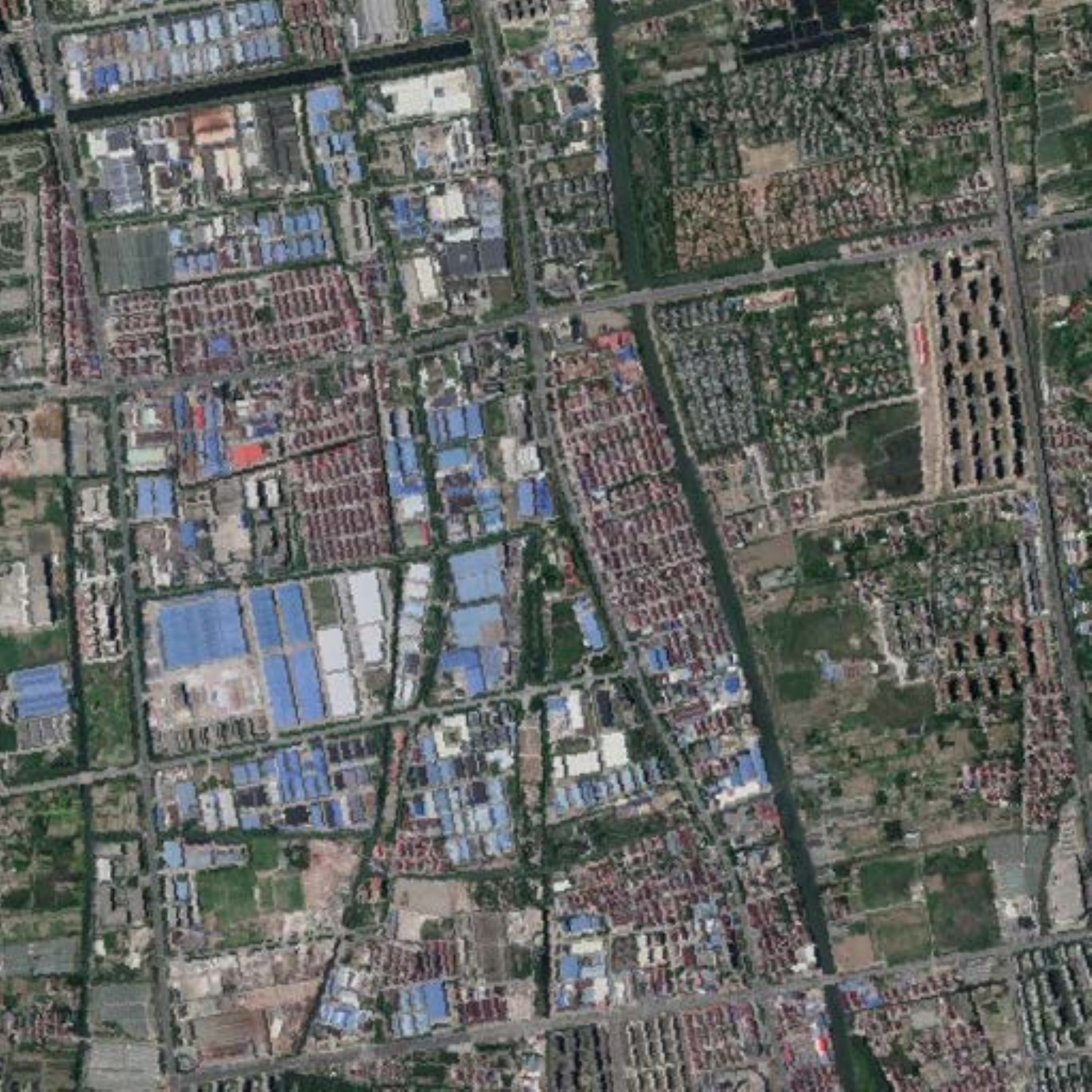}\vspace{4pt}
			\includegraphics[width=1.2in,height=1.2in]{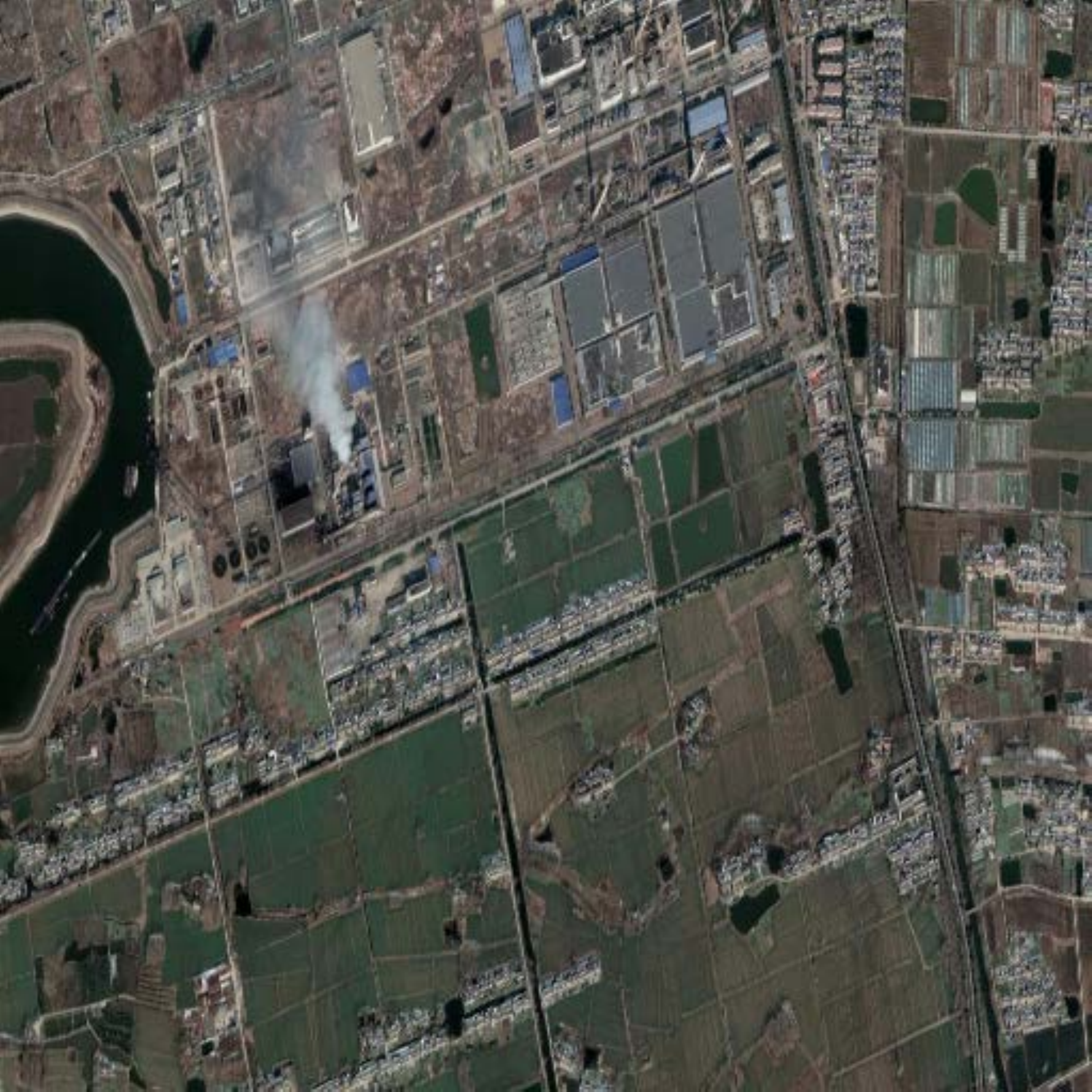}
			\centering{(c)}
		\end{minipage}
	}
	\subfigure{
		\begin{minipage}[b]{0.22\linewidth}
			\includegraphics[width=1.2in,height=1.2in]{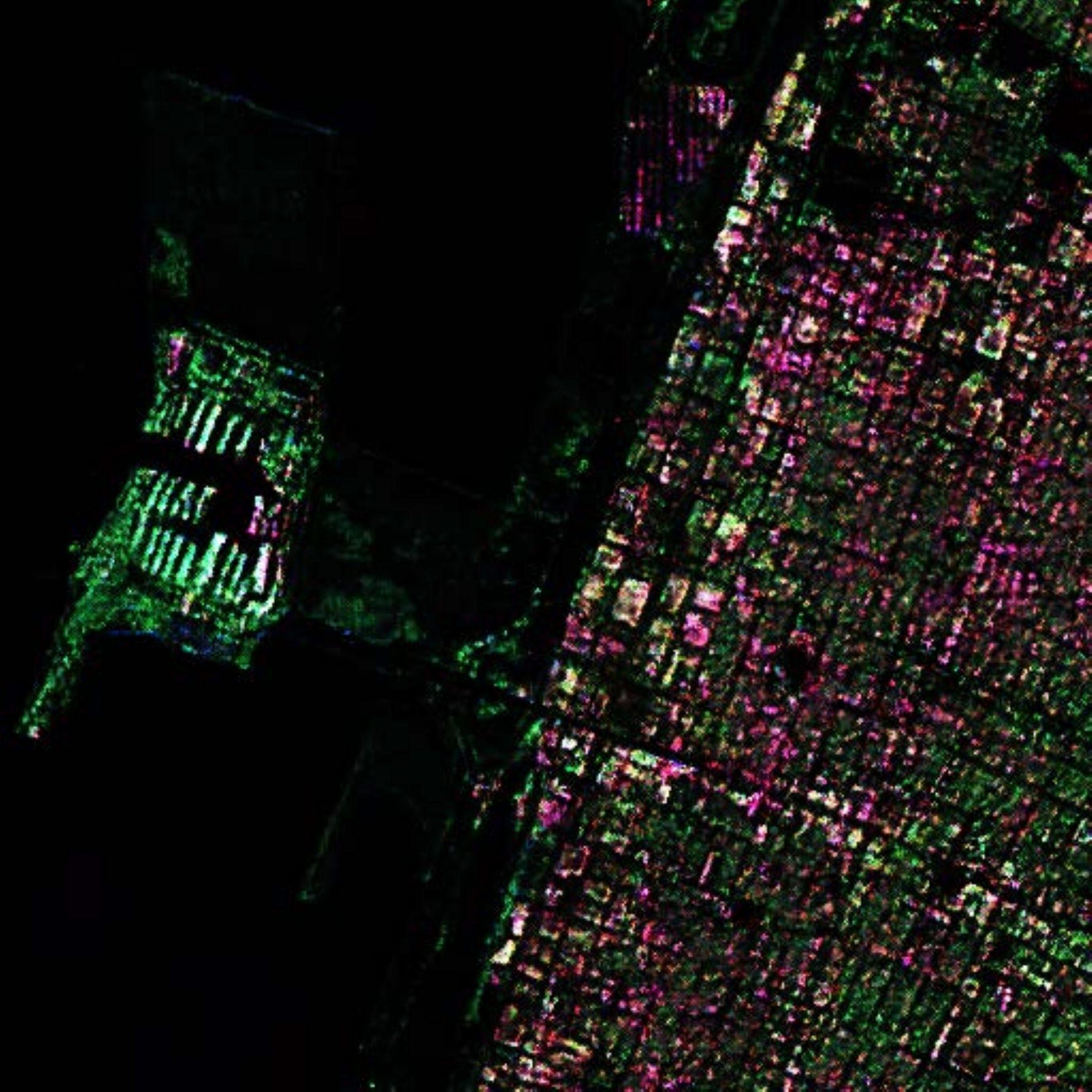}\vspace{4pt}
			\includegraphics[width=1.2in,height=1.2in]{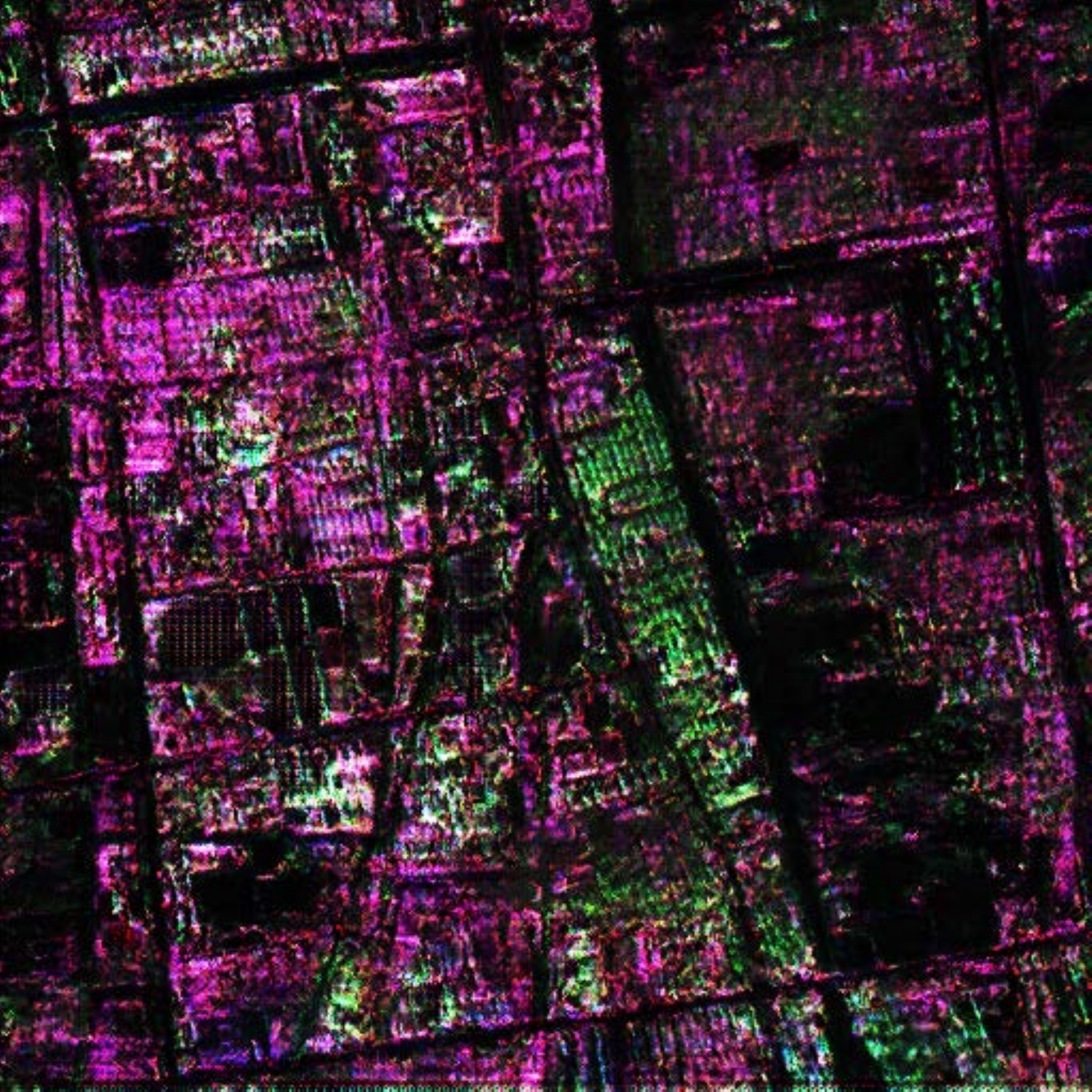}\vspace{4pt}
			\includegraphics[width=1.2in,height=1.2in]{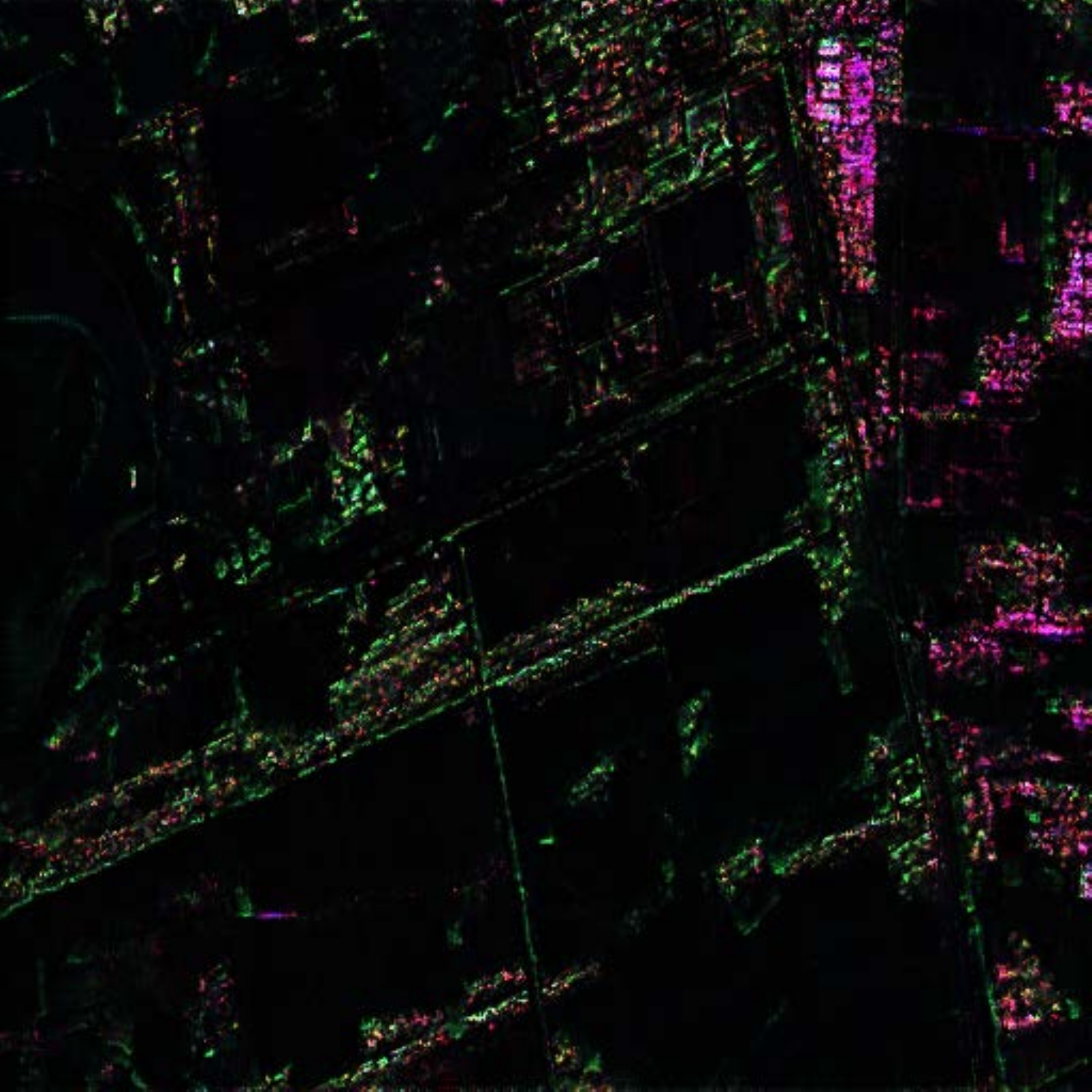}
			\centering{(d)}
		\end{minipage}
	}
	\caption{Images across scenes and across sensors reconstructed by the model pre-trained with 6m UAVSAR images. Images in each row from left to right are the \textbf{(a) real SAR image} and its \textbf{(b) translated optical image}, the \textbf{(c) real optical image} and its \textbf{(a) translated SAR image}. Each row lists a kind of dataset: \textbf{UAVSAR}, \textbf{GF3} and \textbf{ALOS2}.}
	\label{fig:figure17} 
\end{figure}

An experiment is conducted where the model trained using UAVSAR images is used to translate SAR images from UAVSAR, GF3 and ALOS2 acquired at different regions (\autoref{fig:figure17}). Compared to the ground truth, the performance of translation is largely degraded in the case of GF3 and ALOS2. The boundaries of different terrain surfaces are smeared. We believe that this is partially attributed to the fact that SAR images from different platforms are not cross-calibrated.

Note that when applied to processing real SAR image, we prefer to process one large image at a time. Although the network is trained and designed to take inputs of 256×256 patches, it is a fully convolutional network and can be directly extended to process larger size images without any modification. Experiments were conducted to verify the performance of the proposed method when used to process large size images.

\subsection{Enhancement with unsupervised Learning}
Finally, we explore the possibility of further refining the network with unsupervised learning. Note that the unsupervised training starts from the network trained with co-registered images. Then we can feed the SAR or optical images to be tested to the network and train them using large volume of unpaired optical or SAR images. Compared with supervised learning, in which only the prior knowledge from pretraining can be utilized, the model of unsupervised learning can also dynamically learn something new from the extended dataset and refine the results through iterations.

The major experimental procedures are as follows.
\begin{itemize}
	\item Randomly select $n$ pairs of optical and SAR images outside the dataset to be tested. Ensure that the earth surfaces are evenly distributed (slightly more buildings for their difficulty to be reconstructed);
	\item Feed the $N$ test SAR images and the $n$ optical images to the unsupervised network, train until the early stop and save the translated optical images;
	\item Feed the $N$ test optical images and the $n$ SAR images to the unsupervised network, train until the early stop and save the translated SAR images;
	\item Check the results and quantitatively evaluate them with those by supervised learning.
\end{itemize}

\begin{figure}[!htb]
	\scriptsize
	\centering
	\subfigure{
		\begin{minipage}[b]{0.12\linewidth}
			\includegraphics[width=0.7in,height=0.7in]{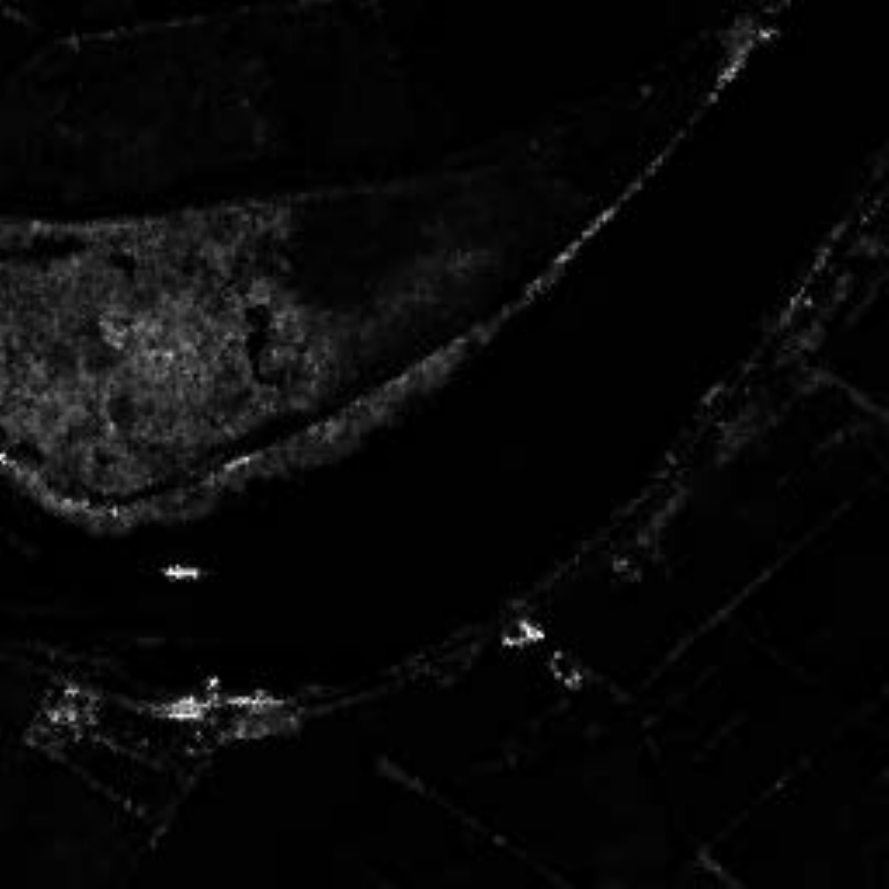}\vspace{4pt}
			\includegraphics[width=0.7in,height=0.7in]{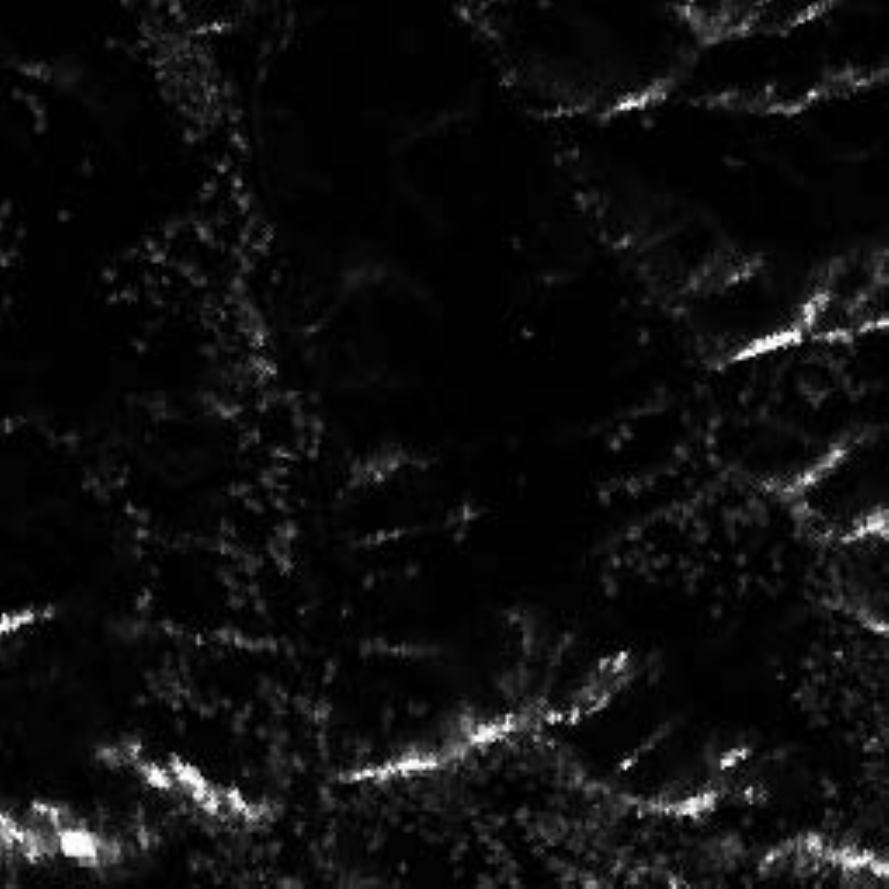}\vspace{4pt}
			\includegraphics[width=0.7in,height=0.7in]{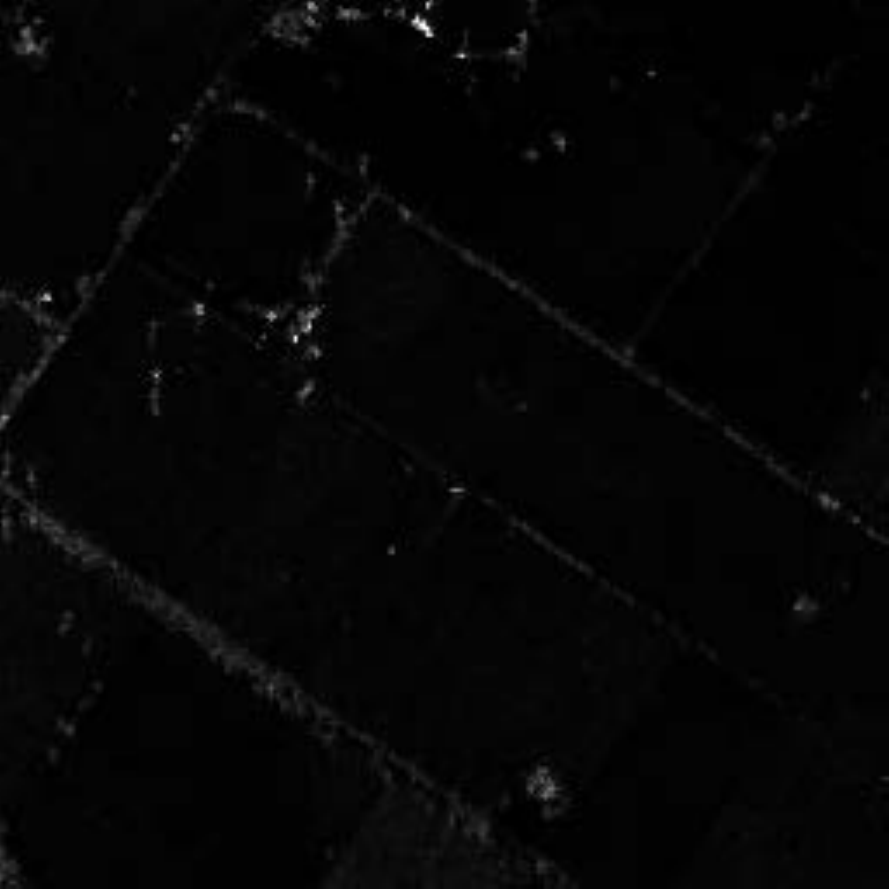}\vspace{4pt}
			\includegraphics[width=0.7in,height=0.7in]{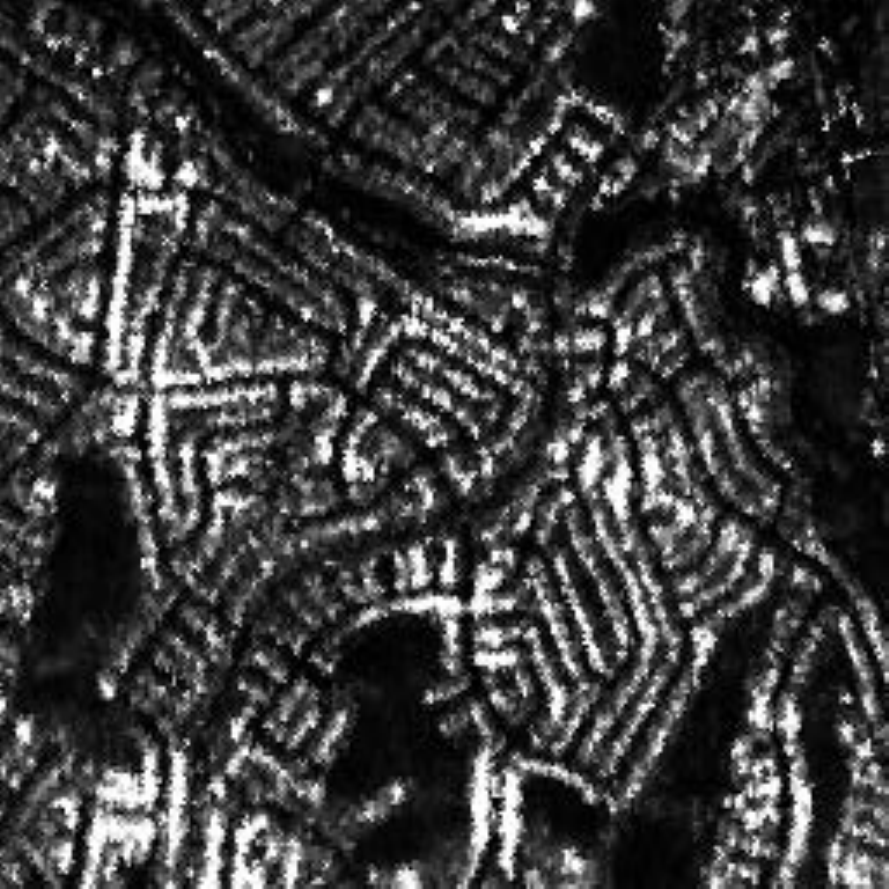}
			\centering{(a)}
		\end{minipage}
	}
	\subfigure{
		\begin{minipage}[b]{0.12\linewidth}
			\includegraphics[width=0.7in,height=0.7in]{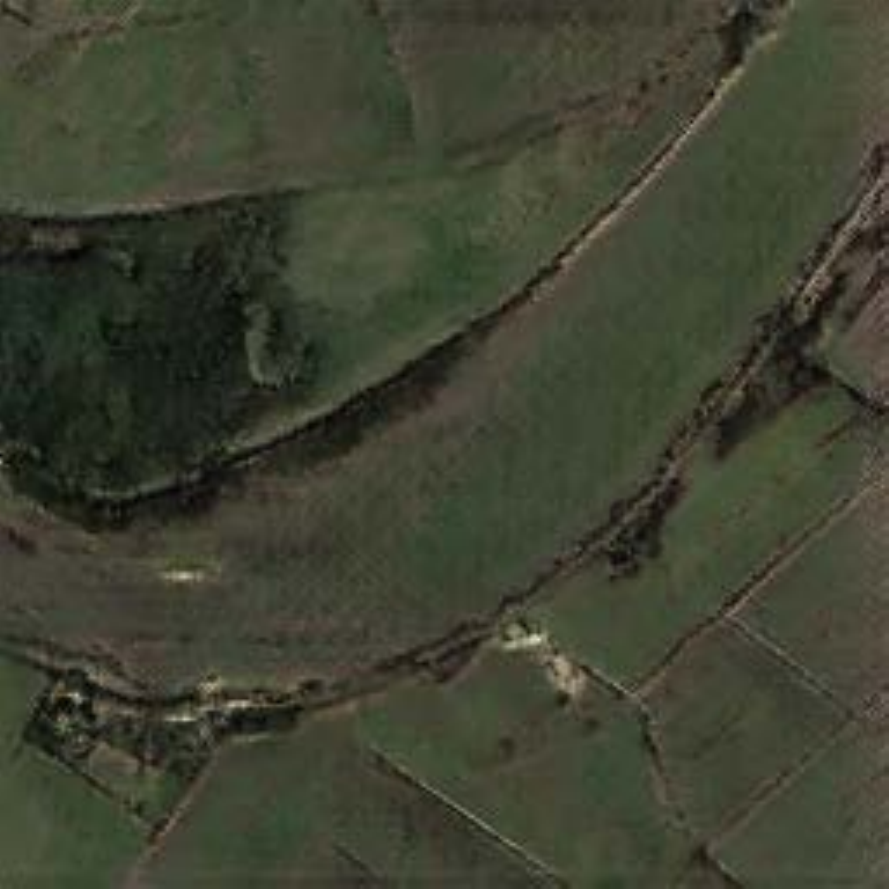}\vspace{4pt}
			\includegraphics[width=0.7in,height=0.7in]{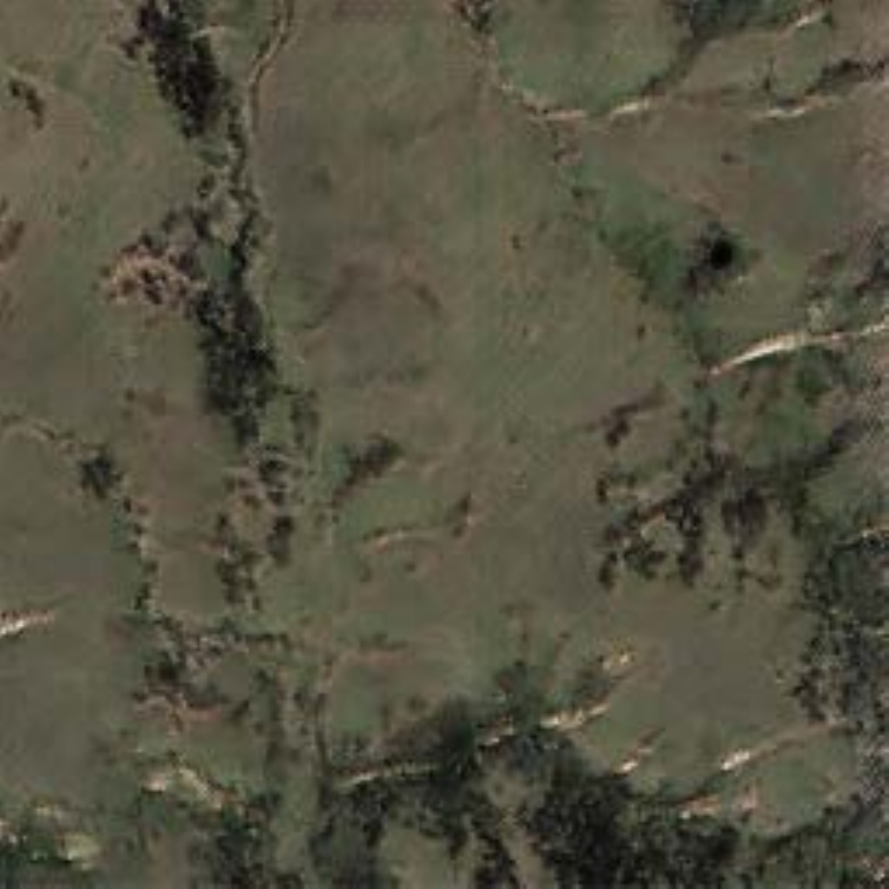}\vspace{4pt}
			\includegraphics[width=0.7in,height=0.7in]{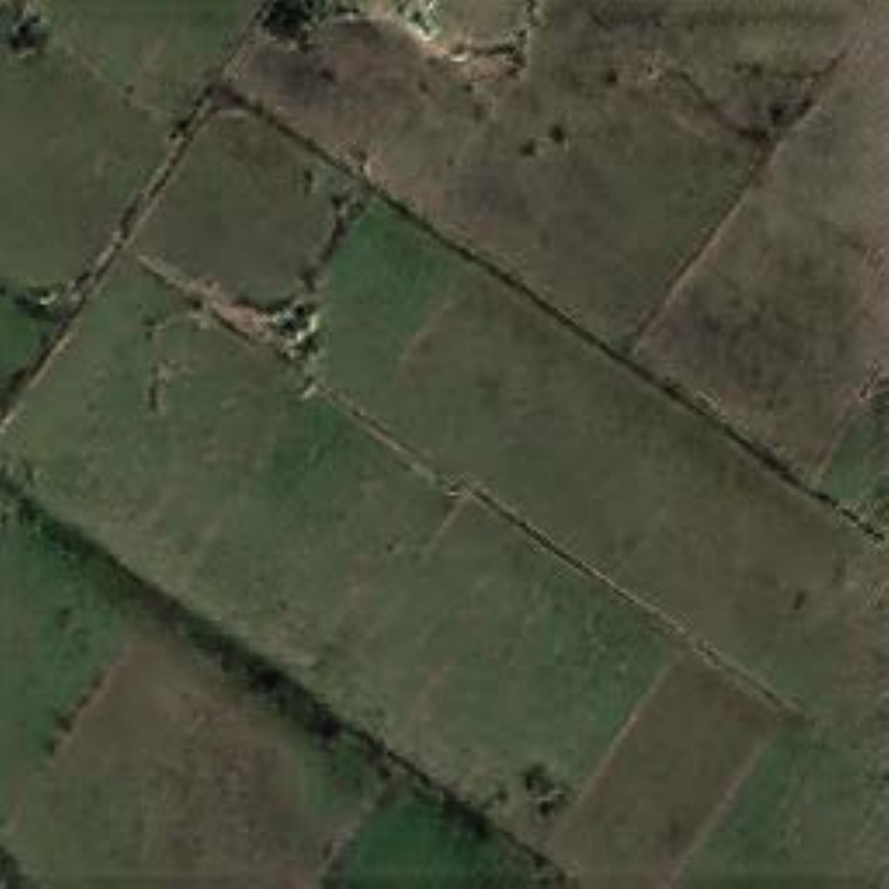}\vspace{4pt}
			\includegraphics[width=0.7in,height=0.7in]{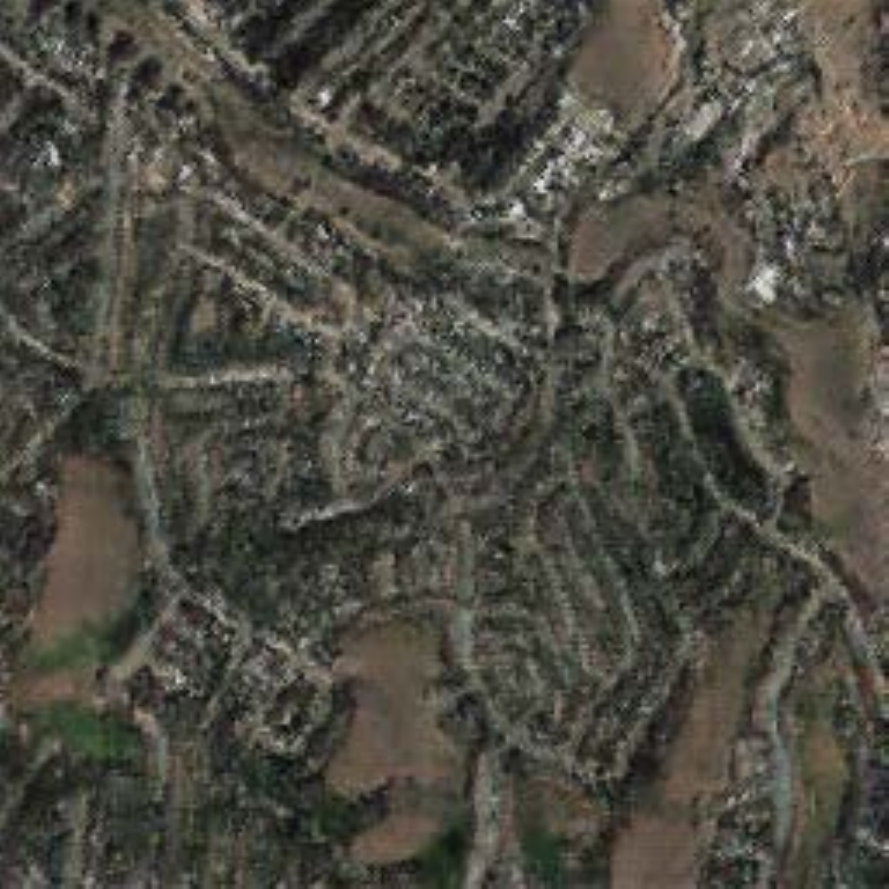}
			\centering{(b)}
		\end{minipage}
	}
	\subfigure{
		\begin{minipage}[b]{0.12\linewidth}
			\includegraphics[width=0.7in,height=0.7in]{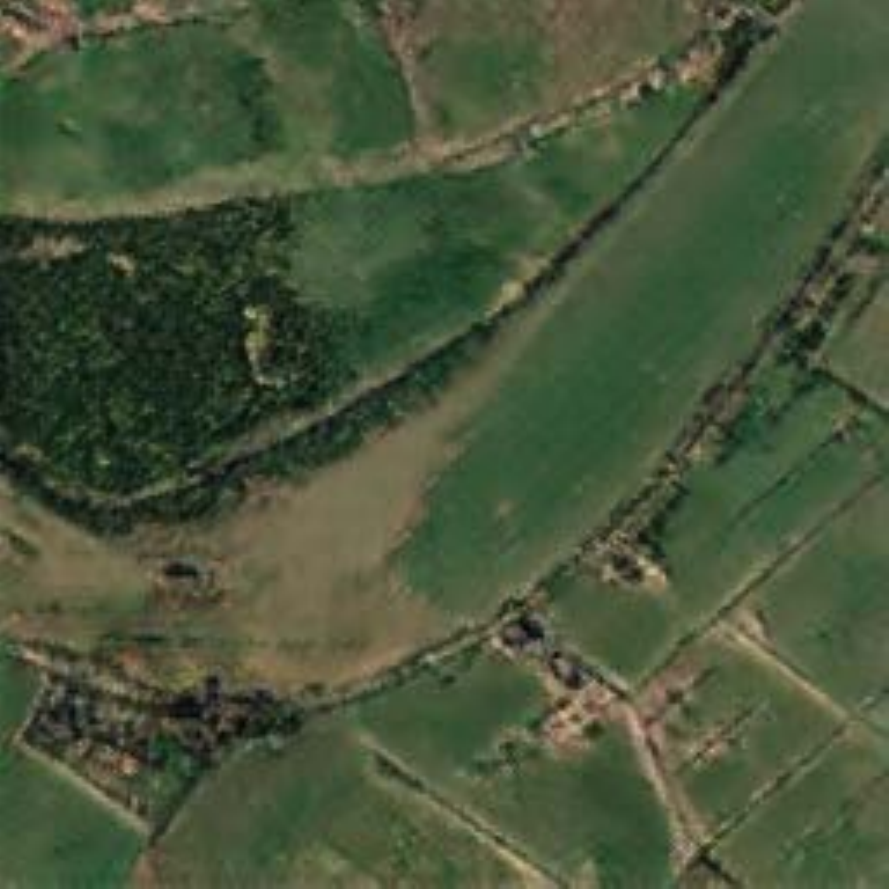}\vspace{4pt}
			\includegraphics[width=0.7in,height=0.7in]{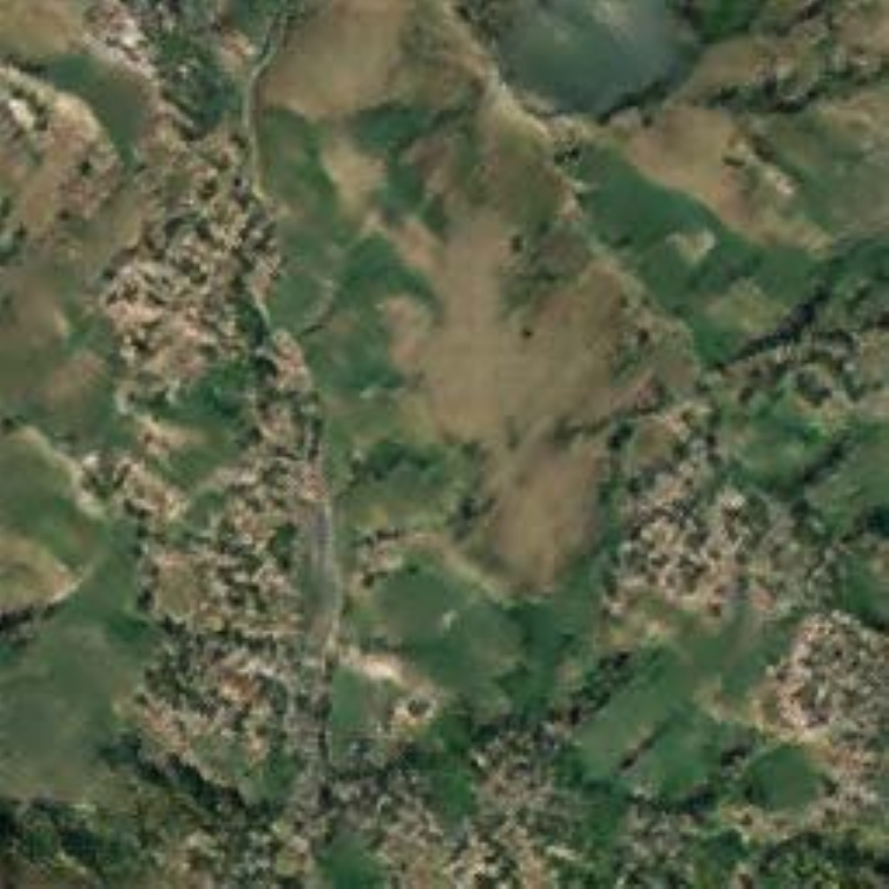}\vspace{4pt}
			\includegraphics[width=0.7in,height=0.7in]{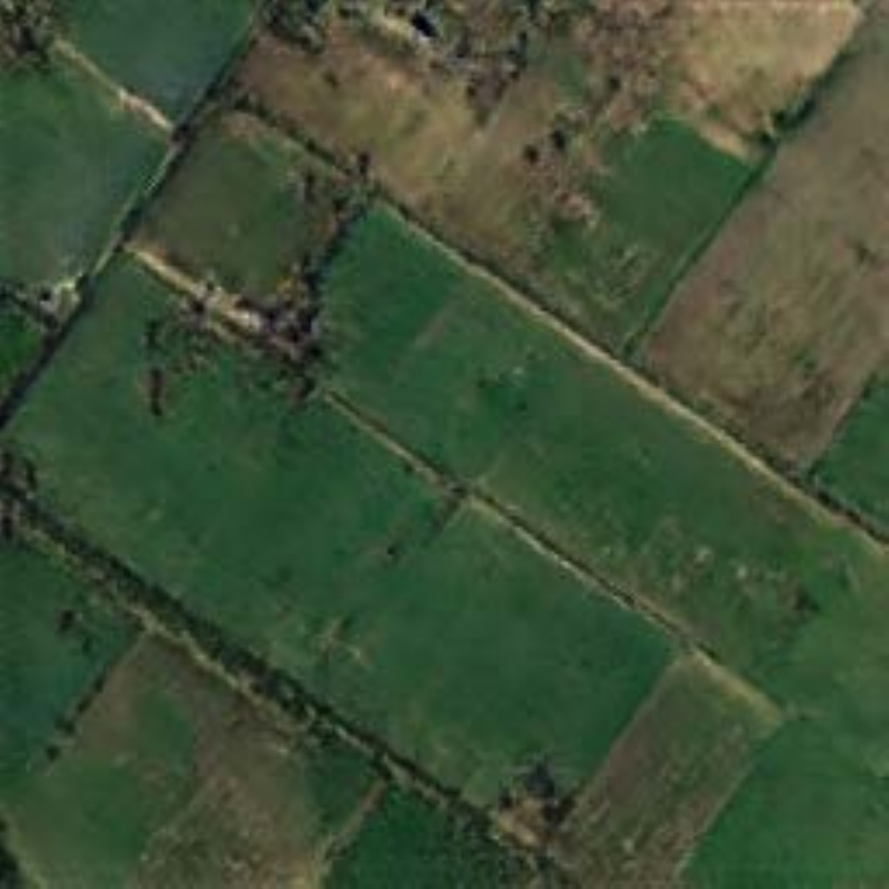}\vspace{4pt}
			\includegraphics[width=0.7in,height=0.7in]{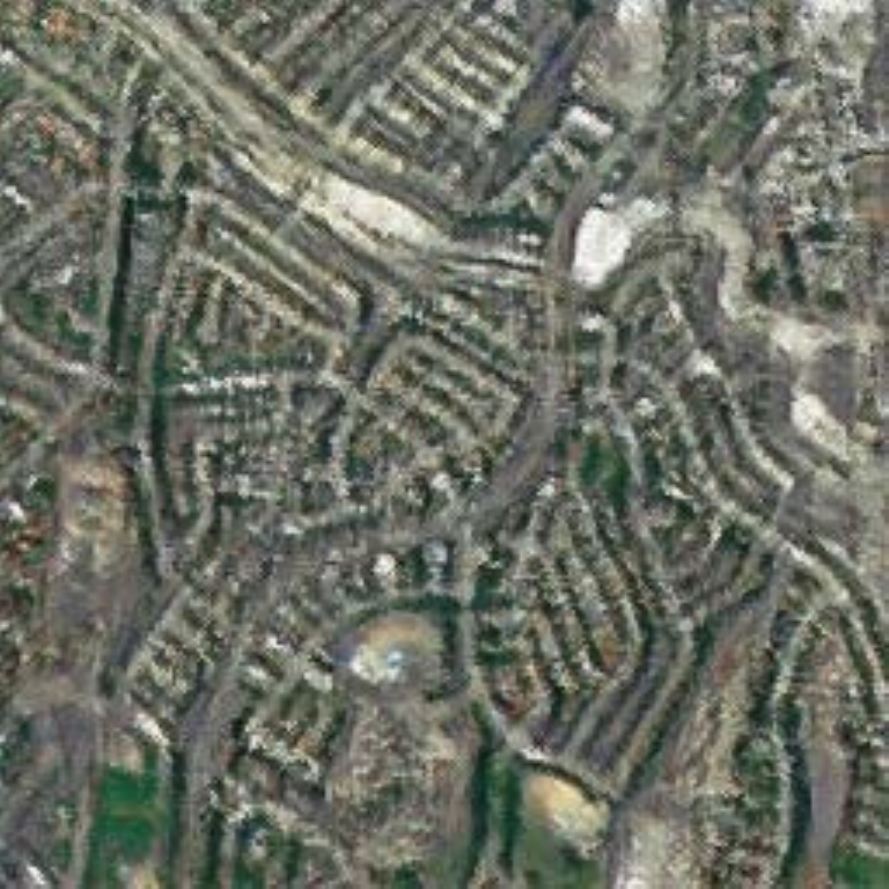}
			\centering{(c)}
		\end{minipage}
	}
	\subfigure{
		\begin{minipage}[b]{0.12\linewidth}
			\includegraphics[width=0.7in,height=0.7in]{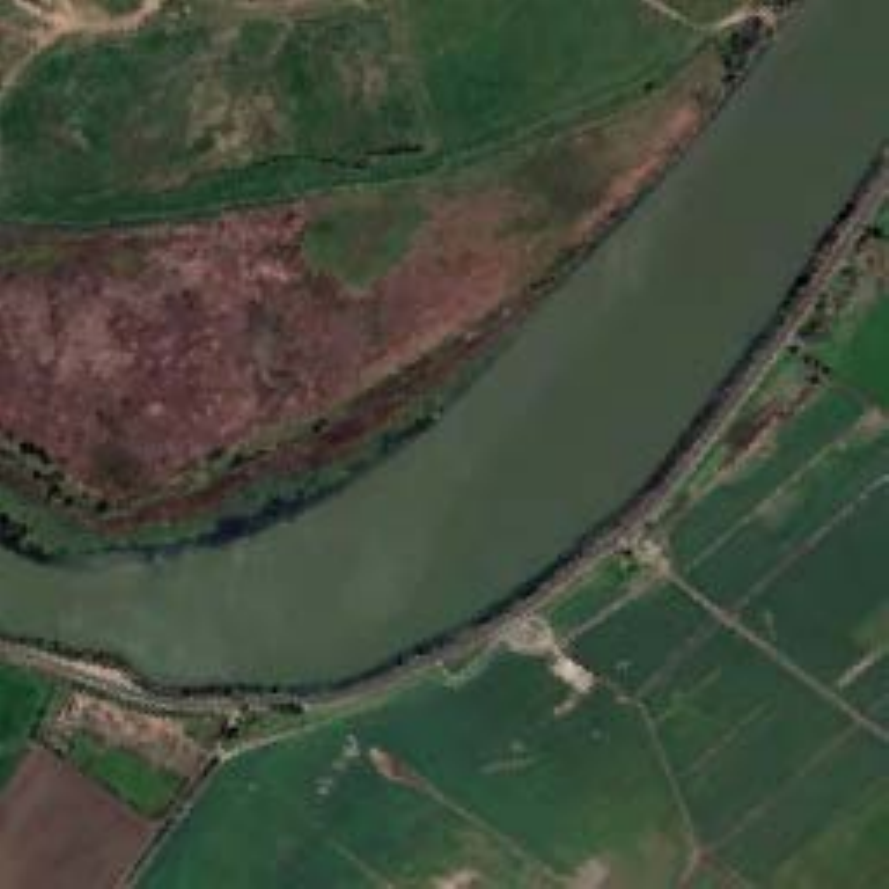}\vspace{4pt}
			\includegraphics[width=0.7in,height=0.7in]{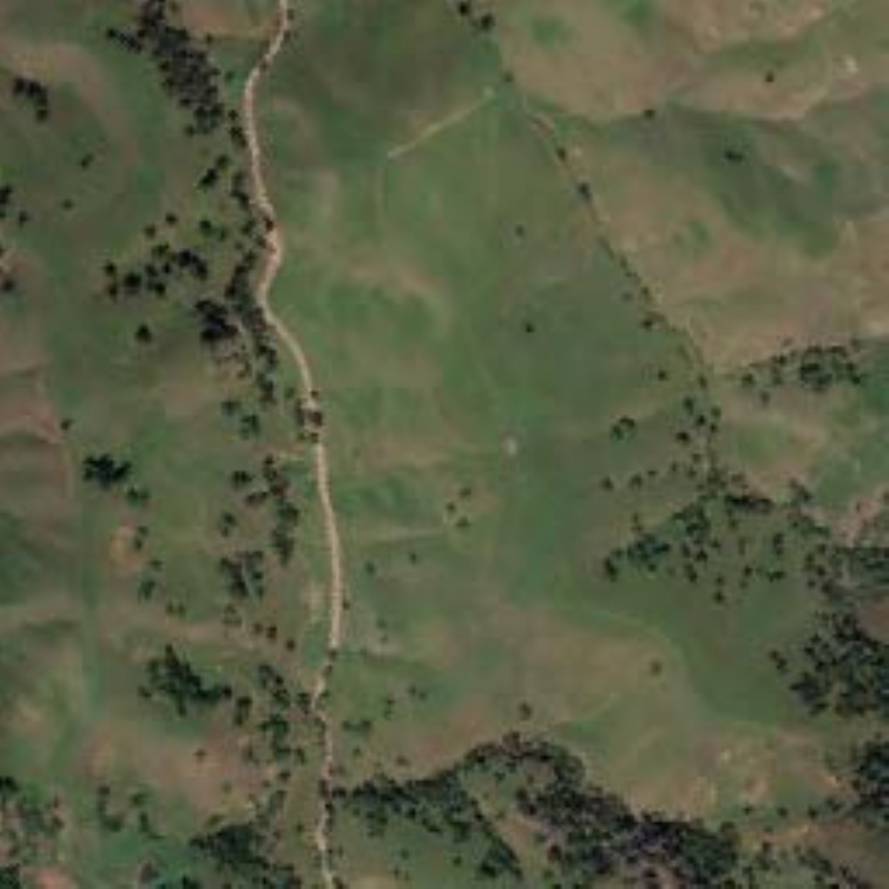}\vspace{4pt}
			\includegraphics[width=0.7in,height=0.7in]{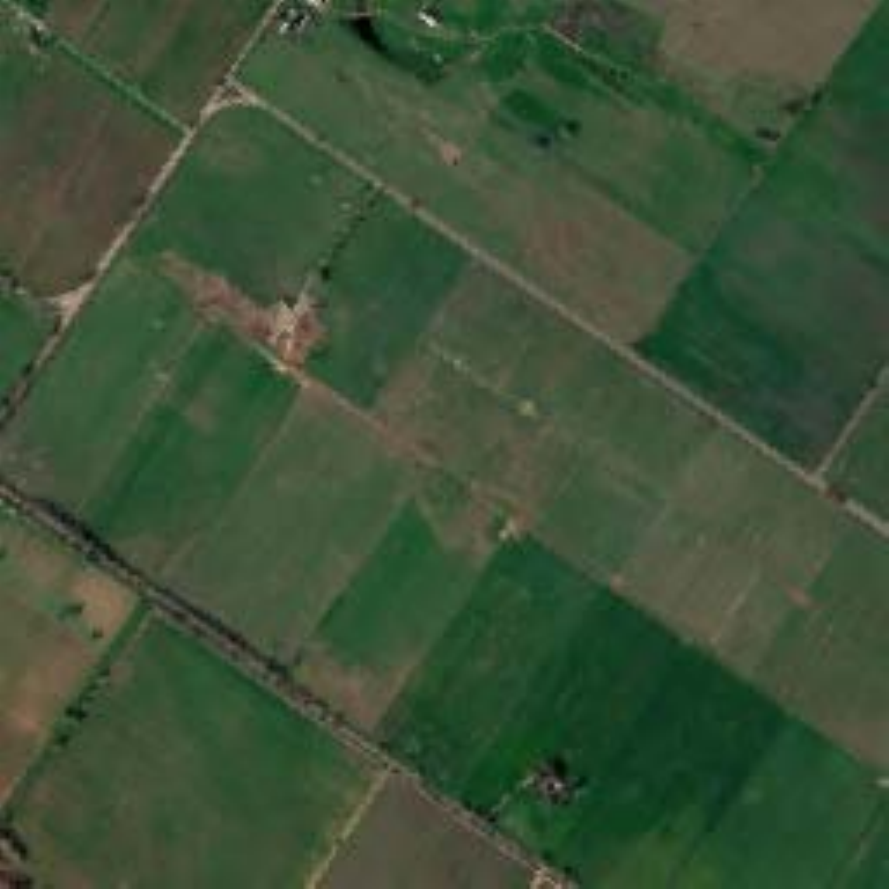}\vspace{4pt}
			\includegraphics[width=0.7in,height=0.7in]{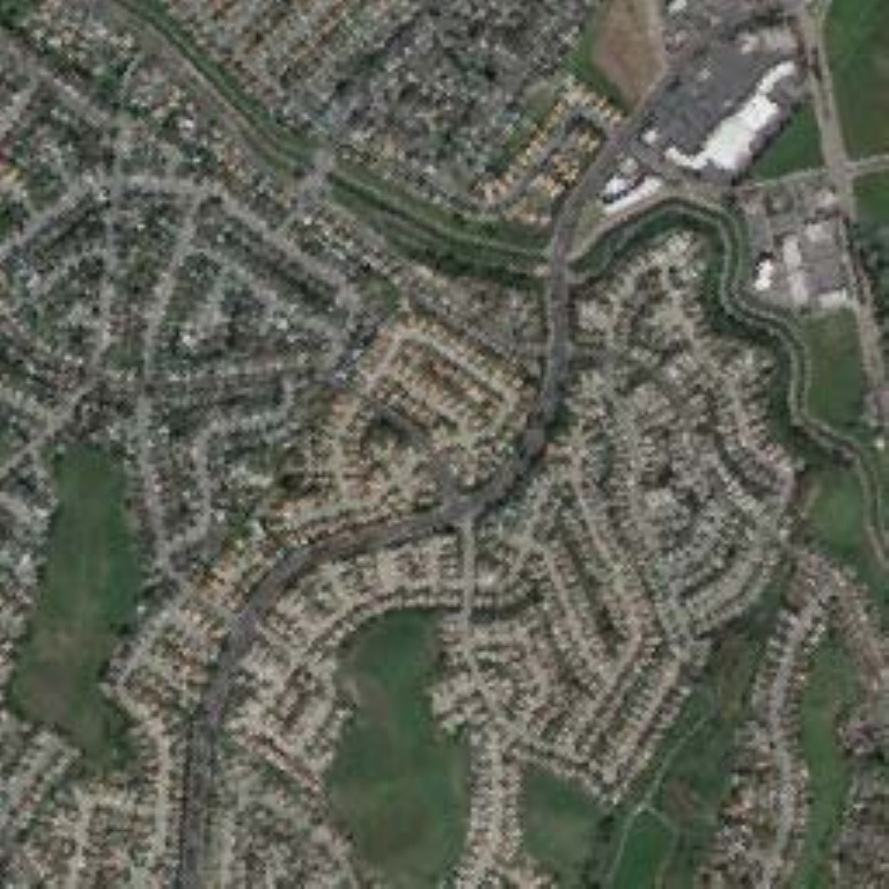}
			\centering{(d)}
		\end{minipage}
	}
	\subfigure{
		\begin{minipage}[b]{0.12\linewidth}
			\includegraphics[width=0.7in,height=0.7in]{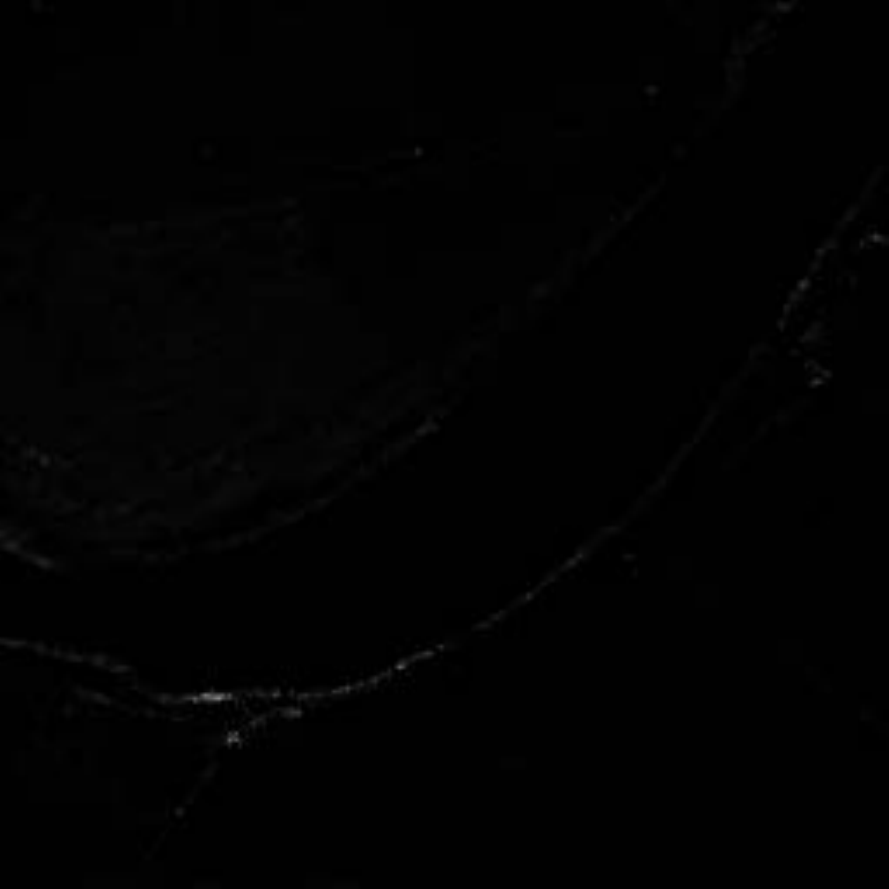}\vspace{4pt}
			\includegraphics[width=0.7in,height=0.7in]{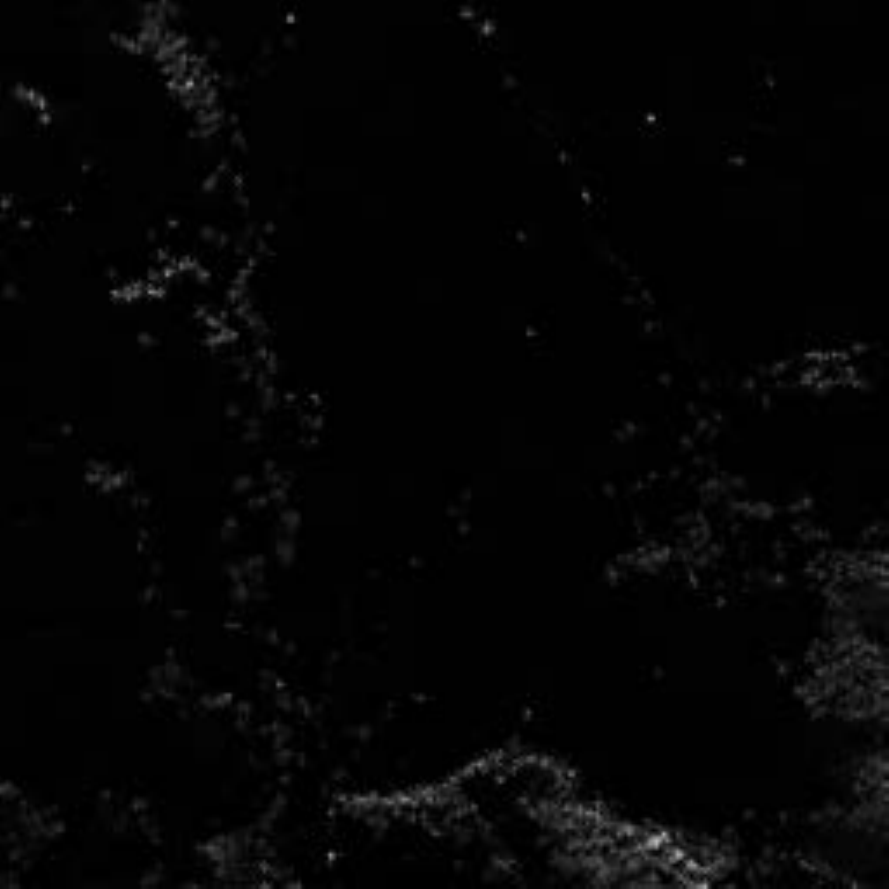}\vspace{4pt}
			\includegraphics[width=0.7in,height=0.7in]{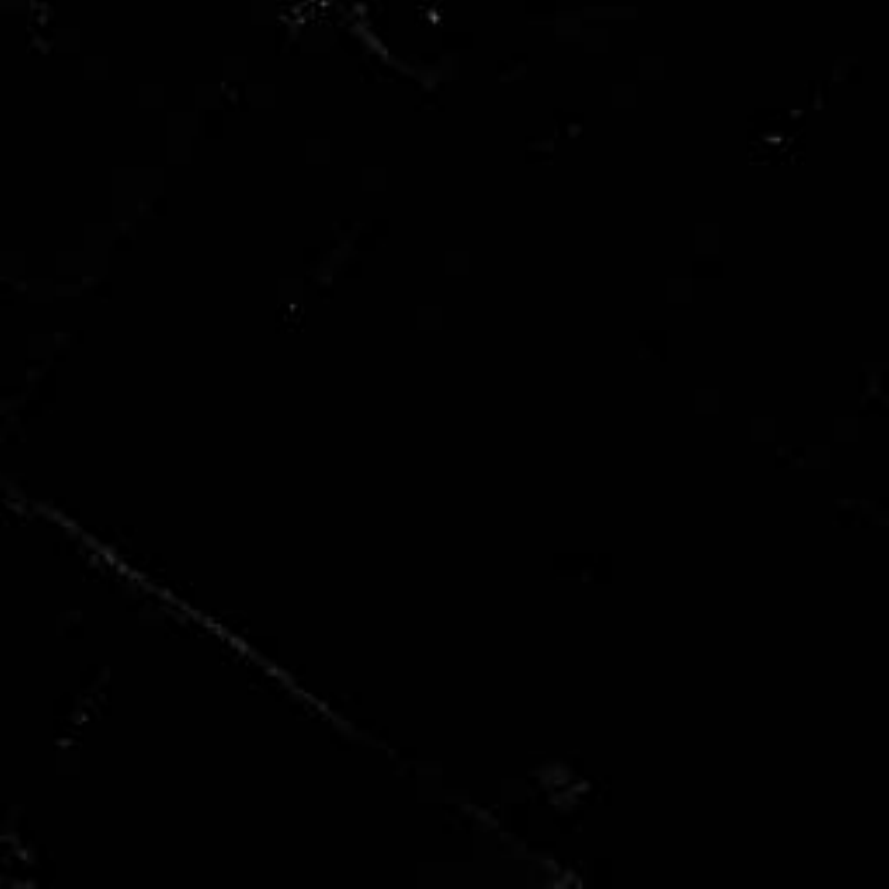}\vspace{4pt}
			\includegraphics[width=0.7in,height=0.7in]{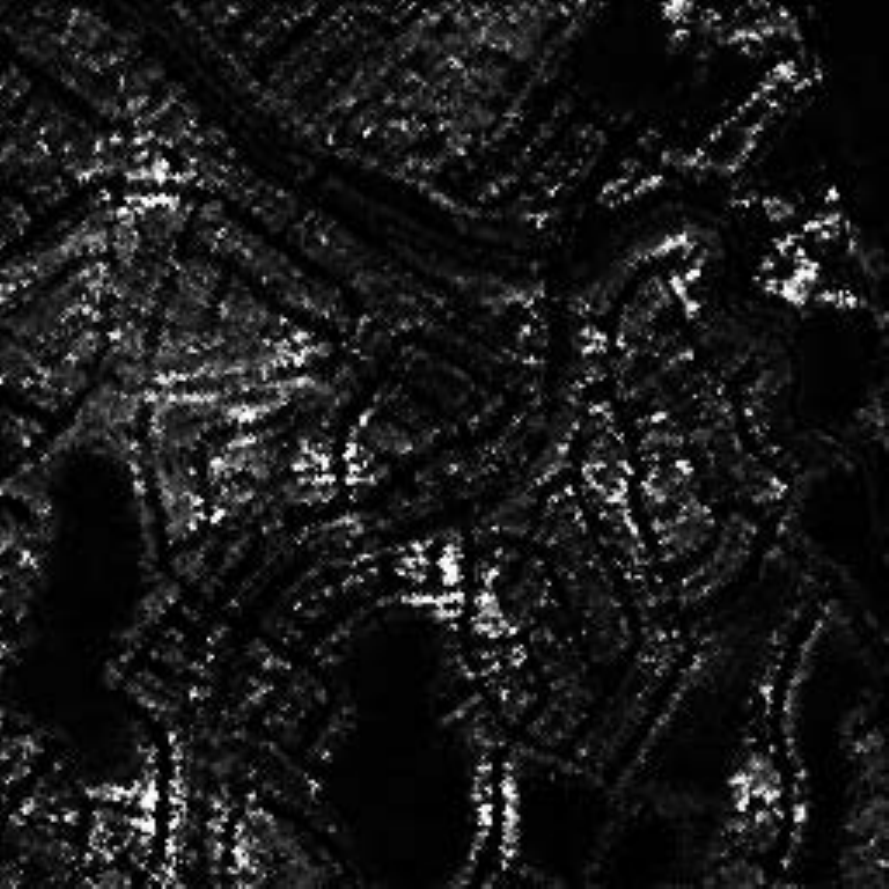}
			\centering{(e)}
		\end{minipage}
	}
	\subfigure{
		\begin{minipage}[b]{0.12\linewidth}
			\includegraphics[width=0.7in,height=0.7in]{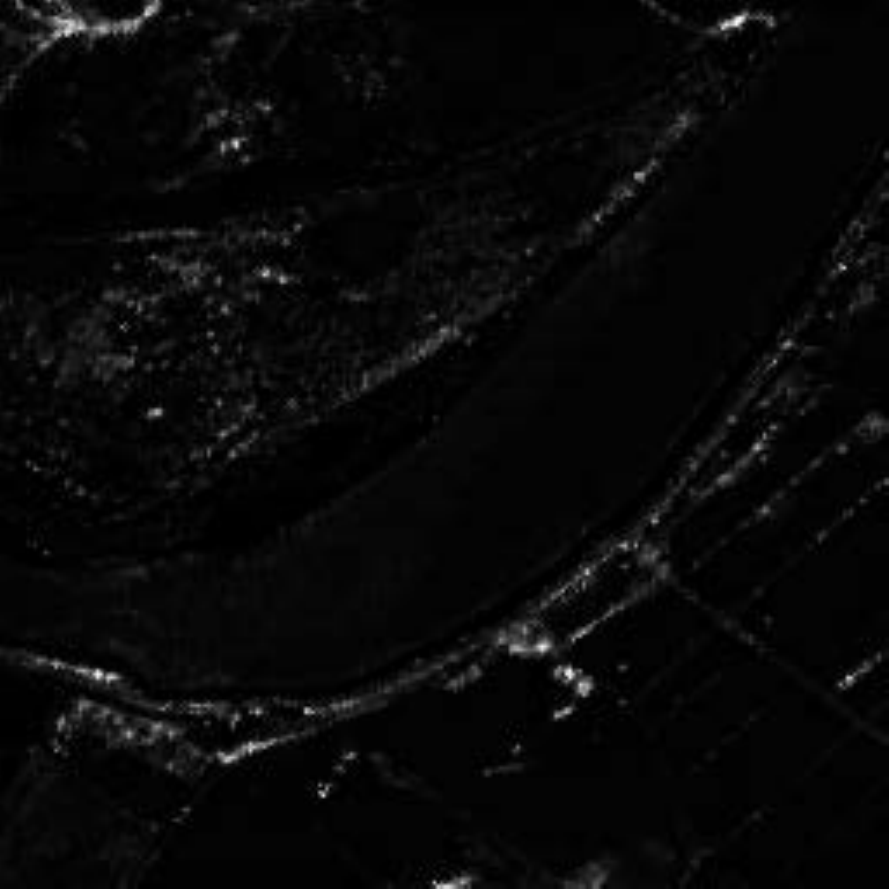}\vspace{4pt}
			\includegraphics[width=0.7in,height=0.7in]{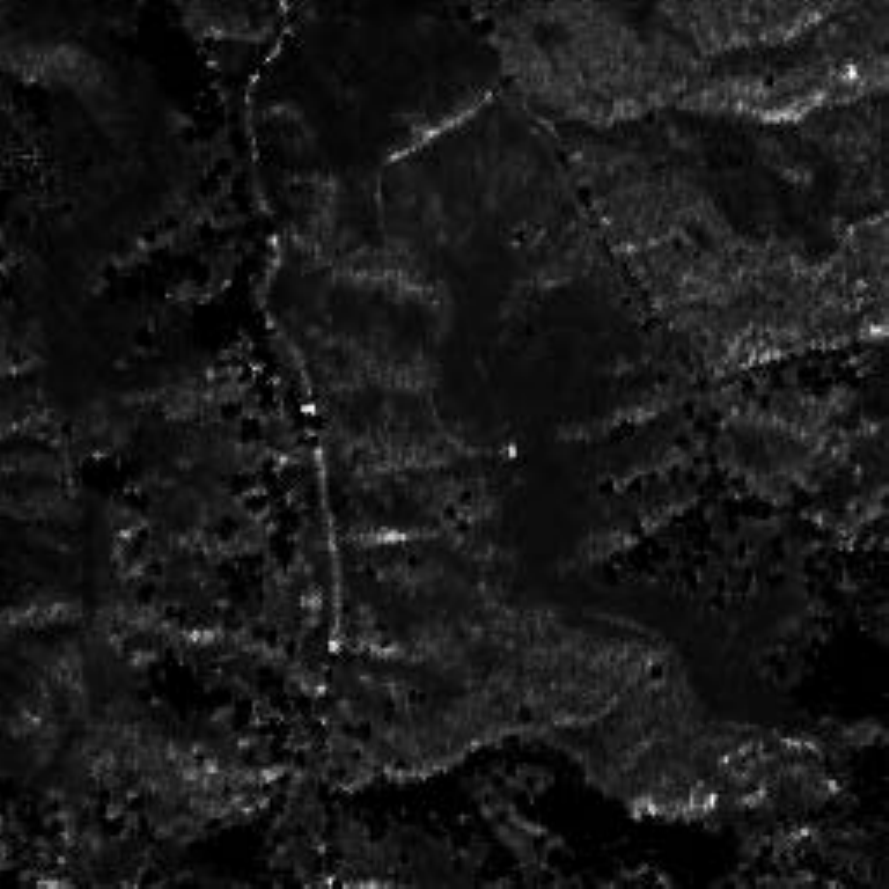}\vspace{4pt}
			\includegraphics[width=0.7in,height=0.7in]{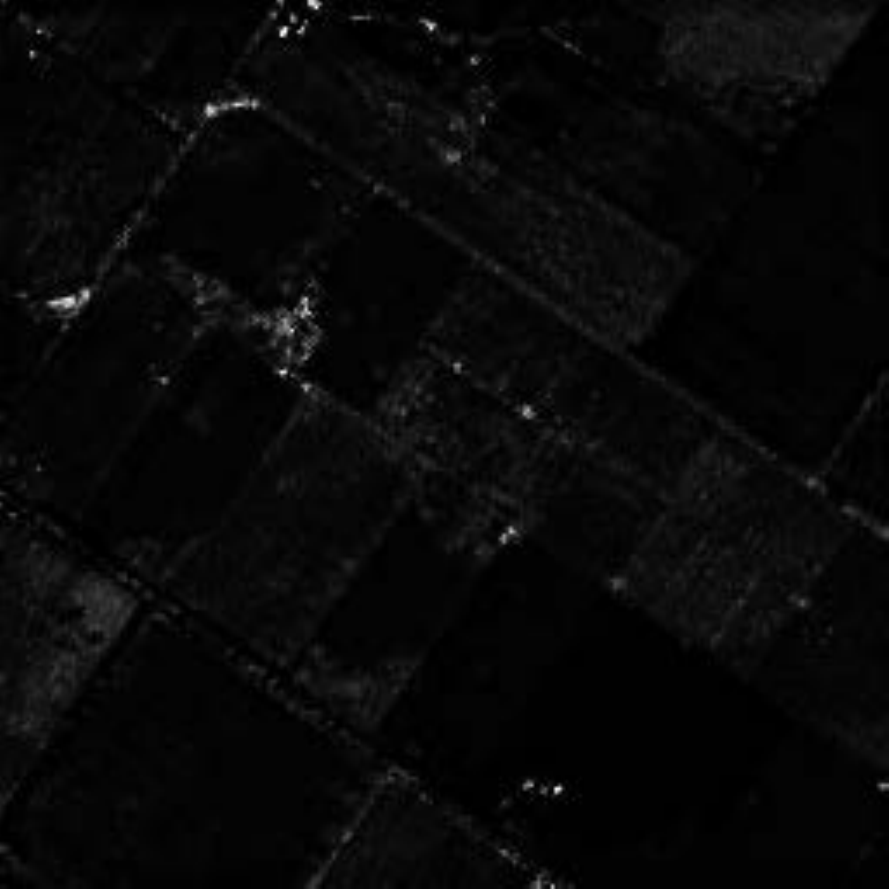}\vspace{4pt}
			\includegraphics[width=0.7in,height=0.7in]{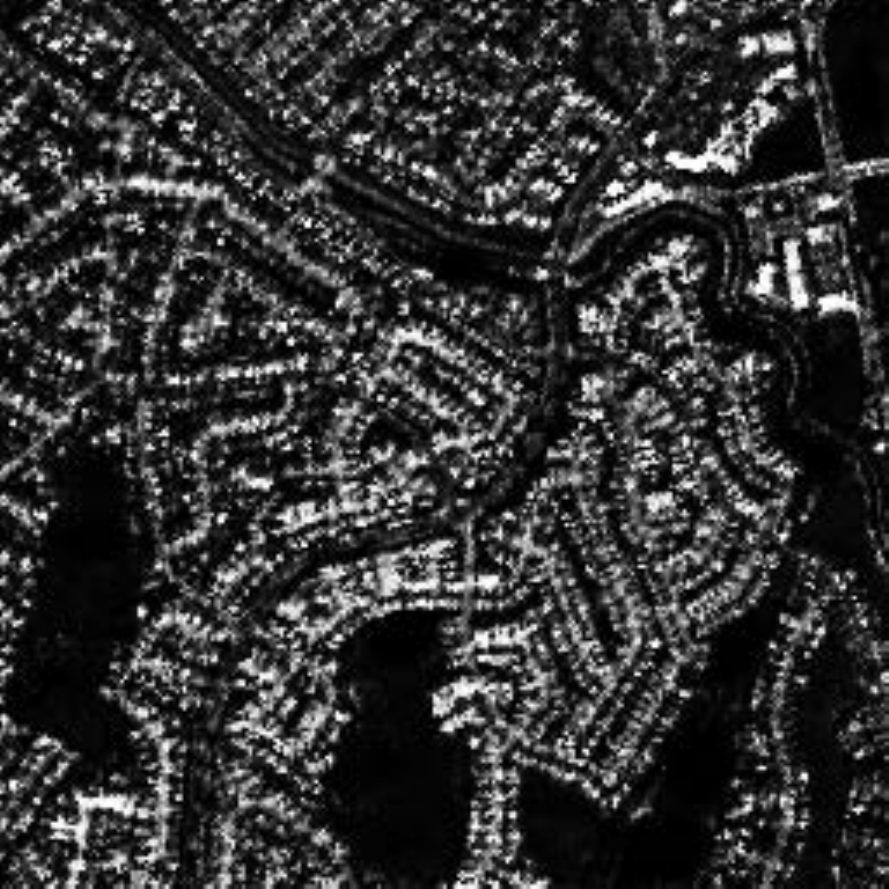}
			\centering{(f)}
		\end{minipage}
	}
	\caption{Translated images further refined with unsupervised learning. Images in each row from left to right are the \textbf{(a) input SAR image}, the \textbf{(b) translated optical image} and the further \textbf{(c) refined optical image} by unsupervised learning, the \textbf{(d) input optical image} and its \textbf{(e) translated SAR image} and the further \textbf{(f) refined optical image} by unsupervised learning. Each row lists a kind of earth surfaces: waters, vegetation, farmlands and buildings.}
	\label{fig:figure18} 
\end{figure}

As shown in \autoref{table6}, it indicates that the translation results are greatly improved with unsupervised learning. The results further refined with unsupervised training are shown in \autoref{fig:figure18}, where we can see that the refined results are more vivid and realistic. However, waters in SAR images don’t differ from those farmlands greatly, which results in the imperfect reconstruction of waters.

\begin{table}
	\scriptsize
	\renewcommand\arraystretch{1.5}
	\setlength{\abovecaptionskip}{0pt}
	\setlength{\belowcaptionskip}{10pt}
	\caption{FIDs of results by supervised and unsupervised learning.}
	\label{table6} 
%	\centering %
	\begin{tabular}{*{3}{m{1.3in}<{\centering}}}
%		\toprule[2pt]
		\hline
		 & Supervised Learning & +Unsupervised Learning \\
		\hline
		Optical & 107.8 & 88.9 \\
		SAR & 58.1 & 41.2 \\
		\hline
%		\bottomrule[2pt]
	\end{tabular}
\end{table}

\subsection{Computational cost}
In this section, the computational performances of the three translation networks, including the computational complexity and the speed of processing images per second, are analyzed. Note that for neural networks, the computation cost is about the same for training and inference per image. Thus, only training performance is analyzed here.

Deep learning models are intensive in resource consumption, mainly measured by the number of trainable parameters and the number of float operations. All the translation networks used are reciprocal. It should be noted that only the convolutional layers are considered, and the trainable parameters and operations generated by LeakyReLU and BN are ignored. \autoref{table7} indicates that the numbers of parameters are almost same, but the number of operations in CycleGAN is approximately twice that of Pix2Pix and CRAN, due to the additional cyclic loop.
\begin{table}
	\scriptsize
	\renewcommand\arraystretch{1.5}
	\setlength{\abovecaptionskip}{0pt}
	\setlength{\belowcaptionskip}{10pt}%设置标题与表格的距离
	\caption{Number of trainable parameters and operations in the three translation networks.}
	\label{table7} 
%	\centering
	\begin{tabular}{ccc}
		\hline
		Model & Number of Parameters & Number of Operations / FLOPs \\
		\hline
		
	    CycleGAN Generator & 113.73M & 152.39M \\
		
		Pix2Pix Generator & 107.16M & 89.50G \\
		
		CRAN Generator & 107.49M & 79.41G \\
		
		Discriminator & 5.35M & 6.53M \\
		\hline
	\end{tabular}
\end{table}

\begin{figure}
	\centering 
	\includegraphics[width=6in]{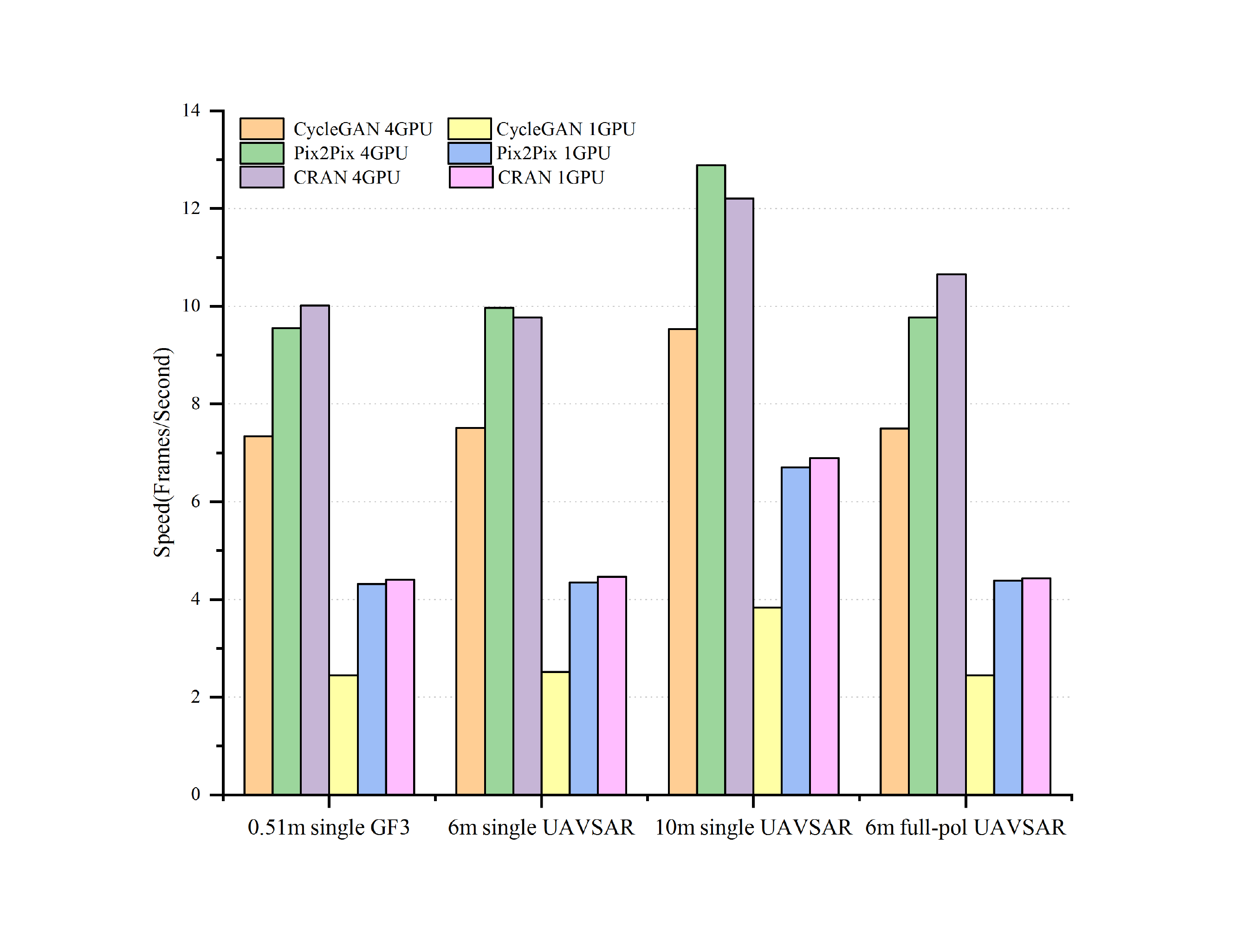} 
	\caption{Speed comparison between different methods, different methods and different number of GPUs. On each dataset, the three translation networks respectively run on four GPUs and one single GPU.}
	\label{fig:figure19} 
\end{figure}

The networks are all implemented on TensorFlow and run on Ubuntu server with 4 Titan X. Here we compare how many pictures can be processed per second respectively by the three methods. From \autoref{fig:figure19} we can find:
\begin{itemize}
	\item For the same method, the training speed using 4 GPUs is approximately $2\sim 3$ times of that using 1 GPU. This is due to the communication overhead and some part of computation that cannot be run distributedly.
	\item The speed of CRAN and Pix2Pix is much faster than CycleGAN, which agrees with the analyses given in \autoref{table7}.
\end{itemize}

\section{Conclusion}
For the purpose of assisted interpretation of SAR imagery by ordinary people, this paper proposes an image translation network architecture for reciprocal translation between SAR and optical remote sensing images. In order to evaluate the translated images from the perspective of human visual perception, the quantitative metric FID is employed. For low-resolution (6m, 10m) UAVSAR dataset, the reconstructed images appear very similar to the true data and the corresponding FID is low. For high-resolution (0.51m) GF3 dataset, the reconstructed results appear reasonable but not exactly capture the geometric features of certain built-up objects such as high-rise buildings. Under the same condition, the proposed network outperforms conventional image translation networks such as CycleGAN and Pix2Pix. Results also show that full-pol SAR image is preferable as input for translation because certain objects are not observable in single-pol SAR images. It is also confirmed that the network does not perform well if generalized across different SAR platforms. Finally, we demonstrate that unsupervised learning could further improve the performance of a translator initially trained with a small number of co-registered image pairs which points the right direction towards general application of assisted SAR image interpretation.

\section*{Acknowledgement}
This work was supported in part by National Key $R\&D$ Program of China no. 2017YFB0502703 and Natural Science Foundation of China no. 61822107, 61571134.

\bibliography{mybibfile}
%\section*{References}
%\begin{thebibliography}{1}
%\bibliography{mybibfile}
%\bibitem{bibitem44}
%Dataset: Gaofen3, China Centre for Resources Satellite Data and Application. Accessed: Jul. 2017 [Online]. Available: http://www.cresda.com/CN/.
%\bibitem{bibitem45}
%Dataset: UAVSAR, NASA 2018, ASF DAAC. Accessed: Dec. 2018 [Online]. Available: https://vertex.daac.asf.alaska.edu/.
%\end{thebibliography}

\end{document}